\documentclass[times,twocolumn,final]{elsarticle}

\usepackage{medima}
\usepackage{multirow}
\usepackage{framed,multirow}
\usepackage[utf8]{inputenc}
\usepackage{amssymb}
\usepackage{latexsym}
\usepackage{graphicx}
\usepackage{supertabular}
\usepackage{url}
\usepackage{xcolor}
\usepackage{longtable}
\usepackage{hyperref}

\definecolor{newcolor}{rgb}{.8,.349,.1}

\journal{Medical Image Analysis}
\DeclareUnicodeCharacter{2212}{-}
\begin{document}

\verso{G J Chowdary \textit{et~al.}}

\begin{frontmatter}

\title{Machine Learning and Deep Learning Methods for Building Intelligent Systems in Medicine and Drug Discovery: A Comprehensive Survey }

\author[1]{Jignesh Chowdary \snm{G}}

\author[1]{Suganya \snm{G}\corref{cor1}}
\cortext[cor1]{Corresponding author: 
  Tel.: +91 9894862359;  
  }
\ead{suganya.g@vit.ac.in}
\author[1]{Premalatha \snm{M}}
\author[1]{Asnath Victy Phamila \snm{Y}}
\author[1]{Karunamurthy \snm{K}}

\address[1]{Vellore Institute of Technology Chennai, Tamilnadu, India}

\received{ }
\finalform{ }
\accepted{ }
\availableonline{ }
\communicated{Suganya G}

\begin{abstract}
With the advancements in computer technology, there is a rapid development of intelligent systems to understand the complex relationships in data to make predictions and classifications. Artificail Intelligence based framework is rapidly revolutionizing the healthcare industry. These intelligent systems are built with machine learning and deep learning based robust models for early diagnosis of diseases and demonstrates a promising supplementary diagnostic method for frontline clinical doctors and surgeons. Machine Learning and Deep Learning based systems can streamline and simplify the steps involved in diagnosis of diseases from clinical and image-based data, thus providing significant clinician support and workflow optimization. They mimic human cognition and are even capable of diagnosing diseases that cannot be diagnosed with human intelligence. This paper focuses on the survey of machine learning and deep learning applications in across 16 medical specialties, namely Dental medicine, Haematology, Surgery, Cardiology, Pulmonology, Orthopedics, Radiology, Oncology, General medicine, Psychiatry, Endocrinology, Neurology, Dermatology, Hepatology, Nephrology, Ophthalmology, and Drug discovery. In this paper along with the survey, we discuss the advancements of medical practices with these systems and also the impact of these systems on medical professionals.

\end{abstract}


\end{frontmatter}

\section{Background}
\begin{figure}[!t]
\centering
\includegraphics[scale=.5]{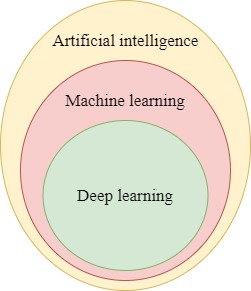}
\caption{ AI, ML and DL}
\label{fik1}
\end{figure}

Machine learning and deep learning are the subsets of artificial intelligence as shown in figure \ref{fik1}, that have witnessed rapid growth over the years. These techniques have proved to be effective in diagnosing diseases across various specialties in medicine.

\subsection{Machine learning}
Machine Learning techniques are statistical models that are used to make predictions and classifications on the given data. The machine learning models are categorized into two types based on the type of learning techniques namely supervised and unsupervised \cite{t1}. In supervised learning \cite{t2}, the machine learning model is trained with a set of input data (or features) that are associated with known output. Once the training of the machine learning model is successful, it is capable of making predictions on the new data. The predictions made by the supervised-learning algorithms can be continuous or discrete. Some of the examples of supervised learning algorithms are random forests, decision trees, logistic regression, K-nearest neighbors, and support vector machines \cite{t3, t4, t5, t6}. These techniques are used for diagnosing diseases like diabetes, thyroid, diabetic retinopathy, cardiomyopathy, etc. On the other hand, unsupervised techniques operate on input data without any labels(or outcomes). The main goal of these algorithms is to identify the undefined patterns in the unlabeled dataset. These techniques are mainly used for reducing the dimensions of the dataset by extracting important features from the data. By reducing the number of features, issues like high computational cost and multicollinearity can be avoided \cite{t7}. Some of the most commonly used unsupervised learning algorithms are latent Dirichlet analysis and principal component analysis. 
\subsection{Deep learning}
\begin{figure*}[tb] 
\centering
\includegraphics[width=\textwidth]{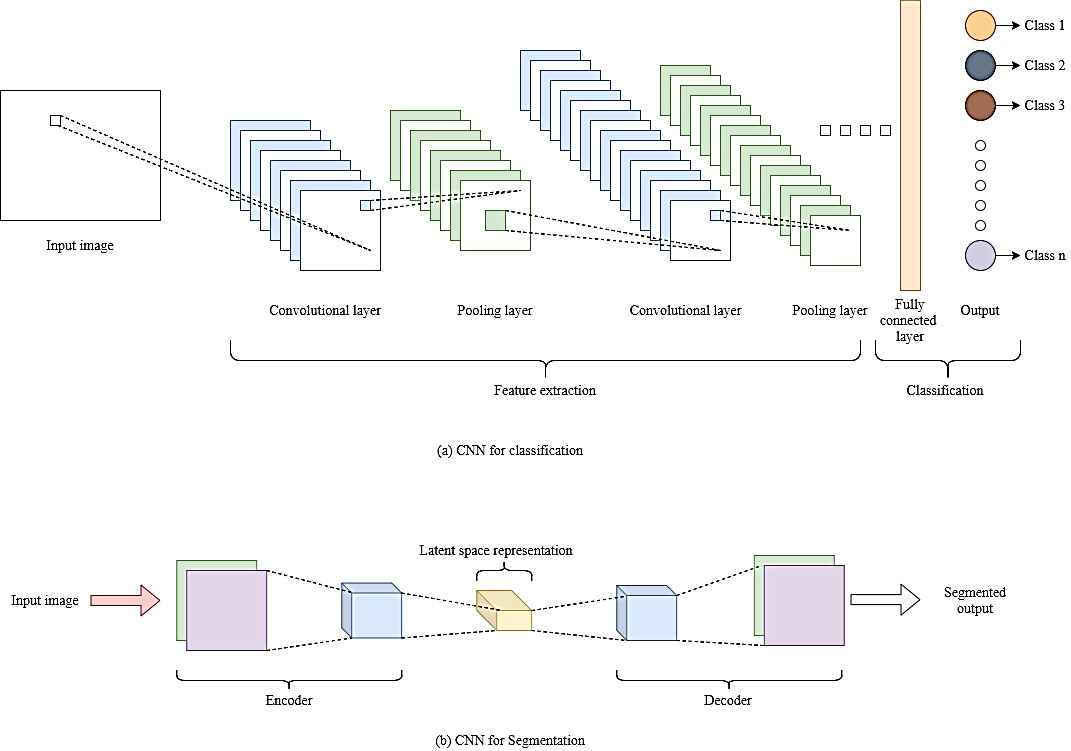}
\caption{Basic CNN architectures for classification and segmentation.}
\label{fik2}
\end{figure*}

Deep learning is a subset of the broad family of machine learning. Deep learning models are successful in solving problems related to image classification and natural language processing. These deep learning algorithms \cite{t8} are based on neural networks, these networks have multiple layers for extracting high-level features and for eliminating problematic features, so the performance of deep learning algorithms is higher than machine learning algorithms. Some of the widely used deep learning architectures \cite{t9} are Convolutional Neural Networks(CNN), Recurrent neural networks(RNN), and generative adversarial neural networks(GAN). Convolutional neural networks \cite{t10}[ are mostly used for solving computer vision and image classification tasks. CNN’s reduce the number of parameters compared to fully connected neural networks. Convolutional neural networks are used for organ segmentation, identification of tumors, classification of various cancers from radiology images. The basic structure of CNN for classification and segmentation are shown in Figure \ref{fik2}.

\begin{figure*}[tb] 
\centering
\includegraphics[width=\textwidth]{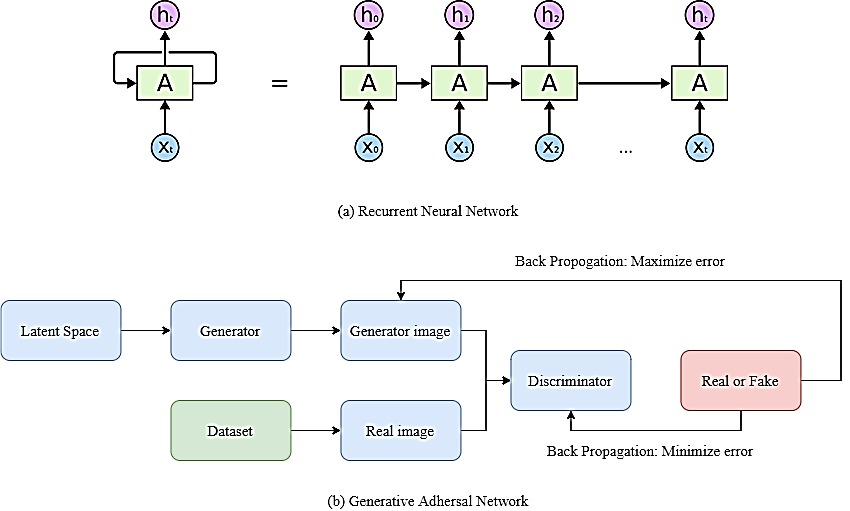}
\caption{Basic structure for Recurrent neural network and Generative adhersal network.}
\label{figfiku3}
\end{figure*}

RNN’s are used for solving audio recognition and natural language processing problems. RNN’s \cite{t11} can be used for sequential data analysis as they show temporal dynamic behavior. This property of RNN can be used for aiding fractional radiotherapy. GAN’s are generative models \cite{t12}, when given a training dataset these models are capable of generating new data that have the same statistics as the training data. GAN’s learn with an adversarial competition between its discriminator and generator. GAN’s are used mainly for medical image synthesis and for image augmentation of the training data to avoid problems like overfitting and data scarcity. The basic structures of RNN and GAN are shown in figure \ref{figfiku3}.

In this paper, we present a survey of various machine learning(ML) and deep learning(DL) models that can be used for building intelligent systems for the diagnosis of various diseases in 16 medical specialties of healthcare. Along with the 16 medical specialties we present the survey of ML and DL models in drug discovery. The 16 medical specialties that are considered for this work are Dental medicine, Haematology, Surgery, Cardiology, Pulmonology, Orthopedics, Radiology, Oncology, General medicine, Psychiatry, Endocrinology, Neurology, Dermatology, Hepatology, Nephrology, and Ophthalmology. Along with the survey, we discuss the advancements in medicine with machine learning and deep learning-based diagnostic systems and also the impact of these systems on medical professionals. The rest of the paper is organised as follows, Section 2 describes the survey methodology, Section 3 describes the literature review, Section 4 discusses the impact of these methods, Section 5 concludes the work and finally Section 6 presents the suggestions for future work.

\section{Survey methodology}
The articles that were surveyed in this paper are presented and published in high-quality conferences and journals of IEEE, Elsevier, ACM, and Springer. The terms that are used for searching these articles include machine learning, deep learning, dental medicine, hematology, surgery, cardiology, pulmonology, orthopedics, radiology, oncology, general medicine, psychiatry, endocrinology, neurology, dermatology, hepatology, nephrology, ophthalmology, and drug discovery. The articles that are considered for this survey are directly related to the topic of machine learning and deep learning applications in medicine and drug discovery. For this work, both empirical and review articles related to the above topics were considered. 

\section{Literature review}
\subsection{Dental medicine}
Table 1 shows some of the popular approaches used by the researchers to diagnose dental diseases. \cite{t13} used machine learning approaches to classify three different diseases, namely dental caries, periodical infection, and periodontitis. The authors considered a dataset of 251 radio videography (RVG) x-ray images for classification, out of which 80 images are of dental caries, 110 images are of periodical infection, and 61 images of periodontitis. Three models were built they are Convolutional neural network(CNN), transfer learning of CNN with VGG16 as a feature extractor, and CNN transfer learning model with fine-tuning.
A fuzzy rule-based model is designed by \cite{t14} to diagnose dental diseases. In this model, different methods are used for feature extraction; they are entropy, edge-value, intensity, and local binary patterns, gradient feature, red-green-blue, and patch level feature. These extracted features are used to make a database. For this work, Fuzzy c-means clustering and Mamdani’s fuzzy interface methods were used.
Machine learning algorithms are used by \cite{t15} for optimizing the dental milling process. In this paper, two models of feedforward networks were designed with Bayesian regularisation and early stopping technique. The artificial neural network structure of a feedforward network built using Bayesian regularisation has two layers, the first layer with 5 hyperbolic target units and the second layer with 30 hidden hyperbolic target units and one output unit. The structure of the second feedforward network built using the early stopping technique has the same layers as the first model, but these two networks are estimated separately using the Lenvenberg-Marquardt method and quasi-Newton methods. The performance of these models is shown Table 1.


\begin{table*}[]
\centering
\label{tab1}
\caption{ML and DL methods in Dental medicine.}
\resizebox{\textwidth}{!}{\begin{tabular}{|l|l|l|l|l|}
\hline
\textbf{Author}              & \textbf{Method}                                                                             & \textbf{Diagnosis/Application}                                                                                        & \textbf{Metrics}          & \textbf{Findings} \\ \hline
\multirow{3}{*}{\cite{t13}} & CNN                                                                                         & \multirow{3}{*}{\begin{tabular}[c]{@{}l@{}}Dental caries, periodical \\ infection, and \\ periodontitis\end{tabular}} & \multirow{3}{*}{Accuracy} & 0.7307            \\ \cline{2-2} \cline{5-5} 
                             & Transfer learning                                                                           &                                                                                                                       &                           & 0.8846            \\ \cline{2-2} \cline{5-5} 
                             & \begin{tabular}[c]{@{}l@{}}Transfer learning with fine \\ tuning\end{tabular}               &                                                                                                                       &                           & 0.8846            \\ \hline
\multirow{3}{*}{\cite{t14}} & \multirow{3}{*}{Fuzzy Interface System}                                                     & \multirow{3}{*}{\begin{tabular}[c]{@{}l@{}}Abnormalities in dental \\ x-ray\end{tabular}}                             & Mean Square Error         & 0.2445            \\ \cline{4-5} 
                             &                                                                                             &                                                                                                                       & Mean Absolute Error       & 0.1264            \\ \cline{4-5} 
                             &                                                                                             &                                                                                                                       & Accuracy                  & 90.29\%           \\ \hline
\multirow{2}{*}{\cite{t15}} & \begin{tabular}[c]{@{}l@{}}Feedforward network \\ with early stopping\end{tabular}          & \multirow{2}{*}{\begin{tabular}[c]{@{}l@{}}Optimization of dental \\ mining process\end{tabular}}                     & Normalized Mean error     & 89.95\%           \\ \cline{2-2} \cline{4-5} 
                             & \begin{tabular}[c]{@{}l@{}}Feedforward network \\ with Bayesian regularisation\end{tabular} &                                                                                                                       & Normalized Mean error     & 87.98\%           \\ \hline
\end{tabular}}
\end{table*}

\subsection{Haematology}
The diagnosis of blood-related diseases is very complicated and time-consuming. Machine learning techniques are used by Haematologists to diagnose diseases accurately in less time. \cite{t16} used machine learning classification techniques to build a model that automates the differential blood counting system. The segmentation of images is done using contour models, namely snakes and balloons. Texture and shape-based features are considered for classification. Four machine learning classifiers are used for classification; they are K-Nearest Neighbour, Linear vector quantization, Multilayer perceptron, and support vector machine. The accuracies of the classifiers are shown in table \ref{tb2}.
The application of machine learning for the segmentation and tracking of thrombus is proposed by \cite{t17}.  Adaboost and Decision tree models are used for segmentation, and the K-nearest neighbor algorithm is used for tracking. These classification algorithms are evaluated based on the Dice coefficient, and positive predictive value and tracking method are evaluated based on the Dice coefficient.
The classification of sickle cell disease using machine learning is proposed by \cite{t18}. For this work, the performances of various machine learning algorithms are analyzed, as mentioned in table 2, from which Levenberg-Marquardt and random forest algorithms produced excellent results.


\begin{table*}[]
\centering
\caption{ML and DL methods in Haematology.}
\resizebox{\textwidth}{!}{\begin{tabular}{|l|l|l|l|l|}
\hline
\textbf{Author}               & \textbf{Method}                                                                                                                                                                   & \textbf{Diagnosis/Application}                                                                               & \textbf{Metrics}         & \textbf{Findings} \\ \hline
\multirow{4}{*}{\cite{t16}}  & K-Nearest Neighbour                                                                                                                                                               & \multirow{4}{*}{\begin{tabular}[c]{@{}l@{}}Automation of differential \\ blood counting system\end{tabular}} & Accuracy                 & 81\%              \\ \cline{2-2} \cline{4-5} 
                              & Linear vector Quantization                                                                                                                                                        &                                                                                                              & Accuracy                 & 83\%              \\ \cline{2-2} \cline{4-5} 
                              & Multi Layer Perceptron                                                                                                                                                            &                                                                                                              & Accuracy                 & 90\%              \\ \cline{2-2} \cline{4-5} 
                              & Support Vector Machine                                                                                                                                                            &                                                                                                              & Accuracy                 & 91\%              \\ \hline
\multirow{5}{*}{\cite{t17}} & \multirow{2}{*}{Decision Tree}                                                                                                                                                    & \multirow{5}{*}{Thrombus}                                                                                    & Dice coefficient         & 0.89 ±0.02        \\ \cline{4-5} 
                              &                                                                                                                                                                                   &                                                                                                              & Positive predicted value & 0.94 ±0.04        \\ \cline{2-2} \cline{4-5} 
                              & \multirow{2}{*}{Ada Boost}                                                                                                                                                        &                                                                                                              & Dice Coefficient         & 0.87 ±0.02        \\ \cline{4-5} 
                              &                                                                                                                                                                                   &                                                                                                              & Positive predicted value & 0.90 ±0.05        \\ \cline{2-2} \cline{4-5} 
                              & Tracking                                                                                                                                                                          &                                                                                                              & Positive predicted value & 0.94 ±0.05        \\ \hline
{\cite{t18}} & \multirow{2}{*}{Random Oracle model}                                                                                                                                              & {Sickle cell disease}                                                                        & Accuracy                 & 53.9\%            \\ \cline{4-5} 
                              &                                                                                                                                                                                   &                                                                                                              & Area Under ROC Curve     & 0.524             \\ \cline{2-2} \cline{4-5} 
                              & {\begin{tabular}[c]{@{}l@{}}Levenberg-Marquardt \\ learning algorithm\end{tabular}}                                                                                &                                                                                                              & Accuracy                 & 99.2\%            \\ \cline{4-5} 
                              &                                                                                                                                                                                   &                                                                                                              & Area Under ROC Curve     & 0.991             \\ \cline{2-2} \cline{4-5} 
                              & {\begin{tabular}[c]{@{}l@{}}Trainable decision tree \\ Classifier\end{tabular}}                                                                                    &                                                                                                              & Accuracy                 & 92.1\%            \\ \cline{4-5} 
                              &                                                                                                                                                                                   &                                                                                                              & Area Under ROC Curve     & 0.95              \\ \cline{2-2} \cline{4-5} 
                              & {\begin{tabular}[c]{@{}l@{}}Random Forest, Decision \\ Tree Ensemble Classifier\end{tabular}}                                                                      &                                                                                                              & Accuracy                 & 91.5\%            \\ \cline{4-5} 
                              &                                                                                                                                                                                   &                                                                                                              & Area Under ROC Curve     & 0.961             \\ \cline{2-2} \cline{4-5} 
                              & {\begin{tabular}[c]{@{}l@{}}Levenberg- Marquardt \\ learning algorithm and \\ Random Forest, combined \\ using Levenberg neural \\ network\end{tabular}}           &                                                                                                              & Accuracy                 & 99.2\%            \\ \cline{4-5} 
                              &                                                                                                                                                                                   &                                                                                                              & Area Under ROC Curve     & 0.988             \\ \cline{2-2} \cline{4-5} 
                              & {\begin{tabular}[c]{@{}l@{}}Levenberg- Marquardt \\ learning neural network \\ and Random Forest, \\ combined using Fischer \\ discriminate analysis\end{tabular}} &                                                                                                              & Accuracy                 & 98.9\%            \\ \cline{4-5} 
                              &                                                                                                                                                                                   &                                                                                                              & Area Under ROC Curve     & 0.995             \\ \cline{2-2} \cline{4-5} 
                              & {Functional link neural network}                                                                                                                                   &                                                                                                              & Accuracy                 & 96.2\%            \\ \cline{4-5} 
                              &                                                                                                                                                                                   &                                                                                                              & Area Under ROC Curve     & 0.972             \\ \cline{2-2} \cline{4-5} 
                              & {Linear Combiner Network}                                                                                                                                          &                                                                                                              & Accuracy                 & 84\%              \\ \cline{4-5} 
                              &                                                                                                                                                                                   &                                                                                                              & Area Under ROC Curve     & 0.849             \\ \hline
\end{tabular}}
\end{table*}

\subsection{Surgery}

Machine learning techniques are used as a decision support system for post-operative planning and as a navigation system for many of the critical surgeries. For understanding the functioning of knee implants after the Surgery, \cite{t19} used machine learning techniques for building a statistical model. The mapping function of the proposed model in this paper is built using the pre and post-operative features, which are extracted using Principle component analysis from the clinical database. The evaluation results are shown in Table 3.
Minimal invasive surgeries are being performed by surgeons to avoid complications during the surgery and to reduce patient's recovery time. Most of the minimally invasive surgeries are operated using robots. For avoiding the tremors during the procedure while operating with robots, \cite{t20} presented a method for attenuating the tremors using machine learning. The proposed method is based on optimizing parameters using particle swarm optimization technique and building a hybrid support vector machine with an integrated kernel to attenuate tremors.
Stapedotomy is a surgical procedure performed to improve hearing, which requires the estimation of the thickness of the stapes bone, which is usually done by drilling a hole through the bone. \cite{t21} used machine learning techniques for the estimation of thickness. For the estimation, the author used a machine learning scheme known as Fuzzy Lattice Neurocomputing.
In complex trauma surgeries, the surgeons are using augmented reality to understand the relationship between anatomy, tools, and implants. Olivier \cite{t22} proposed a model to improve surgical scene understanding using augmented reality. For this work, the random forest was used to identify objects in the scene, and the pixel-wise alpha map was created using an object-specific lookup table for image fusion. The data for this model is obtained from the C- arm devices with the Kinect RGB-Depth sensor.
\cite{t23} used machine learning algorithms like K-nearest neighbor, naive Bayes, and perceptron algorithm with uneven margins to classify semi-structured and unstructured pathology reports. He highlighted that the K-nearest neighbor algorithm built with binary occurrence type word vector, stop word filter, and pruning produced high accuracy in classifying semi-structured reports. 

\begin{table*}[]
\centering
\caption{ML and DL methods in Surgery.}
\resizebox{\textwidth}{!}{\begin{tabular}{|l|l|l|l|l|}
\hline
\textbf{Author}             & \textbf{Methods}                                                                                                                                        & \textbf{Diagnosis/Application}                                                                                               & \textbf{Metrics}                                                         & \textbf{Findings} \\ \hline
            {\cite{t19}} &             {\begin{tabular}[c]{@{}l@{}}Statistical Kinematics \\ model\end{tabular}}                                                                &             {\begin{tabular}[c]{@{}l@{}}Understanding the function \\ of knee implants after the \\ surgery\end{tabular}} & \begin{tabular}[c]{@{}l@{}}mean correlation \\ coefficients\end{tabular} & 0.79              \\ \cline{4-5} 
                            &                                                                                                                                                         &                                                                                                                              & \begin{tabular}[c]{@{}l@{}}mean\\ root-mean-squared-error\end{tabular}   & 3.44 mm           \\ \hline
\cite{t23}                  & \begin{tabular}[c]{@{}l@{}}K-nearest neighbour \\ algorithm with binary \\ occurrence type word \\ vector, stop word \\ filter and pruning\end{tabular} & \begin{tabular}[c]{@{}l@{}}Classification of unstructured \\ pathology reprots\end{tabular}                                  & Accuracy                                                                 & 99.4\%            \\ \hline
\end{tabular}}
\end{table*}

\subsection{Cardiology}
Cardiovascular diseases are life-threatening diseases that require a high degree of accuracy and precision for their diagnosis. Machine learning has increased the rate of accuracy in diagnosing diseases. \cite{t24} described the use of machine learning to monitor patients having cardiovascular diseases.
The application of Deep neural networks to detect myocardial infarction(heart attack) from Electrocardiogram(ECG) signals was proposed by \cite{t25}. The neural network used for this work consists of 11 convolution layers out of which six layers are activated by Rectified Linear Unit(ReLU) activation function, and the last layer is activated with soft-max function, the pooling function used is Max-pooling. The evaluation of results is done for both noisy and noiseless beat signals in the dataset.
A novel approach was proposed by Peshala \cite{t26} to detect heart failure with preserved ejection fraction by using a statistical model along with a supervised algorithm. The data for this work is preprocessed by using the KNNimpute approach, and three principal component analysis models were built for the spatiotemporal representation of rest and exercise, and distance weighted nearest neighbor algorithm is used for the diagnosis.
A comparative analysis was performed by \cite{t27} in using machine learning network structures to diagnose diabetes and cardiovascular diseases. According to this analysis, the network built using the Levenberg-Marquardt learning algorithm(LMLA) and Naive Bayesian network achieved better accuracy compared with other proposed models considered from 2008 to 2017. 
Hypertrophic cardiomyopathy(HCM) is a chronic heart disease that leads to the thickening of the heart muscle and restricts the flow of blood. \cite{t28} proposed a methodology to detect hypertrophic cardiomyopathy from 10 seconds, 12-lead electrocardiogram(ECG) signals. For this work, features are selected using information gain criterion, and random forest, and support vector machines are used for diagnosis.
The application of machine learning techniques to diagnose heart disease in patients is proposed by \cite{t29}. The data for this work is preprocessed to make it suitable for classification. A set of five machine learning algorithms like logistic regression, random forest, gradient boosting, support vector machine, and naive Bayes are used to make classification's on the data. The accuracies of these classifiers are mentioned in table 4, out of which logistic regression has maximum accuracy.
An ensemble learning methodology was proposed by \cite{t30} to diagnose heart diseases. This methodology is based on the ensemble voting of Logistic Regression, artificial neural network, Gaussian naïve Bayes, random forest, and K-nearest neighbor algorithms. Their model achieved an accuracy of 89\begin{math}\%\end{math}, precision of 91.6\begin{math}\%\end{math}, sensitivity of 86\begin{math}\%\end{math}, and specificity of 91\begin{math}\%\end{math}.
Atherosclerosis is the primary factor for cardiovascular diseases like a heart attack. A novel methodology was proposed by \cite{t31}, which focuses on the detection of risk factors from the different stages of carotid atherosclerosis that are leading to cardiovascular diseases. This method focuses on the use of a decision tree algorithm with a neural network, logistic regression, and support vector machine for making accurate predictions.
Seismocardiographic signals are used to study cardiopulmonary interactions. \cite{t32} used k-means clustering along with a time-domain amplitude feature to cluster Seismocardiographic signals. 
Single-photon emission computed tomography(SPET) imaging is a nuclear imaging technique used to capture the functioning of the heart using gamma rays. \cite{t33} used machine learning to combine clinical information and SPET imaging data to predict major adverse cardiac events(MACE). 
The use of machine learning for rehabilitation assessment is proposed by \cite{t34}. Thermal and heart rate data is processed, and the Neural network with sigmoidal and probabilistic transfer function is used for the assessment. 


\begin{table*}[]
\centering
\caption{ML and DL methods in Cardiology.}
\resizebox{\textwidth}{!}{\begin{tabular}{|l|l|l|l|l|}
\hline
\textbf{Author}              & \textbf{Method}                                                                                                                                                                                  & \textbf{Diagnosis/Application}                                                                             & \textbf{Metric}           & \textbf{Findings}                                                                   \\ \hline
\multirow{3}{*}{\cite{t25}} & \multirow{3}{*}{Convolutional Neural Network}                                                                                                                                                    & \multirow{3}{*}{Myocardial infarction}                                                                     & Accuracy                  & \begin{tabular}[c]{@{}l@{}}noisy beats: 93.53\\ noiseless beats: 95.22\end{tabular} \\ \cline{4-5} 
                             &                                                                                                                                                                                                  &                                                                                                            & Sensitivity               & \begin{tabular}[c]{@{}l@{}}noisy beats: 93.71\\ noiseless beats: 95.49\end{tabular} \\ \cline{4-5} 
                             &                                                                                                                                                                                                  &                                                                                                            & Specificity               & \begin{tabular}[c]{@{}l@{}}noisy beats: 92.83\\ noiseless beats: 94.19\end{tabular} \\ \hline
\multirow{4}{*}{\cite{t26}} & \multirow{4}{*}{\begin{tabular}[c]{@{}l@{}}Statistical model and Distance-\\ weighted -nearest-neighbor (DWNN)\end{tabular}}                                                                     & \multirow{4}{*}{Heart failure}                                                                             & Sensitivity               & 86\%                                                                                \\ \cline{4-5} 
                             &                                                                                                                                                                                                  &                                                                                                            & Specificity               & 82\%                                                                                \\ \cline{4-5} 
                             &                                                                                                                                                                                                  &                                                                                                            & Area under ROC Curve      & 0.89                                                                                \\ \cline{4-5} 
                             &                                                                                                                                                                                                  &                                                                                                            & Accuracy                  & 85\%                                                                                \\ \hline
\multirow{4}{*}{\cite{t27}} & \multirow{2}{*}{\begin{tabular}[c]{@{}l@{}}Multilayer feedforward neural network with \\ Levenberg-Marquardt learning algorithm.\end{tabular}}                                                   & \multirow{4}{*}{\begin{tabular}[c]{@{}l@{}}Diabetes and cardiovascular \\ disease in general\end{tabular}} & \multirow{4}{*}{Accuracy} & 87.29\%(diabetes)                                                                   \\ \cline{5-5} 
                             &                                                                                                                                                                                                  &                                                                                                            &                           & \begin{tabular}[c]{@{}l@{}}89.38\%\\(Cardio Vascular Disease)\end{tabular}                                                    \\ \cline{2-2} \cline{5-5} 
                             & \multirow{2}{*}{Naïve Bayesian network}                                                                                                                                                          &                                                                                                            &                           & 99.51\%(diabetes)                                                                   \\ \cline{5-5} 
                             &                                                                                                                                                                                                  &                                                                                                            &                           & \begin{tabular}[c]{@{}l@{}}97.52\%\\(Cardio Vascular Disease)\end{tabular}                                                    \\ \hline
\multirow{4}{*}{\cite{t28}} & \multirow{4}{*}{\begin{tabular}[c]{@{}l@{}}For Random Forest and Support \\ Vector Machine\end{tabular}}                                                                                         & \multirow{4}{*}{Hypertrophic cardiomyopathy}                                                               & Precision                 & 0.84                                                                                \\ \cline{4-5} 
                             &                                                                                                                                                                                                  &                                                                                                            & Recall                    & 0.89                                                                                \\ \cline{4-5} 
                             &                                                                                                                                                                                                  &                                                                                                            & Specificity               & 0.93                                                                                \\ \cline{4-5} 
                             &                                                                                                                                                                                                  &                                                                                                            & F-Measure                 & 0.86                                                                                \\ \hline
\multirow{5}{*}{\cite{t29}} & Logistic Regression                                                                                                                                                                              & \multirow{5}{*}{Heart disease in general}                                                                  & \multirow{5}{*}{Accuracy} & 0.8651685                                                                           \\ \cline{2-2} \cline{5-5} 
                             & Random Forest                                                                                                                                                                                    &                                                                                                            &                           & 0.8089888                                                                           \\ \cline{2-2} \cline{5-5} 
                             & Naive Bayes                                                                                                                                                                                      &                                                                                                            &                           & 0.8426966                                                                           \\ \cline{2-2} \cline{5-5} 
                             & Gradient Boosting                                                                                                                                                                                &                                                                                                            &                           & 0.8426875                                                                           \\ \cline{2-2} \cline{5-5} 
                             & Support Vector Machine                                                                                                                                                                           &                                                                                                            &                           & 0.7977528                                                                           \\ \hline
\multirow{4}{*}{\cite{t30}} & \multirow{4}{*}{\begin{tabular}[c]{@{}l@{}}Ensemble Voting of Logistic regression, \\ artificial neural network, gaussian naive \\ bayes, random forest and K-nearest \\ neighbor.\end{tabular}} & \multirow{4}{*}{Heart disease in general}                                                                  & Accuracy                  & 89\%                                                                                \\ \cline{4-5} 
                             &                                                                                                                                                                                                  &                                                                                                            & Precision                 & 91.6\%                                                                              \\ \cline{4-5} 
                             &                                                                                                                                                                                                  &                                                                                                            & Sensitivity               & 86\%                                                                                \\ \cline{4-5} 
                             &                                                                                                                                                                                                  &                                                                                                            & Specificity               & 91\%                                                                                \\ \hline
\cite{t31}                  & \begin{tabular}[c]{@{}l@{}}Decision Tree algorithm with support \\ vector machine,\\ logical regression and neural networks\end{tabular}                                                         & Carotid atherosclerosis                                                                                    & Accuracy                  & 87.2\%                                                                              \\ \hline
\end{tabular}}
\end{table*}

\subsection{Pulmonology}
Respiratory diseases are very complex diseases that affect not only the lungs but also the other organs of the body. Predicting these diseases in advance can reduce the complications while treating these diseases. Many studies are performed by researchers in diagnosing respiratory diseases by using machine learning techniques. Acute diseases like Respiratory distress syndrome(ARDS) is diagnosed using machine learning by \cite{t35}. For this work, four different versions of the Support vector machine were built. The performance of these algorithms is evaluated based on accuracy and area under the receiver operating characteristic(ROC) curve shown in table 5.
Audio auscultation is the primary examination that a specialist performs when he sees a patient. Audio auscultation is considered to be the best primary practice to detect abnormalities in the heart and lungs. An approach was performed by \cite{t36} to detect respiratory diseases using audio auscultation. In this approach, the automatic free sound extractor was used to extract features from the ICBHI-2017 challenge database, and a boosted decision tree algorithm is used to detect sounds with respiratory diseases. 
A study was performed by \cite{t37} in diagnosing respiratory pathologies from pulmonary acoustic signals using support vector machine and K-nearest neighbor algorithms.
A comparative study was performed by \cite{t38} to detect obstructive sleep apnoea by using electrocardiogram derived respiratory signal(EDR) and respiratory chest signals. Three models, namely Support Vector machine, linear discriminant analysis, and extreme learning machine was built. These three models diagnosed the disease with high accuracy when they are trained with respiratory chest signals.
Seismocardiogram cycles were used to identify respiratory phases like inspiration and expiration using machine learning by \cite{t39}. For building, this model support vector machine was employed and trained with the 50 percent data approved by healthcare authorities of British Columbia.
A novel methodology named FissureNet was proposed by \cite{t40} to detect pulmonary fissure from computed tomography(CT) images. This FissureNet is a source of a fine cascade of two Seg3DNet neural network algorithms. The proposed methodology achieved higher accuracy when compared to models built with U-Net architecture and Hessian matrix.

\begin{table*}[]
\caption{ML and DL methods in Pulmonology.}
\resizebox{\textwidth}{!}{\begin{tabular}{|l|l|l|l|l|}
\hline
\textbf{Author}              & \textbf{Method}                                                                                                                                & \textbf{Disease}                                                                                        & \textbf{Metrics}                                                                    & \textbf{Findings} \\ \hline
\multirow{8}{*}{\cite{t35}} & \multirow{2}{*}{\begin{tabular}[c]{@{}l@{}}Support Vector Machine with a \\ class weighted cost function \\ Sampled Randomly\end{tabular}}     & \multirow{8}{*}{Respiratory distress syndrome}                                                          & Accuracy                                                                            & 0.7478            \\ \cline{4-5} 
                             &                                                                                                                                                &                                                                                                         & \begin{tabular}[c]{@{}l@{}}Area under \\ ROC Curve\end{tabular}                     & 0.7703            \\ \cline{2-2} \cline{4-5} 
                             & \multirow{2}{*}{\begin{tabular}[c]{@{}l@{}}Support Vector Machine with a \\ class weighted cost function \\ without any sampling\end{tabular}} &                                                                                                         & Accuracy                                                                            & 0.7094            \\ \cline{4-5} 
                             &                                                                                                                                                &                                                                                                         & \begin{tabular}[c]{@{}l@{}}Area under \\ ROC Curve\end{tabular}                     & 0.7122            \\ \cline{2-2} \cline{4-5} 
                             & \multirow{2}{*}{\begin{tabular}[c]{@{}l@{}}Support Vector Machine with \\ Label Uncertainty sampled randomly\end{tabular}}                     &                                                                                                         & Accuracy                                                                            & 0.7698            \\ \cline{4-5} 
                             &                                                                                                                                                &                                                                                                         & \begin{tabular}[c]{@{}l@{}}Area under \\ ROC Curve\end{tabular}                     & 0.7989            \\ \cline{2-2} \cline{4-5} 
                             & \multirow{2}{*}{\begin{tabular}[c]{@{}l@{}}Support Vector Machine with Label \\ Uncertainity without any sampling\end{tabular}}                &                                                                                                         & Accuracy                                                                            & 0.7188            \\ \cline{4-5} 
                             &                                                                                                                                                &                                                                                                         & \begin{tabular}[c]{@{}l@{}}Area under \\ ROC Curve\end{tabular}                     & 0.7431            \\ \hline
\cite{t36}                  & Boosted Decision Tree                                                                                                                          & \begin{tabular}[c]{@{}l@{}}Respiratory abnormalities \\ from audio auscultation\end{tabular}            & Accuracy                                                                            & 85\%              \\ \hline
\cite{t39}                  & Support Vector Machine                                                                                                                         & \begin{tabular}[c]{@{}l@{}}Respiratory phase identification\\ (inspiration and expiration)\end{tabular} & Accuracy                                                                            & 88.4\%            \\ \hline
\cite{t40}                  & \begin{tabular}[c]{@{}l@{}}FissureNet\\ (Coarse to fine cascade of two Seg3DNets)\end{tabular}                                                  & Pulmonary fissure                                                                                       & \begin{tabular}[c]{@{}l@{}}Area under \\ the precision \\ recall curve\end{tabular} & 0.991             \\ \hline
\end{tabular}}
\end{table*}

\subsection{Orthopaedics}
The risk of having diseases like osteoporosis, arthritis is predicted using machine learning. A methodology was proposed by \cite{t56} for the risk assessment of osteoporosis. In this methodology, the Support Vector machine is used for risk assessment, and this method was tested on the Korean government's Survey's dataset.
The diagnosis of rheumatoid arthritis from X-ray images was proposed by \cite{t57}. Support Vector Machine, along with a histogram of oriented gradients, is used for the diagnosis. The accuracy of the model detecting finger joints, erosion estimation, and Joint space narrowing(JSN) is shown in table 6.
The assessment of bone density from x-ray images was proposed by \cite{t58}. In the proposed methodology, the traditional GoogleNet convolutional neural network was modified by adding large number filters with a multilayer perceptron. This modified neural network produced better results than the traditional neural network.

\begin{table*}[]
\caption{ML and DL methods in Orthopaedics.}
\resizebox{\textwidth}{!}{\begin{tabular}{|l|l|l|l|l|}
\hline
\textbf{Author}              & \textbf{Method}                                                                          & \textbf{Diagnosis/Application}                   & \textbf{Metric}           & \textbf{Findings}                                                              \\ \hline
\multirow{4}{*}{\cite{t56}} & \multirow{4}{*}{Support Vector Machine}                                                  & \multirow{4}{*}{Risk assessment of osteoporosis} & Area under ROC Curve      & 0.827                                                                          \\ \cline{4-5} 
                             &                                                                                          &                                                  & Accuracy                  & 76.7\%                                                                         \\ \cline{4-5} 
                             &                                                                                          &                                                  & Sensitivity               & 77.8\%                                                                         \\ \cline{4-5} 
                             &                                                                                          &                                                  & Specificity               & 76.0\%                                                                         \\ \hline
\multirow{3}{*}{\cite{t57}} & \multirow{3}{*}{Support Vector Machine}                                                  & \multirow{3}{*}{Rheumatoid arthritis}            & \multirow{3}{*}{Accuracy} & \begin{tabular}[c]{@{}l@{}}81.4\% \\ (detection of finger joints)\end{tabular} \\ \cline{5-5} 
                             &                                                                                          &                                                  &                           & \begin{tabular}[c]{@{}l@{}}50.09 \\ (erosion estimation)\end{tabular}          \\ \cline{5-5} 
                             &                                                                                          &                                                  &                           & \begin{tabular}[c]{@{}l@{}}64.3\% \\ (JSN score)\end{tabular}                  \\ \hline
\cite{t58}                  & \begin{tabular}[c]{@{}l@{}}Modified GoogleNet \\ convolution neural network\end{tabular} & Bone density assessment                          & Accuracy                  & 91.3\%                                                                         \\ \hline
\end{tabular}}
\end{table*}

\subsection{Radiology}
Radiology deals with accurate, precise diagnosis and treatment of many diseases. Machine learning algorithms were used by \cite{t59} for gating and tracking of lung cancer tumors. In this work, Principle component analysis and neural network with backpropagation algorithm are used for gating, and an artificial neural network is used for tracking. The results are shown in Table 7.
\cite{t60} compared the performance of three convolutional neural network architectures namely GoogleNet, InceptionV3, and ResNet50 to detect abnormalities in a chest x-ray. For this work, the images from the Stanford Normal Radiology Diagnostic Dataset are used and are preprocessed using histogram equalization before training the CNN models.
Detecting fractured bones from x-ray images was proposed by \cite{t61}. For this work, many types of features were fused for building a stacked random forest model to detect the fractures of bones. This stacked random forest showed enhanced performance compared to the support vector machine. 
Machine learning classifiers and active learning methods are used by \cite{t62}, to classify radiology reports. In this work, four active learning methods like Self-Confident, Simple, Balance EE, and Kernel Farthest-First are analyzed to enhance the performance of the Support vector machine classifier for the classification of reports. The rule-based preprocessing method is used to preprocess the data.
A study was performed by \cite{t63} on using machine learning for the classification of Malignant and Benign microcalcifications. In this work, five different machine learning and deep learning algorithms, namely Support Vector Machine, Kernel Fisher Discriminant, Relevance Vector Machines, Feed Forward Neural Network, and Ada Boost, were studied and their performances were evaluated. The performances of the classifiers are described in table 7.
\cite{t64} used machine-learning to detect breast cancer from mammogram images. For this approach, a three-layered feedforward neural network with a backpropagation algorithm was built to detect breast cancer. This neural network is trained on features extracted from mammograms by experienced radiologists.
Abnormalities from orthopedic trauma radiographs are diagnosed using neural network approaches by Jakub \cite{t65}. Five freely available networks from the Caffe library are considered, and they are retrained with 13 epochs. The performance of all the networks are similar and exhibited an accuracy of 90

\begin{table*}[]
\caption{ML and DL methods in Radiology.}
\resizebox{\textwidth}{!}{\begin{tabular}{|l|l|l|l|l|}
\hline
\textbf{Author}              & \textbf{Method}                                                                                                  & \textbf{Application/Diagnosis}                              & \textbf{Metrics}                                                                 & \textbf{Findings} \\ \hline
\multirow{3}{*}{\cite{t59}} & \begin{tabular}[c]{@{}l@{}}Principal Component Analysis and \\ Artificial Neural Network. (Gatting)\end{tabular} & \multirow{3}{*}{Gating and tracking of lung cancer tumours} & Precision                                                                        & 96.5              \\ \cline{2-2} \cline{4-5} 
                             & \multirow{2}{*}{Artificial Neural Network(tracking)}                                                             &                                                             & mean localization error                                                          & 2.1 pixels        \\ \cline{4-5} 
                             &                                                                                                                  &                                                             & \begin{tabular}[c]{@{}l@{}}maximum error at 95\%\\ confidence level\end{tabular} & 4.6 pixels        \\ \hline
\multirow{2}{*}{\cite{t60}} & \multirow{2}{*}{CNN with GoogLeNet architecture}                                                                 & \multirow{2}{*}{Abnormality detection of chest x-ray}       & Accuracy                                                                         & 0.8               \\ \cline{4-5} 
                             &                                                                                                                  &                                                             & F1 Score                                                                         & 0.66              \\ \hline
cite\{t61\}                  & Stacked Random Forests Feature fusion.                                                                           & Bone fractures                                              & Precision                                                                        & 24.7\%            \\ \hline
\multirow{2}{*}{\cite{t62}} & \multirow{2}{*}{Support Vector Machine}                                                                          & \multirow{2}{*}{Classification of radiology reports}        & Sensitivity                                                                      & 98.25\%           \\ \cline{4-5} 
                             &                                                                                                                  &                                                             & Specificity                                                                      & 96.14\%           \\ \hline
\multirow{5}{*}{\cite{t63}} & Support Vector Machine                                                                                           & \multirow{5}{*}{Malignant and Benign microcalcifications}   & Standard Deviation                                                               & 0.0259            \\ \cline{2-2} \cline{4-5} 
                             & Kernel Fisher Discriminant                                                                                       &                                                             & Standard Deviation                                                               & 0.0254            \\ \cline{2-2} \cline{4-5} 
                             & Relevance Vector Machines                                                                                        &                                                             & Standard Deviation                                                               & 0.0243            \\ \cline{2-2} \cline{4-5} 
                             & Feed Forward Neural Network                                                                                      &                                                             & Standard Deviation                                                               & 0.0266            \\ \cline{2-2} \cline{4-5} 
                             & Ada Boost                                                                                                        &                                                             & Standard Deviation                                                               & 0.0268            \\ \hline
\cite{t64}                  & Artificial Neural network                                                                                        & Breast cancer                                               & Area under ROC Curve                                                             & 1                 \\ \hline
\end{tabular}}
\end{table*}


\subsection{General medicine}
Computer-aided diagnosis is used in general medicine to diagnose diseases based on the symptoms of the patient. The Survey of machine learning models in general medicine is shown in Table 9. A machine learning approach was proposed by \cite{t74} to improve the performance of computer-based diagnosis systems. For this purpose, a semi-supervised learning algorithm named co-Forest was used.
\cite{t75} proposed a model to diagnose fever. This model was build using the Naive Bayes algorithm with Dempster Shafer theory.
Dengue Hemorrhagic Fever is a fatal disease that was found mostly in the southern parts of Asia. For estimating the risk level of this disease, a machine learning model was proposed by \cite{t76}. The risk level of the disease is predicted using an extreme learning machine.
For avoiding aspiration pneumonia, a swallowing assessment model was developed using machine learning techniques by \cite{t77}. The features are extracted using linear predictive coding, and a support vector machine is used to classify patients with dysphagia.
Convolutional neural networks were used by \cite{t78} to diagnose malaria. The performance of the neural network is superior to the performance of the support vector machine.

\begin{table*}[]
\centering
\caption{ML and DL methods in General Medicine.}
\begin{tabular}{|l|l|l|l|l|}
\hline
\textbf{Author}              & \textbf{Method}                                                                                                  & \textbf{Application/Diagnosis}                                                       & \textbf{Metric}                                                             & \textbf{Findings} \\ \hline
\cite{t75}                  & \begin{tabular}[c]{@{}l@{}}Dempster Shafer with Naive \\ Bayes\end{tabular}                                      & Fever                                                                                & Accuracy                                                                    & 56.25\%           \\ \hline
\multirow{2}{*}{\cite{t76}} & \multirow{2}{*}{Extreme learning machine}                                                                        & \multirow{2}{*}{\begin{tabular}[c]{@{}l@{}}Dengue \\ Hemorrhagic Fever\end{tabular}} & Mean Absolute Error                                                         & 0.08698           \\ \cline{4-5} 
                             &                                                                                                                  &                                                                                      & \begin{tabular}[c]{@{}l@{}}Mean Absolute \\ Percentage Error\end{tabular}   & 3.00536           \\ \hline
\multirow{2}{*}{\cite{t77}} & \multirow{2}{*}{\begin{tabular}[c]{@{}l@{}}Support Vector Machine with \\ Linear predictive coding\end{tabular}} & \multirow{2}{*}{Dysphagia}                                                           & Sensitivity                                                                 & 82.4\%            \\ \cline{4-5} 
                             &                                                                                                                  &                                                                                      & Specificity                                                                 & 86\%              \\ \hline
\multirow{6}{*}{\cite{t78}} & \multirow{6}{*}{Convolutional neural network}                                                                    & \multirow{6}{*}{Malaria}                                                             & Accuracy                                                                    & 97.37\%           \\ \cline{4-5} 
                             &                                                                                                                  &                                                                                      & Sensitivity                                                                 & 96.99\%           \\ \cline{4-5} 
                             &                                                                                                                  &                                                                                      & Specificity                                                                 & 97.75\%           \\ \cline{4-5} 
                             &                                                                                                                  &                                                                                      & F1-Score                                                                    & 97.36\%           \\ \cline{4-5} 
                             &                                                                                                                  &                                                                                      & Precision                                                                   & 97.73\%           \\ \cline{4-5} 
                             &                                                                                                                  &                                                                                      & \begin{tabular}[c]{@{}l@{}}Matthews correlation \\ coefficient\end{tabular} & 94.75\%           \\ \hline
\end{tabular}
\end{table*}


\subsection{Oncology}
Diseases like cancer need accurate prediction for their diagnosis and treatment. A comparative study was performed by \cite{t66} on using machine learning to diagnose cervical cancers. Various existing machine learning models were analyzed in this study.
\cite{t67} performed an analysis of using machine learning for diagnosing cancer. Machine learning techniques like Xgboost, Deepboost, BoostI, and support vector machine are studied. The performances of these algorithms were shown in table 8.
A model was proposed by \cite{t68} to diagnose breast cancer using a genetic algorithm. This genetic algorithm obtained an accuracy of 97.2
The detection of colon cancer from CT colonography images was proposed by \cite{t69}. A histogram of oriented gradients is employed to extract features, and a fully connected neural network was used for diagnosis. Convolutional neural networks performed better than other existing algorithms for diagnosis.
Many cancer-related deaths in men are due to prostate cancer. A neural network model was developed by \cite{t70} to detect prostate cancers from multi-parametric MRI scan images. This convolutional neural network has 5 convolutional layers, and it is evaluated based on the area under the ROC curve.
Machine learning and deep learning techniques are applied to classify cancerous profiles by \cite{t71} For this work, a feedforward neural network with a backpropagation algorithm and support vector machine is used for clustering and classification of data. 
Neural network approaches were used by \cite{t72} to detect lung cancer from chest x-ray images. For this work, 121 layered neural network and transfer learning methods are employed for building the model. The performance of the model over three different datasets are shown in Table 8.
The waiting times of patients in radiational oncology were predicted using machine learning techniques by \citep{t73}. For this study, a set of four different algorithms, namely, support vector machine, linear regression, random forest, and decision tree, were analyzed. Random Forest algorithm performed best when compared to the remaining three algorithms.

\begin{table*}[]
\caption{Ml and DL methods in Oncology.}
\resizebox{\textwidth}{!}{\begin{tabular}{|l|l|l|l|l|}
\hline
\textbf{Author}              & \textbf{Method}                                                                                                & \textbf{Diagnosis/Application}                                                                            & \textbf{Metrics}                                                         & \textbf{Findings}                                                                                  \\ \hline
\multirow{8}{*}{\cite{t67}} & \multirow{2}{*}{Xgboost}                                                                                       & \multirow{8}{*}{\begin{tabular}[c]{@{}l@{}}Thyroid cancer, colon \\ cancer and liver cancer\end{tabular}} & \begin{tabular}[c]{@{}l@{}}Average mean \\ area under Curve\end{tabular} & \begin{tabular}[c]{@{}l@{}}Thyroid nodule: 0.811\\ Colon: 0.872\\ Liver: 0.797\end{tabular}        \\ \cline{4-5} 
                             &                                                                                                                &                                                                                                           & AMACC                                                                    & \begin{tabular}[c]{@{}l@{}}Thyroid nodule: 0.798\\ Colon: 0.857\\ Liver: 0.775\end{tabular}        \\ \cline{2-2} \cline{4-5} 
                             & \multirow{2}{*}{Support Vector Machine}                                                                        &                                                                                                           & \begin{tabular}[c]{@{}l@{}}Average mean \\ area under Curve\end{tabular} & \begin{tabular}[c]{@{}l@{}}Thyroid nodule: 0.750\\ Colon: 0.897\\ Liver: 0.897\end{tabular}        \\ \cline{4-5} 
                             &                                                                                                                &                                                                                                           & AMACC                                                                    & \begin{tabular}[c]{@{}l@{}}Thyroid nodule: 0.726\\ Colon: 0.890\\ Liver: 0.895\end{tabular}        \\ \cline{2-2} \cline{4-5} 
                             & \multirow{2}{*}{DeepBoost}                                                                                     &                                                                                                           & \begin{tabular}[c]{@{}l@{}}Average mean \\ area under Curve\end{tabular} & \begin{tabular}[c]{@{}l@{}}Thyroid nodule: 0.758\\ Colon: 0.822\\ Liver: 0.791\end{tabular}        \\ \cline{4-5} 
                             &                                                                                                                &                                                                                                           & AMACC                                                                    & \begin{tabular}[c]{@{}l@{}}Thyroid nodule: 0.744\\ Colon: 0.833\\ Liver: 0.786\end{tabular}        \\ \cline{2-2} \cline{4-5} 
                             & \multirow{2}{*}{Boost I}                                                                                       &                                                                                                           & \begin{tabular}[c]{@{}l@{}}Average mean \\ area under Curve\end{tabular} & \begin{tabular}[c]{@{}l@{}}Thyroid nodule: 0.500\\ Colon: 0.500\\ Liver: 0.500\end{tabular}        \\ \cline{4-5} 
                             &                                                                                                                &                                                                                                           & AMACC                                                                    & \begin{tabular}[c]{@{}l@{}}Thyroid nodule: 0.658\\ Colon: 0.694\\ Liver: 0.652\end{tabular}        \\ \hline
\cite{t68}                  & Genetic Algorithm                                                                                              & Breast cancer                                                                                             & Accuracy                                                                 & 97.2\%                                                                                             \\ \hline
\multirow{3}{*}{\cite{t69}} & \multirow{3}{*}{Convolutional Neural Network}                                                                  & \multirow{3}{*}{Colon cancer}                                                                             & Accuracy                                                                 & \begin{tabular}[c]{@{}l@{}}Colon detection: 87.03\%\\ Polyp Detection In colon: 88.56\end{tabular} \\ \cline{4-5} 
                             &                                                                                                                &                                                                                                           & Sensitivity                                                              & Polyp Detection In Colon: 88.77\%                                                                  \\ \cline{4-5} 
                             &                                                                                                                &                                                                                                           & Specificity                                                              & Polyp Detection In Colon: 87.35\%                                                                  \\ \hline
cite\{t70\}                  & Deep Convolutional Neural Network                                                                              & Prostate cancer                                                                                           & Area under ROC Curve                                                     & 0.903±0.009                                                                                        \\ \hline
\multirow{2}{*}{\cite{t71}} & \begin{tabular}[c]{@{}l@{}}Feed Forward Network and\\ (clustering)\end{tabular}                                & \multirow{2}{*}{\begin{tabular}[c]{@{}l@{}}Cancer profile \\ classification\end{tabular}}                 & Accuracy                                                                 & 87.51 (for 100 samples)                                                                            \\ \cline{2-2} \cline{4-5} 
                             & Support Vector Machine(classification)                                                                         &                                                                                                           & Accuracy                                                                 & 93.57 (for 100 samples)                                                                            \\ \hline
\multirow{3}{*}{\cite{t72}} & \multirow{3}{*}{\begin{tabular}[c]{@{}l@{}}121-layer Densely Con-\\ nected Convolutional Network\end{tabular}} & \multirow{3}{*}{Lung cancer}                                                                              & Mean Accuracy                                                            & 74.43±6.01\%                                                                                       \\ \cline{4-5} 
                             &                                                                                                                &                                                                                                           & Mean Specificity                                                         & 74.96±9.85\%                                                                                       \\ \cline{4-5} 
                             &                                                                                                                &                                                                                                           & Mean Sensitivity                                                         & 74.68±15.33\%                                                                                      \\ \hline
\multirow{4}{*}{\cite{t73}} & \multirow{4}{*}{Random Forest Regressor}                                                                       & \multirow{4}{*}{\begin{tabular}[c]{@{}l@{}}Prediction of patients \\ waiting time\end{tabular}}           & Median Absolute Error {[}min{]}                                          & 3.3                                                                                                \\ \cline{4-5} 
                             &                                                                                                                &                                                                                                           & Mean Absolute Error {[}min{]}                                            & 4.6                                                                                                \\ \cline{4-5} 
                             &                                                                                                                &                                                                                                           & Standard Deviation Error {[}min{]}                                       & 6.1                                                                                                \\ \cline{4-5} 
                             &                                                                                                                &                                                                                                           & R-squared                                                                & 0.47                                                                                               \\ \hline
\end{tabular}}
\end{table*}


\subsection{Psychiatry}
Psychiatric disorders are common health problems with which millions of people are suffering around the world. Machine learning and deep learning helps to diagnose problems like depression based on data from many sources. The models surveyed to diagnose psychiatric disorders are shown in Table 10.  
An approach was proposed by \cite{t79} to classify people with depression based on the data from social networking sites. A multilayer neural network was used for the classification of users with depression from social network data. 
Another methodology was proposed by \cite{t80} to classify depression using deep  learning. For this method, the particle swarm optimization algorithm is used for feature extraction, and Meta Cognitive neural network with projection-based learning is used for classification. 
A study was performed by \cite{t81} to diagnose depression using machine learning in diabetes mellitus type 2 patients. For this study, four machine learning algorithms, namely the K-means, Fuzzy C-mean, probabilistic neural network, and support vector machine, were built. The performance of the Support vector machine is better compared to the remaining algorithms.
For the detection of stress, a machine learning framework was proposed by \cite{t82}. Stress at different levels was diagnosed from Electrocardiogram signals using a logistic regression model.
Psychiatric disorders are found in newborn babies, which were inherited from their depressed parents. A study was performed by \cite{t83} to detect familiar depression in patients from MRI scans using machine learning and deep learning. In this study, logistic regression, graph convolutional neural network, and Support vector machine were analyzed.
Various machine learning algorithms were compared by \cite{t84} to diagnose depression and anxiety among patients. Five different classifiers, namely Logistic Regression, Random Forest, CatBoost, Naive Bayes, and Support Vector machine, were evaluated.
A comprehensive survey was performed by \cite{t85} on using machine learning to build systems that detect mental stress. In this Survey, popular feature selection methods are discussed.
A comparative study was performed by \cite{t86} on using random forest and neural networks to classify depression in patients based on motor activity. The performance of random forest is better when compared to the neural network.
A hybrid model was developed by \cite{t87} to predict the emotional state of learners from brain waves using machine learning techniques. WEKA software is used for this approach, and the emotional states were predicted using the K-nearest neighbor algorithm. 
\begin{table*}[]
\centering
\caption{ML and DL methods in Psychiatry.}
\resizebox{\textwidth}{!}{\begin{tabular}{|l|l|l|l|l|}
\hline
\textbf{Author}             & \textbf{Algorithm}                                                                                                                                                           & \textbf{Application/Diagnosis}                                                           & \textbf{Metrics}                                                            & \textbf{Findings}                                                               \\ \hline
\                  {\cite{t80}} & \                  {\begin{tabular}[c]{@{}l@{}}Meta-Cognitive Neural \\ Network using \\ Projection-based learning \\ framework and particle \\ swarm optimization\end{tabular}} & \                  {\begin{tabular}[c]{@{}l@{}}Classification of \\ depression\end{tabular}} & \                  {Efficiency}                                                 & 85.94\% (Average)                                                               \\ \cline{5-5} 
                            &                                                                                                                                                                              &                                                                                          &                                                                             & 89.4\% (Overall)                                                                \\ \cline{4-5} 
                            &                                                                                                                                                                              &                                                                                          & \                  {Sensitivity}                                                & 74.11\% (Average)                                                               \\ \cline{5-5} 
                            &                                                                                                                                                                              &                                                                                          &                                                                             & 96.55\% (Average)                                                               \\ \hline
\                  {\cite{t81}} & Support Vector Machine                                                                                                                                                       & \                  {Depression}                                                              & \                  {Accuracy}                                                   & 96.475                                                                          \\ \cline{2-2} \cline{5-5} 
                            & Fuzzy C\_MEAN                                                                                                                                                                &                                                                                          &                                                                             & 95.455                                                                          \\ \cline{2-2} \cline{5-5} 
                            & Probabilistic Neural Network                                                                                                                                                 &                                                                                          &                                                                             & 93.75                                                                           \\ \cline{2-2} \cline{5-5} 
                            & K-Means                                                                                                                                                                      &                                                                                          &                                                                             & 87.879                                                                          \\ \hline
\                  {\cite{t82}} & \                  {Logistic Regression Classifier}                                                                                                                              & \                  {Stress}                                                                  & \                  {Accuracy}                                                   & \begin{tabular}[c]{@{}l@{}}94.6\% for \\ two level identification\end{tabular}  \\ \cline{5-5} 
                            &                                                                                                                                                                              &                                                                                          &                                                                             & \begin{tabular}[c]{@{}l@{}}83.4\% for \\ multilevel identification\end{tabular} \\ \hline
\                  {\cite{t83}} & \begin{tabular}[c]{@{}l@{}}Logistic Regression with \\ regularisation\end{tabular}                                                                                           & \                  {Familiar depression}                                                     & \                  {Accuracy}                                                   & 97.78\%                                                                         \\ \cline{2-2} \cline{5-5} 
                            & Support Vector Machine                                                                                                                                                       &                                                                                          &                                                                             & 93.67\%                                                                         \\ \cline{2-2} \cline{5-5} 
                            & \begin{tabular}[c]{@{}l@{}}Graph convolution neural \\ network\end{tabular}                                                                                                  &                                                                                          &                                                                             & 89.58\%                                                                         \\ \hline
\                  {\cite{t84}} & CatBoost                                                                                                                                                                     & \                  {Depression and anxiety}                                                  & \                  {Accuracy}                                                   & 89.3\%                                                                          \\ \cline{2-2} \cline{5-5} 
                            & Logistic Regression                                                                                                                                                          &                                                                                          &                                                                             & 87.5\%                                                                          \\ \cline{2-2} \cline{5-5} 
                            & Support Vector Machine                                                                                                                                                       &                                                                                          &                                                                             & 82.1\%                                                                          \\ \cline{2-2} \cline{5-5} 
                            & Naive Bayes                                                                                                                                                                  &                                                                                          &                                                                             & 82.1\%                                                                          \\ \cline{2-2} \cline{5-5} 
                            & Random Forest                                                                                                                                                                &                                                                                          &                                                                             & 78.6\%                                                                          \\ \hline
\                  {\cite{t85}} & \                  {Random Forest}                                                                                                                                               & \                  {Stress}                                                                  & F1-Score                                                                    & 73\%                                                                            \\ \cline{4-5} 
                            &                                                                                                                                                                              &                                                                                          & \begin{tabular}[c]{@{}l@{}}Matthews correlation \\ coefficient\end{tabular} & 0.44                                                                            \\ \cline{2-2} \cline{4-5} 
                            & \                  {Neural Network}                                                                                                                                              &                                                                                          & F1-Score                                                                    & 69\%                                                                            \\ \cline{4-5} 
                            &                                                                                                                                                                              &                                                                                          & \begin{tabular}[c]{@{}l@{}}Matthews correlation \\ coefficient\end{tabular} & 0.35                                                                            \\ \hline
\                  {\cite{t86}} & \                  {K-Nearest Neighbour}                                                                                                                                         & \                  {Depression}                                                              & Precision                                                                   & 79.2\% - 84\%                                                                   \\ \cline{4-5} 
                            &                                                                                                                                                                              &                                                                                          & Recall                                                                      & 78.7\% - 82.2\%                                                                 \\ \cline{4-5} 
                            &                                                                                                                                                                              &                                                                                          & F-Measure                                                                   & 79.2\% - 83.5\%                                                                 \\ \cline{4-5} 
                            &                                                                                                                                                                              &                                                                                          & KAPPA                                                                       & 0.78                                                                            \\ \cline{4-5} 
                            &                                                                                                                                                                              &                                                                                          & Sensitivity                                                                 & 82.27\%                                                                         \\ \hline
\end{tabular}}
\end{table*}

\subsection{Endocrinology}
\subsubsection{Diabetes}
Diabetes is a major cause of many deadly diseases like cardiovascular diseases, brain strokes, and many more. The observations from surveyed papers are recorded in table 11. An approach was proposed by \cite{t88} to diagnose diabetes using machine learning. For this approach, a random forest classifier was used, and its performance was superior compared with logistic regression and support vector machine.
A study was performed by \cite{t89} to diagnose diabetes mellitus using a support vector machine. For this study, linear and non-linear support vector machines were analyzed and tested on a standard dataset from the UCI repository. 
A study was performed by \cite{t90} on using machine learning for preemptive diagnosis of diabetes mellitus. In this study, four machine learning algorithms, namely naive Bayes, k-nearest neighbor, support vector machine, and neural network, were analyzed. Artificial neural networks outperformed all the remaining three classifiers.
Using a Support vector machine with radial bias function(RBF) kernel for diagnosing diabetes mellitus was proposed by \cite{t91}. For this work, the data is sampled using a K-means clustering algorithm.
An approach was proposed by \cite{t92} to predict the rate of re-admission of diabetes mellitus patients at the hospital. C.45 algorithm is used for prediction on a standard dataset from the UCI machine learning repository.

\begin{table*}[]
\centering
\caption{ML and DL methods in Endocrinology(Diabetes).}
\resizebox{\textwidth}{!}{\begin{tabular}{|l|l|l|l|l|}
\hline
\textbf{Author}              & \textbf{Method}                         & \textbf{Diagnosis/Application}                                & \textbf{Metric}           & \textbf{Findings} \\ \hline
\cite{t88}                  & Random Forest                           & Diabetes                                                      & Accuracy                  & 84\%              \\ \hline
\cite{t89}                  & Support Vector Machine                  & Diabetes mellitus                                             & Accuracy                  & 80.20\%           \\ \hline
                  {\cite{t90}} & Artificial Neural Network               &                   {Diabetes mellitus}                            &                   {Accuracy} & 77.50\%           \\ \cline{2-2} \cline{5-5} 
                             & Support Vector Machine                  &                                                               &                           & 71.25             \\ \cline{2-2} \cline{5-5} 
                             & K-Nearest Neighbor                      &                                                               &                           & 73.75\%           \\ \cline{2-2} \cline{5-5} 
                             & Naive Bayes                             &                                                               &                           & 67.5\%            \\ \hline
                  {\cite{t91}} &                   {Support Vector Machine} &                   {Diabetes mellitus}                            & Accuracy                  & 94\%              \\ \cline{4-5} 
                             &                                         &                                                               & Sensitivity               & 93\%              \\ \cline{4-5} 
                             &                                         &                                                               & Specificity               & 94\%              \\ \hline
\cite{t92}                  & C4.5 Algorithm                          & Prediction readmission rate of diabetes patients at hospitals & Accuracy                  & 74.5\%            \\ \hline
\end{tabular}}
\end{table*}

\subsubsection{Thyroid}
A comparative study was performed by \cite{t93} to diagnose thyroid disease. In his study, he analyzed six models, namely perceptron network, a neural network with Radial based function(RBF) kernel, artificial immune recognition system, Learning Vector Quantization, and feedforward network with backpropagation algorithm. The performance of these models is shown in table 12.
Classifying the type of thyroid using machine learning was proposed by \cite{t94}. A support vector machine classifier is used to make classifications on the real-time data obtained from the UCI repository and a well-known hospital in Pakistan. 
The segmentation and classification of thyroid images from ultrasound scans were proposed by \cite{t95}. For this model, the Extreme learning machine and support vector machine is used for segmentation and classification of ultrasound images. The performance of these models is shown in Table 12.

\begin{table*}[]
\centering
\caption{ML and DL methods in Endocrinology(Thyroid).}
\resizebox{\textwidth}{!}{\begin{tabular}{|l|l|l|l|l|}
\hline
\textbf{Author}             & \textbf{Method}                                                                 & \textbf{Diagnosis/Application} & \textbf{Metrics}          & \textbf{Findings} \\ \hline
                  {\cite{t93}} & A Neural network with RBF                                                       &                   {Thyroid}       &                   {Accuracy} & 95.35\%           \\ \cline{2-2} \cline{5-5} 
                            & Perceptron                                                                      &                                &                           & 91.16\%           \\ \cline{2-2} \cline{5-5} 
                            & \begin{tabular}[c]{@{}l@{}}Artificial Immune Recognition \\ system\end{tabular} &                                &                           & 93.5\%            \\ \cline{2-2} \cline{5-5} 
                            & Back Propagation algorithm                                                      &                                &                           & 69.77\%           \\ \cline{2-2} \cline{5-5} 
                            & Multi-layer Perceptron                                                          &                                &                           & 96.74\%           \\ \cline{2-2} \cline{5-5} 
                            & Learning Vector Quantization                                                    &                                &                           & 93.50\%           \\ \hline
\cite{t94}                  & Support Vector Machine                                                          & Thyroid                        & Accuracy                  & 95.7\%            \\ \hline
                  {\cite{t95}} & Support Vector Machine                                                          &                   {Thyroid}       & Accuracy                  & 84.78             \\ \cline{2-2} \cline{4-5} 
                            & Extreme Learning Machine                                                        &                                & Accuracy                  & 93.56             \\ \hline
\end{tabular}}
\end{table*}

\subsection{Neurology}
Machine learning and deep learning methods are used in neurology to diagnose neurological disorders at an early stage. The performance of the models used to diagnose neurological disorders is shown in Table 13.
The diagnosis of chronic diseases like schizophrenia from MRI scans was proposed by \cite{t96}. For this approach, the Boruta algorithm is used for feature selection, and various machine learning algorithms were built, namely Support vector machine, Linear discriminant analysis, Naive Bayes, K-nearest neighbor, C5.0 Decision tree, Gaussian process classifier, and random forest.
A method was proposed by \cite{t97} to classify normal and abnormal brain Computed Tomography(CT) scan images. The features for this approach were extracted based on greyscale, shape and texture, symmetric features. See5 and Neural network with radial bias function was built based on these features.
Cerebral microbleeds are chronic brain hemorrhages that lead to the death of a person. The diagnosis of cerebral microbleeds from Susceptibility weighted imaging(SWI) scans using deep learning was proposed by \cite{t98}. Eight layered Convolutional neural network model was built and trained with 8000 images for diagnosis.
An approach was proposed by \cite{t99} to recognize and anticipate the seriousness of tremor and dyskinesia using machine learning and deep learning methods. For this work three models namely the hidden Markov model, support vector machines, and neural networks, were built for the detection of tremors and dyskinesia. The seriousness of tremor and dyskinesia was anticipated using the Bayesian maximum likelihood classifier.
A diagnostic approach for dementia and mild cognitive impairment using machine learning was proposed by Daniel \cite{t100}. For this approach three models namely extreme gradient boosting, stochastic gradient boosting, and the random forest was built using the features extracted by the RelieF method.
A two-step approach for classifying Parkinson's disease gait was proposed by \cite{t101}. In this approach, the first step is to normalize the data using a multi regression approach, and the second step is classification using random forest. The performance of this approach is superior to the performance of the support vector machine and kernel Fisher discriminant.
The detection of epileptic convulsions from accelerometric signals using machine learning was proposed by \cite{t102}. For this work, the classification of seizure and non-seizure epochs is done using the least-square support vector machine. The data for this work is collected from the patients using four accelerometers tied to their wrists and ankles. 
\cite{t103} proposed a neural network model for the assessment of Parkinson’s disease. For this work convolutional neural network with a rectified linear unit(ReLU) activation function was built, and it showed better performance compared to K-nearest neighbor, support vector machine, AdaBoost, and partial decision tree.
Bradykinesia is an important factor in the diagnosis of Parkinson's disease. A method was proposed by \cite{t104} to predict the scores of each patient based on their moments using a support vector machine. Movements like finger tapping, diadochokinesis, and toe-tapping were considered.
Classifying neurological gait disorders like Parkinson's disease and stroke using multi-task feature learning was proposed by \cite{t105}. For building this multi-task feature learning method, a support vector machine and neural network were used. 
The segmentation of glioblastomas from brain MRI scans was proposed by \cite{t106}. The proposed method focuses on replacing manual segmentation of brain MRI scans with automated segmentation using machine learning. In this approach, segmentation is done using K-means clustering, and the segmentation accuracy of this model is better than the accuracy of manual segmentation.
The detection of chronic traumatic Encephalopathy using machine learning was proposed by \cite{t107}. For this work, Random classifier, Random Forest, Support vector machine(RBF kernel), and K-nearest neighbor, were built. These models were trained with data from two sequences, namely PRESS and L-COSY of magnetic resonance spectroscopy. The accuracy of the classifiers is better when they are trained with L-COSY.


\begin{table*}[]
\centering
\caption{ML and DL methods in Neurology.}
\resizebox{\textwidth}{!}{\begin{tabular}{|l|l|l|l|l|}
\hline
\textbf{Author}               & \textbf{Method}                                                                                              & \textbf{Diagnosis/Application}                                                                        & \textbf{Metrics}                                                & \textbf{Findings}                                                                       \\ \hline
                  {\cite{t96}}  & Support vector machine                                                                                       &                   {Schizophrenia}                                                                        &                   {Accuracy}                                       & 94.12\%                                                                                 \\ \cline{2-2} \cline{5-5} 
                              & C4.5 Decision tree                                                                                           &                                                                                                       &                                                                 & 91.18\%                                                                                 \\ \cline{2-2} \cline{5-5} 
                              & Random Forest                                                                                                &                                                                                                       &                                                                 & 91.18\%                                                                                 \\ \cline{2-2} \cline{5-5} 
                              & K-Nearest Neighbours                                                                                         &                                                                                                       &                                                                 & 79.41\%                                                                                 \\ \cline{2-2} \cline{5-5} 
                              & Linear discriminant analysis                                                                                 &                                                                                                       &                                                                 & 79.41\%                                                                                 \\ \cline{2-2} \cline{5-5} 
                              & \begin{tabular}[c]{@{}l@{}}Gaussian process classifier\\ (laplacedot)\end{tabular}                           &                                                                                                       &                                                                 & 92\%                                                                                    \\ \cline{2-2} \cline{5-5} 
                              & Naive Bayes                                                                                                  &                                                                                                       &                                                                 & 85.29\%                                                                                 \\ \hline
                  {\cite{t97}}  & See5                                                                                                         &                   {Brain abnormalities}                                                                  &                   {Accuracy}                                       & 90\%-94\%                                                                               \\ \cline{2-2} \cline{5-5} 
                              & \begin{tabular}[c]{@{}l@{}}Radial Bayes Function Neural \\ Network\end{tabular}                              &                                                                                                       &                                                                 & 77\%-82\%                                                                               \\ \hline
\cite{t98}                   & Convolutional neural network                                                                                 & Cerebral microbleeds                                                                                  & Sensitivity                                                     & 97.29\%                                                                                 \\ \hline
                  {\cite{t99}}  & Neural Network                                                                                               &                   {\begin{tabular}[c]{@{}l@{}}Detection of tremor \\ and dyskinesia\end{tabular}}        &                   {Global Error Rate}                              & \begin{tabular}[c]{@{}l@{}}Tremor: 6.2\%\\ Dyskinesia: 8.8\%\end{tabular}               \\ \cline{2-2} \cline{5-5} 
                              & Support Vector Machine                                                                                       &                                                                                                       &                                                                 & \begin{tabular}[c]{@{}l@{}}Tremor: 7.2\%\\ Dyskinesia: 9.1\%\end{tabular}               \\ \cline{2-2} \cline{5-5} 
                              & Hidden Markov models                                                                                         &                                                                                                       &                                                                 & \begin{tabular}[c]{@{}l@{}}Tremor: 6.1\%\\ Dyskinesia: 12.3\%\end{tabular}              \\ \cline{2-5} 
                              & \begin{tabular}[c]{@{}l@{}}Bayesian maximum likelihood \\ classifier\end{tabular}                            & \begin{tabular}[c]{@{}l@{}}Seriousness of tremor \\ and dyskinesia\end{tabular}                       & Sensitivity                                                     & \begin{tabular}[c]{@{}l@{}}Tremor: 96.3\%\\ Dyskinesia: 99.3\%\end{tabular}             \\ \hline



\multirow{3}{*}{\cite{t100}} & \begin{tabular}[c]{@{}l@{}}Extreme Gradient\\ Boosting\end{tabular}      & \multirow{3}{*}{\begin{tabular}[c]{@{}l@{}}Dementia and mild \\ cognitive impairment\end{tabular}} & \multirow{3}{*}{Accuracy} & 0.88 \\ \cline{2-2} \cline{5-5} 
                              & \begin{tabular}[c]{@{}l@{}}Stochastic\\ Gradient\\ Boosting\end{tabular} &                                                                                                    &                           & 0.87 \\ \cline{2-2} \cline{5-5} 
                              & Random Forest                                                            &                                                                                                    &                           & 0.87 \\ \hline

                  {\cite{t101}} & kernel Fisher discriminant                                                                                   &                   {Parkinson’s disease}                                                                  &                   {Accuracy}                                       & 86.2\%                                                                                  \\ \cline{2-2} \cline{5-5} 
                              & Support Vector Machine                                                                                       &                                                                                                       &                                                                 & 80.4\%                                                                                  \\ \cline{2-2} \cline{5-5} 
                              & Random Forest                                                                                                &                                                                                                       &                                                                 & 92.6\%                                                                                  \\ \hline
                  {\cite{t102}} &                   {\begin{tabular}[c]{@{}l@{}}Least-squares support vector ma-\\ chine classifier\end{tabular}} &                   {Epileptic convulsions}                                                                & Median sensitivity                                              & \begin{tabular}[c]{@{}l@{}}100\% \\ (For seizures \\longer than\\ 30 seconds)\end{tabular}   \\ \cline{4-5} 
                              &                                                                                                              &                                                                                                       & \begin{tabular}[c]{@{}l@{}}False detection\\ rate\end{tabular}  & \begin{tabular}[c]{@{}l@{}}0.39 h−1 \\(For seizures \\longer than\\ 30 seconds\end{tabular} \\ \hline
\cite{t103}                  & Convolutional Neural Network                                                                                 & Parkinson’s disease                                                                                   & Accuracy                                                        & 90.9                                                                                    \\ \hline
                  {\cite{t104}} &                   {Support Vector Machine}                                                                      &                   {Parkinson’s disease}                                                                  &                   {Classification Errors}                          & \begin{tabular}[c]{@{}l@{}}15-16.5\%\\ (finger tapping)  \end{tabular}                                                            \\ \cline{5-5} 
                              &                                                                                                              &                                                                                                       &                                                                 & \begin{tabular}[c]{@{}l@{}}9.3-9.8\%\\ (diadochokinesis)\end{tabular}                                                             \\ \cline{5-5} 
                              &                                                                                                              &                                                                                                       &                                                                 & \begin{tabular}[c]{@{}l@{}}18.2-20.2\%\\(toe tapping)\end{tabular}                                                                \\ \hline
\cite{t105}                  & Multi Tasking Feature Learning                                                                               & Parkinson’s disease                                                                                   & \begin{tabular}[c]{@{}l@{}}Area under \\ ROC Curve\end{tabular} & 0.96                                                                                    \\ \hline
\cite{t106}                  & K-Means Clustering                                                                                           & Glioblastomas                                                                                         & Dice Coefficiet                                                 & 0.82321                                                                                 \\ \hline
                  {\cite{t107}} & Random Classifier(L-COSY)                                                                                    &                   {\begin{tabular}[c]{@{}l@{}}Chronic traumatic \\ Encephalopathy\end{tabular}}          &                   {Accuracy}                                       & 61\%                                                                                    \\ \cline{2-2} \cline{5-5} 
                              & \begin{tabular}[c]{@{}l@{}}Support Vector \\ Machine-RBF(L-COSY)\end{tabular}                                &                                                                                                       &                                                                 & 38\%                                                                                    \\ \cline{2-2} \cline{5-5} 
                              & Random Forest(L-COSY)                                                                                        &                                                                                                       &                                                                 & 61\%                                                                                    \\ \cline{2-2} \cline{5-5} 
                              & K-Nearest Neighbours(L-COSY)                                                                                 &                                                                                                       &                                                                 & 87\%                                                                                    \\ \cline{2-2} \cline{5-5} 
                              & Random Classifier(PRESS)                                                                                     &                                                                                                       &                                                                 & 45\%                                                                                    \\ \cline{2-2} \cline{5-5} 
                              & \begin{tabular}[c]{@{}l@{}}Support Vector \\ Machine-RBF(PRESS)\end{tabular}                                 &                                                                                                       &                                                                 & 60\%                                                                                    \\ \cline{2-2} \cline{5-5} 
                              & Random Forest(PRESS)                                                                                         &                                                                                                       &                                                                 & 70\%                                                                                    \\ \cline{2-2} \cline{5-5} 
                              & K-Nearest Neighbours(PRESS)                                                                                  &                                                                                                       &                                                                 & 75\%                                                                                    \\ \hline

\end{tabular}}

\end{table*}

\subsection{Dermatology}
The diagnosis of dermatological diseases using machine learning and deep learning was proposed by \cite{t108}. Three machine learning models, namely neural networks, decision tree, and an nth nearest neighbor were built and trained using a standard dataset.
An approach was performed by \cite{t109} to diagnose skin diseases from colored images using artificial neural networks. Various methods, like image cropping and color gradient generation, are applied to preprocess the data. A feedforward neural network with a backpropagation algorithm was used for diagnosis.
The segmentation of skin ulcer images using machine learning was proposed by \cite{t110}. In this study, several models like Extreme learning machine, J48, Support vector machine, K-nearest neighbor, Random Forest, Multilayer perceptron, and Naive Bayes were analyzed. The performance of these algorithms in the segmentation of ulcers is shown in table 14.
A survey was performed by \cite{t111} on using machine learning algorithms for classifying skin diseases based on color and texture features. For this survey, skin diseases, namely Eczema, Lichen plants, and plague psoriasis, were diagnosed using linear discriminant analysis, artificial neural networks, naive Bayes, and support vector machine.
A novel method was proposed by \cite{t112} to diagnose erythematosquamous diseases using ensemble modeling. Support vector machine, multilayer perceptron, and k-nearest neighbor algorithms were built using the features extracted based on rough set theory. The results of these algorithms were combined using the majority voting method(Ensemble method).
The diagnosis of diseases, namely melanocytes nevus, seborrheic keratosis, basal cell carcinoma, and psoriasis using neural network approaches were proposed by \cite{t113}. In this approach, transfer learning of GoogleNet InceptionV3 is used for diagnosis. For this approach, two different datasets with the same diseases and features were used to train the model.
The early diagnosis of skin tumors is possible by estimating the constituent elements of the skin. An approach was proposed by \cite{t114} to estimate the skin elements using machine learning. The support vector regressor model was used for estimating skin constituents. 

\begin{table*}[]
\centering
\caption{ML and DL methods in Dermatology.}
\resizebox{\textwidth}{!}{\begin{tabular}{|l|l|l|l|l|}
\hline
\textbf{Author}              & \textbf{Method}                                                                                                                           & \textbf{Application/Diagnosis}                                                                                                                                                       & \textbf{Metrics}          & \textbf{Findings}                                                                                                                             \\ \hline
{\cite{t108}} & Decision Tree                                                                                                                             & {\begin{tabular}[c]{@{}l@{}}Psoriasis, seborrheic \\ dermatitis, lichen planus, \\ pityriasis rosea, chronic \\ dermatitis, pityriasis rubra \\ pilaris\end{tabular}} & {Accuracy} & 95\%                                                                                                                                          \\ \cline{2-2} \cline{5-5} 
                             & Neural Networks                                                                                                                           &                                                                                                                                                                                      &                           & 95\%                                                                                                                                          \\ \cline{2-2} \cline{5-5} 
                             & Kth Nearest Neighbour                                                                                                                     &                                                                                                                                                                                      &                           & 95\%                                                                                                                                          \\ \hline
{\cite{t109}} & {\begin{tabular}[c]{@{}l@{}}feed-forward\\ back-propagation ANN\end{tabular}}                                              & {\begin{tabular}[c]{@{}l@{}}Acne, eczema, psoriasis, \\ tinea corporis, scabies and \\ vitiligo\end{tabular}}                                                         & {Accuracy} & 95.99\% for diseases skin detection                                                                                                           \\ \cline{5-5} 
                             &                                                                                                                                           &                                                                                                                                                                                      &                           & 94.016\% for disease identification                                                                                                           \\ \hline
{\cite{t110}} & Naive Bayes                                                                                                                               & {Skin ulcer}                                                                                                                                                          & {Accuracy} & \begin{tabular}[c]{@{}l@{}}full regions recognition: 0.64\\ Border Region Recognition: 0.67 \\ Center Region Recognition: 0.65\end{tabular}   \\ \cline{2-2} \cline{5-5} 
                             & K-Nearest Neighbour                                                                                                                       &                                                                                                                                                                                      &                           & \begin{tabular}[c]{@{}l@{}}full regions recognition: 0.93\\ Border regions recognition: 0.71\\ Center Region Recognition: 0.95\end{tabular}   \\ \cline{2-2} \cline{5-5} 
                             & ELM                                                                                                                                       &                                                                                                                                                                                      &                           & \begin{tabular}[c]{@{}l@{}}full regions recognition: 0.87\\ Border regions recognition: 0.67\\ Center Region Recognition: 0.88\end{tabular}   \\ \cline{2-2} \cline{5-5} 
                             & Support Vector Machine                                                                                                                    &                                                                                                                                                                                      &                           & \begin{tabular}[c]{@{}l@{}}full regions recognition: 0.94 \\ Border regions recognition: 0.71 \\ Center Region Recognition: 0.95\end{tabular} \\ \cline{2-2} \cline{5-5} 
                             & J48                                                                                                                                       &                                                                                                                                                                                      &                           & \begin{tabular}[c]{@{}l@{}}full regions recognition: 0.92 \\ Border regions recognition: 0.69\\ Center Region Recognition: 0.93\end{tabular}  \\ \cline{2-2} \cline{5-5} 
                             & Naive Bayes                                                                                                                               &                                                                                                                                                                                      &                           & \begin{tabular}[c]{@{}l@{}}full regions recognition: 0.64\\ Border Region Recognition: 0.67 \\ Center Region Recognition: 0.65\end{tabular}   \\ \cline{2-2} \cline{5-5} 
                             & Random Forest                                                                                                                             &                                                                                                                                                                                      &                           & \begin{tabular}[c]{@{}l@{}}full regions recognition: 0.93\\ Border regions recognition: 0.71\\ Center Region Recognition: 0.95\end{tabular}   \\ \cline{2-2} \cline{5-5} 
                             & Multi-Layer Perceptron                                                                                                                    &                                                                                                                                                                                      &                           & \begin{tabular}[c]{@{}l@{}}full regions recognition: 0.94\\ Border regions recognition: 0.70\\ Center Region Recognition: 0.95\end{tabular}   \\ \hline
{\cite{t111}} & {Artificial Neural Network}                                                                                                & {\begin{tabular}[c]{@{}l@{}}Eczema, Lichen plants, \\ and plague psoriasis\end{tabular}}                                                                              & Accuracy                  & 62.9                                                                                                                                          \\ \cline{4-5} 
                             &                                                                                                                                           &                                                                                                                                                                                      & Standard Deviation        & 33.11                                                                                                                                         \\ \cline{2-2} \cline{4-5} 
                             & {Linear Discriminant Analysis}                                                                                             &                                                                                                                                                                                      & Accuracy                  & 80.9                                                                                                                                          \\ \cline{4-5} 
                             &                                                                                                                                           &                                                                                                                                                                                      & Standard Deviation        & 23.62                                                                                                                                         \\ \cline{2-2} \cline{4-5} 
                             & {Naive Bayes}                                                                                                              &                                                                                                                                                                                      & Accuracy                  & 67.42                                                                                                                                         \\ \cline{4-5} 
                             &                                                                                                                                           &                                                                                                                                                                                      & Standard Deviation        & 21.84                                                                                                                                         \\ \cline{2-2} \cline{4-5} 
                             & {Support Vector Machine}                                                                                                   &                                                                                                                                                                                      & Accuracy                  & 81.61                                                                                                                                         \\ \cline{4-5} 
                             &                                                                                                                                           &                                                                                                                                                                                      & Standard Deviation        & 22.87                                                                                                                                         \\ \hline
\cite{t112}                  & \begin{tabular}[c]{@{}l@{}}Majority voting of Multilayer \\ perceptron, K-Nearest \\ Neighbour and Support \\ Vector Machine\end{tabular} & Erythematosquamous                                                                                                                                                                   & Accuracy                  & 97.78\%                                                                                                                                       \\ \hline
{\cite{t113}} & {\begin{tabular}[c]{@{}l@{}}GoogleNet Inception v3 \\ with transfer learning\end{tabular}}                                 & {\begin{tabular}[c]{@{}l@{}}Melanocytes nevus, \\ seborrheic keratosis, basal \\ cell carcinoma, and psoriasis\end{tabular}}                                          & Average accuracy          & 86.54\%                                                                                                                             \\ \cline{4-5} 
                             &                                                                                                                                           &                                                                                                                                                                                      & Standard deviation        & 3.63\%                                                                                                                            \\ \hline
\end{tabular}}
\end{table*}

\subsection{Hepatology}
A comparative study of using machine learning techniques to diagnose the advanced stage of liver fibrosis in patients suffering from Hepatitis C was performed by \cite{t115}. In this study, five machine learning algorithms were analyzed, namely alternating decision tree, particle swarm optimization, multi-linear regression, alternating decision tree with criterion point zero, and genetic algorithm.
An approach was proposed by \cite{t116} to diagnose liver fibrosis from liver biopsies using machine learning and search algorithms. In this approach, Related Objects, Skipper algorithm is used to query the database, and a rule-based subsumption algorithm is used for diagnosis.
Predicting the stage of liver fibrosis is essential for the treatment of the disease. 
\cite{t117} proposed a methodology to predict the stage of liver fibrosis using machine learning. In this methodology, liver fibrosis is classified into five stages using a decision tree algorithm. 
A methodology was proposed by \cite{t118} for predicting the stage of fibrosis and inflammatory activity from serum indices data of Hepatitis C patients using the Extreme learning machine algorithm.

\begin{table*}[]
\centering
\caption{ML and DL methods in Hepatology.}
\resizebox{\textwidth}{!}{\begin{tabular}{|l|l|l|l|l|}
\hline
\textbf{Author}                & \textbf{Method}                                                                                                            & \textbf{Diagnosis/Application}                                                                       & \textbf{Metric}           & \textbf{Findings}              \\ \hline
{\cite{t115}} & {Multi-linear regression}                                                                                   & {Liver fibrosis}                                                                     & Accuracy                  & 69.1\%                         \\ \cline{4-5} 
                               &                                                                                                                            &                                                                                                      & Sensitivity               & 69.0\%                         \\ \cline{4-5} 
                               &                                                                                                                            &                                                                                                      & Specificity               & 69.1\%                         \\ \cline{2-2} \cline{4-5} 
                               & {Genetic Algorithms}                                                                                        &                                                                                                      & Accuracy                  & 69.6\%                         \\ \cline{4-5} 
                               &                                                                                                                            &                                                                                                      & Sensitivity               & 68.9\%                         \\ \cline{4-5} 
                               &                                                                                                                            &                                                                                                      & Specificity               & 69.7\%                         \\ \cline{2-2} \cline{4-5} 
                               & {Alternating decision tree}                                                                                 &                                                                                                      & Accuracy                  & 66.3\%                         \\ \cline{4-5} 
                               &                                                                                                                            &                                                                                                      & Sensitivity               & 73.0\%                         \\ \cline{4-5} 
                               &                                                                                                                            &                                                                                                      & Specificity               & 65.0\%                         \\ \cline{2-2} \cline{4-5} 
                               & {Particle Swarm Optimization}                                                                               &                                                                                                      & Accuracy                  & 66.4\%                         \\ \cline{4-5} 
                               &                                                                                                                            &                                                                                                      & Sensitivity               & 70.4\%                         \\ \cline{4-5} 
                               &                                                                                                                            &                                                                                                      & Specificity               & 65.6\%                         \\ \cline{2-2} \cline{4-5} 
                               & {\begin{tabular}[c]{@{}l@{}}Alternating Decision Tree \\ model with criteria point \\ of zero\end{tabular}} &                                                                                                      & Accuracy                  & 84.4\%                         \\ \cline{4-5} 
                               &                                                                                                                            &                                                                                                      & Sensitivity               & 70\%                           \\ \cline{4-5} 
                               &                                                                                                                            &                                                                                                      & Specificity               & 99\%                           \\ \hline
\cite{t116}                   & \begin{tabular}[c]{@{}l@{}}Related Objects Skipper \\ algorithm with \\ Subsumptions Rule-Based \\ Classifier\end{tabular} & Liver fibrosis                                                                                       & Accuracy                  & 99.48\%                        \\ \hline
\cite{t117}                   & Decision Tree Classifier                                                                                                   & Liver fibrosis                                                                                       & Accuracy                  & 93.7\%                         \\ \hline
{\cite{t118}}  & {Extreme Learning machine}                                                                                  & {\begin{tabular}[c]{@{}l@{}}Liver fibrosis and \\ inflammatory activity\end{tabular}} & {Accuracy} & {\begin{tabular}[c]{@{}l@{}}69.11\%\\ (fibrosis stage)\end{tabular}}        \\ \cline{5-5} 
                               &                                                                                                                            &                                                                                                      &                           & {\begin{tabular}[c]{@{}l@{}}69.92\%\\ (inflammatory activity)\end{tabular}} \\ \hline
\end{tabular}}
\end{table*}

\subsection{Nephrology}
The prediction of chronic kidney disease using machine learning was proposed by \cite{t119}. For this work, four machine learning algorithms, namely support vector machine, decision tree, k-nearest neighbor, and logistic regression, were explored to find the best classifier. The support vector machine was found to be the best classifier.
A methodology for the diagnosis of chronic kidney disease using a support vector machine was proposed by \cite{120}. For this work, the features are selected based on the standard deviation of data.
A model was proposed by \cite{t121} to suggest the best diet plan for chronic kidney disease patients using machine learning. The best diet plan is recommended by this system by predicting the potassium zone of the patient. This experiment is carried out by various multi-class classifiers, namely decision tree, decision jungle, logistic regression, and neural network. Better performance was found in the decision tree. The surveyed models are tabulated in Table 16. 

\begin{table*}[]
\centering
\caption{ML and DL methods in Nephrology.}
\resizebox{\textwidth}{!}{\begin{tabular}{|l|l|l|l|l|}
\hline
\textbf{Author}                & \textbf{Method}                                                          & \textbf{Diagnosis/Application}                        & \textbf{Metric}           & \textbf{Findings} \\ \hline
{\cite{t119}} & {Support Vector Machine}                                  & {Chronic kidney disease}              & Accuracy                  & 98.3\%            \\ \cline{4-5} 
                               &                                                                          &                                                       & Sensitivity               & 99\%              \\ \cline{4-5} 
                               &                                                                          &                                                       & Specificity               & 98\%              \\ \cline{2-2} \cline{4-5} 
                               & {Logistic Regression}                                     &                                                       & Accuracy                  & 96.55\%           \\ \cline{4-5} 
                               &                                                                          &                                                       & Sensitivity               & 94\%              \\ \cline{4-5} 
                               &                                                                          &                                                       & Specificity               & 98\%              \\ \cline{2-2} \cline{4-5} 
                               & {Decision Tree}                                           &                                                       & Accuracy                  & 94.8\%            \\ \cline{4-5} 
                               &                                                                          &                                                       & Sensitivity               & 93\%              \\ \cline{4-5} 
                               &                                                                          &                                                       & Specificity               & 96\%              \\ \cline{2-2} \cline{4-5} 
                               & {K-Nearest Neighbour}                                     &                                                       & Accuracy                  & 98.1\%            \\ \cline{4-5} 
                               &                                                                          &                                                       & Sensitivity               & 96\%              \\ \cline{4-5} 
                               &                                                                          &                                                       & Specificity               & 99\%              \\ \hline
\cite{t120}                   & Support Vector Machine                                                   & Chronic kidney disease                                & Accuracy                  & 82\%              \\ \hline
{\cite{t121}}  & \begin{tabular}[c]{@{}l@{}}Multiclass Logistic\\ Regression\end{tabular} & {Diet plan for chronic kidney disease} & {Accuracy} & 89.17\%           \\ \cline{2-2} \cline{5-5} 
                               & \begin{tabular}[c]{@{}l@{}}Multiclass Neural\\ Network\end{tabular}      &                                                       &                           & 82.50\%           \\ \cline{2-2} \cline{5-5} 
                               & Multiclass Decision Forest                                               &                                                       &                           & 99.17\%           \\ \cline{2-2} \cline{5-5} 
                               & \begin{tabular}[c]{@{}l@{}}Multiclass Decision\\ Jungle\end{tabular}     &                                                       &                           & 97.50\%           \\ \hline
\end{tabular}}
\end{table*}

\subsection{Ophthalmology}
\cite{t122} and \cite{t123} performed a study on using a convolutional network to detect ophthalmological diseases like cataract, glaucoma, and retinal disorders. 
Various models, namely decision tree, neural network, Naive Bayes, and random forest, were analyzed by \cite{t124} for detecting eye diseases. The performances of these algorithms are shown in table 18. From these algorithms, random forest and decision tree showed better performance.  
Vision impairing diseases like diabetic eye disease are major problems in today's world. The diagnosis of diabetic eye disease from thermographic images using machine learning is proposed by \cite{t125}. For this work, texture-based features are used by the support vector machine classifier for diagnosis. 
Exudates and hemorrhages are the two primary causes of diabetic retinopathy. The diagnosis of diabetic retinopathy from retinal scans was proposed by \cite{t126}. For this approach, feature extraction and detection of hard exudates and hemorrhages from retinal images are done using principal component analysis and support vector machine.
\cite{t127} proposed a method to detect Diabetic retinopathy and glaucoma from retinal images using a neural network framework. This work focuses on changing parametric values of the VGG 19 neural network for an accurate diagnosis.

\begin{table*}[]
\centering
\caption{ML and DL methods in Ophthalmology.}
\begin{tabular}{|l|l|l|l|l|}
\hline
\textbf{Author}               & \textbf{Method}                         & \textbf{Diagnosis/Application}                                                                                   & \textbf{Metrics}          & \textbf{Findings} \\ \hline
\multirow{4}{*}{\cite{t124}} & Decision Tree                           & \multirow{4}{*}{\begin{tabular}[c]{@{}l@{}}Glaucoma, Unspecified primary \\ angle-closure glaucoma\end{tabular}} & \multirow{4}{*}{Accuracy} & 85.81\%           \\ \cline{2-2} \cline{5-5} 
                              & Random Forest                           &                                                                                                                  &                           & 86.63\%           \\ \cline{2-2} \cline{5-5} 
                              & Naive Bayes                             &                                                                                                                  &                           & 81.53\%           \\ \cline{2-2} \cline{5-5} 
                              & Neural Network                          &                                                                                                                  &                           & 85.98\%           \\ \hline
\multirow{3}{*}{\cite{t125}} & \multirow{3}{*}{Support vector machine} & \multirow{3}{*}{Diabetic eye disease}                                                                            & Accuracy                  & 86.22\%           \\ \cline{4-5} 
                              &                                         &                                                                                                                  & Sensitivity               & 94.07\%           \\ \cline{4-5} 
                              &                                         &                                                                                                                  & Specificity               & 79.17\%           \\ \hline
\multirow{2}{*}{\cite{t126}} & \multirow{2}{*}{Support Vector machine} & \multirow{2}{*}{Diabetic retinopathy}                                                                            & Accuracy                  & 96\%              \\ \cline{4-5} 
                              &                                         &                                                                                                                  & Sensitivity               & 94\%              \\ \hline
\end{tabular}
\end{table*}


\subsection{Drug discovery}
This section deals with the application of machine learning algorithms in the Pharma industry. \cite{t128} used machine learning for drug design. For this work, the prediction of inhibition of dihydrofolate reductase from pyrimidines is made using a support vector machine. The data for this research is taken from the University of California Irvine(UCI) machine learning repository. In this paper, the performance of Support Vector Machine is compared with the Prune network, multilayer perceptron(MLP) feedforward network, radial bias function network(RBF), dynamic network, and C5.0. The support vector machine performed best among the above algorithms with a training time of 93 seconds.
Machine learning is used by \cite{t129} to predict proximal tubular toxicity in humans. For this work, the researchers developed a one-step protocol to differentiate human induced pluripotent stem cells(hiPCS) into proximal tubular like cells. A random forest algorithm is used for prediction. This algorithm is evaluated based on sensitivity, specificity, and accuracy and is shown in table 18. 
Drug likeliness is predicted using machine learning by \cite{t130}. In this work, Bayesian classifier, recursive portioning models were developed to predict the drug likeliness, and these models are evaluated based on the compounds from the Traditional Chinese medicine compound database(TCMCD). Bayesian classifier with the LCFP\_6 fingerprint set and physicochemical properties outperformed the Recursive partitioning model. 
A study was performed by \cite{t131} on using machine learning and deep learning techniques in drug discovey. In this paper the authors surveyed the different machine learning and deep learning models that were used in target identification and validation, design and optimization of small molecules, discovery of biomarkers, prediction of drug sensitivity and in computational pathology.

\begin{table*}[]
\centering
\caption{ML and DL methods in Drug Discovery.}
\begin{tabular}{|l|l|l|l|l|}
\hline
\textbf{Author}               & \textbf{Method}                            & \textbf{Application}                       & \textbf{Metrics}                                                  & \textbf{Findings} \\ \hline
\cite{t128}                  & SVM                                        & Drug design                                & Error                                                             & 0.1269            \\ \hline
{\cite{t129}} & {Random Forest}             & {Proximal tubular toxicity} & Sensitivity                                                       & 89.0\%            \\ \cline{4-5} 
                              &                                            &                                            & Specificity                                                       & 85.0\%            \\ \cline{4-5} 
                              &                                            &                                            & Accuracy                                                          & 87.0\%            \\ \hline
{\cite{t130}} & {Naive Bayesian Classifier} & {Drug likeliness}           & \begin{tabular}[c]{@{}l@{}}Accuracy\\ (test set)\end{tabular}     & 90.9\%            \\ \cline{4-5} 
                              &                                            &                                            & \begin{tabular}[c]{@{}l@{}}Accuracy\\ (training set)\end{tabular} & 91.4\%            \\ \hline
\end{tabular}
\end{table*}

\subsection{COVID-19}
ML and DL models play a prominent role in current pandemic situation, COVID-19. Researchers have done a extensive use of ML and DL methods for helping the healthcare professionals in diagnosing the disease through multiple sources ranging from X-rays till individual's comorbities. The authors have done a thorough literature study and the results obtained are shown in table \ref{c}.
\cite{t41} proposed a deep learning model named COVIDX-Net consisting of seven convolutional neural networks(CNN). This COVIDX-Net model diagnosed COVID-19 infection from x-ray images with an F1-score of 0.91. A deep learning open source model named COVID-Net was proposed by \cite{t42} to diagnose COVID-19. This model was able to classify the x-ray images into three classes namely normal, Pneumonia, and COVID-19 with an accuracy of 92\begin{math}\%\end{math}. \cite{t43} performed a comparative study on using state-of-the-art deep learning architectures namely VGG19, MobileNet, Xception, Inception, and Inception ResNet v2 for the diagnosis of COVID-19. According to this study, VGG19 and MobileNet showed enhanced performance compared to the other models with an accuracy of 98.75\begin{math}\%\end{math} and 97.40\begin{math}\%\end{math}. Another comparative study was performed by \cite{t44} on using ResNet50, ResNet152, ResNet101, Inception-ResNetV2 and InceptionV3 to diagnose COVID-19 from chest x-ray images of three different datasets. According to this study, it was observed that ResNet50 performed better than other models with the accuracies of 96.1\begin{math}\%\end{math}, 99.5\begin{math}\%\end{math}, and 99.7\begin{math}\%\end{math} on all the three datasets. \cite{t45} diagnosed COVID-19 from x-ray images using a support vector machine trained with features extracted by the ResNet50 model. \citep{t46} developed a deep learning-based diagnosis system known as DeepPneumonia to diagnose COVID-19 and bacterial pneumonia. In this work, the authors used data collected from 86 healthy persons, 101 patients diagnosed with bacterial pneumonia, and 88 patients diagnosed with COVID-19. This data was collected from 19 hospitals in China. This model was tested with images from two datasets. \cite{t48} developed a weakly-supervised deep learning model for the diagnosis of COVID-19 from chest CT images. In this work, a pre-trained U-Net architecture is used to segment the lung region from the image and a 3D deep neural network was used to diagnose the COVID-19 infection in the lungs. This system is very fast at it takes only 1.93s to process a CT image of a patient and diagnoses the disease with an accuracy of 90.1\begin{math}\%\end{math} and specificity of 91.1\begin{math}\%\end{math}. \cite{t49} proposed a methodology to diagnose COVID-19 pneumonia and Influenza-A viral pneumonia. In this methodology, a 3D CNN model is used to segment the candidate regions in the lungs and a location-attention classifier is used to classify these regions into COVID-19 pneumonia, influenza-A viral pneumonia and as normal and finally, a noisy or Bayesian function is used to calculate the infection probability. \cite{t50} performed a study on using a support vector machine as a classifier trained with features extracted using five different algorithms for the diagnosis of COVID-19 CT images. The five feature extraction algorithms that were used in this work are Local Directional Pattern(LDP), Grey Level Co-occurrence Matrix(GLCM), Grey-Level Size Zone Matrix (GLSZM), Grey Level Run Length Matrix (GLRLM), and Discrete Wavelet Transform (DWT). According to this study support vector machine trained with features extracted using the GLSZM algorithm performed better than other feature extraction algorithms with an accuracy of 99.68\begin{math}\%\end{math}. \cite{t51} proposed a methodology to diagnose COVID-19 using multi-level thresholding and a support vector machine. \cite{t52} designed a deep learning model using transfer learning UNet for the diagnosis of COVID-19. For this work, 46,096 anonymous CT images are collected from 106 patients admitted to Renmin hospital in Wuhan city were used, out of which 51 patients are diagnosed with COVID-19. \cite{to53} proposed a transfer learning model by fine-tuning the pre-trained model VGG16 for the diagnosis of COVID-19.  \cite{t54} compared two deep learning models that were built using the transfer learning of residual networks ResNet34 and ResNet50 for the diagnosis of pneumonia and COVID-19. According to this paper, ResNet50 performed better than ResNet32 with an accuracy of 72.38\begin{math}\%\end{math}. \cite{t55} proposed a deep learning model that is designed using VGG16 and a faster R-CNN framework to diagnose COVID-19. 

\begin{table*}[]
\centering
\caption{ML and DL methods for the diagnosis of COVID-19.}
\label{c}
\begin{tabular}{|l|l|l|l|l|}
\hline
\textbf{Author}              & \textbf{Method}                                                                                                              & \textbf{Disease}                                                                                 & \textbf{Metrics} & \textbf{Findings} \\ \hline
\cite{t41}                  & \begin{tabular}[c]{@{}l@{}}Ensemble of 7 \\ convolutional \\ neural networks\end{tabular}                                    & COVID-19                                                                                         & f1-score         & 0.91              \\ \hline
\cite{t42}                  & Deep CNN                                                                                                                     & \begin{tabular}[c]{@{}l@{}}Pneumonia and \\ COVID-19\end{tabular}                                & Accuracy         & 92\%              \\ \hline
\cite{t45}                  & \begin{tabular}[c]{@{}l@{}}Support vector \\ machine trained \\ with features \\ extracted by VGG16\end{tabular}             & COVID-19                                                                                         & Accuracy         & 95.38\%           \\ \hline
  {\cite{t46}} &   {Deep CNN}                                                                                                    &   {\begin{tabular}[c]{@{}l@{}}COVID-19 and \\ bacterial pneumonia\end{tabular}}     & Accuracy         & 86\%              \\ \cline{4-5} 
                             &                                                                                                                              &                                                                                                  & Sensitivity      & 96\%              \\ \hline
\cite{t47}                  & \begin{tabular}[c]{@{}l@{}}Transfer learning \\ of InceptionV3\end{tabular}                                                  & \begin{tabular}[c]{@{}l@{}}COVID-19 and viral \\ pneumonia\end{tabular}                          & Accuracy         & 89.5\%            \\ \hline
\cite{t48}                  & \begin{tabular}[c]{@{}l@{}}U-Net based \\ segmentation and \\ weakly supervised \\ learning based \\ classifier\end{tabular} & COVID-19                                                                                         & Accuracy         & 90.1\%            \\ \hline
\cite{t49}                  & \begin{tabular}[c]{@{}l@{}}3D CNN based \\ segmentation with \\ attention based classifier\end{tabular}                      & \begin{tabular}[c]{@{}l@{}}COVID-19 pneumonia \\ and Influenza A viral \\ Pneumonia\end{tabular} & Accuracy         & 86.7\%            \\ \hline
  {\cite{t51}} &   {\begin{tabular}[c]{@{}l@{}}Multi-level thresholding \\ and SVM\end{tabular}}                                 &   {COVID-19}                                                                        & Accuracy         & 97.84\%           \\ \cline{4-5} 
                             &                                                                                                                              &                                                                                                  & Sensitivity      & 95.76\%           \\ \cline{4-5} 
                             &                                                                                                                              &                                                                                                  & Specificity      & 99.7\%            \\ \hline
\cite{t52}                  & \begin{tabular}[c]{@{}l@{}}Transfer learning of \\ U-Net\end{tabular}                                                        & COVID-19                                                                                         & Accuracy         & 98.85\%           \\ \hline
\cite{to53}                  & \begin{tabular}[c]{@{}l@{}}Transfer learning by \\ fine tuning VGG16\end{tabular}                                            & COVID-19                                                                                         & Accuracy         & 96\%              \\ \hline
  {\cite{t55}} &   {\begin{tabular}[c]{@{}l@{}}CNN designed using \\ VGG16 and Faster \\ RCNN framework\end{tabular}}            &   {COVID-19}                                                                        & Accuracy         & 97.36\%           \\ \cline{4-5} 
                             &                                                                                                                              &                                                                                                  & Sensitivity      & 97.65\%           \\ \cline{4-5} 
                             &                                                                                                                              &                                                                                                  & Precision        & 99.28\%           \\ \hline
\end{tabular}
\end{table*}

\section{Discussion}
\subsection{Smart Healthcare Systems using ML and DL}
The usage of ML and DL models created a massive impact across all domains especially, the healthcare industry. Even with the advent of many reports and scans, human decision making is the only way for diagnosing diseases. This may sometimes lead to unreliable diagnosis due to bias in human decisions. Smart intelligent techniques like ML and DL methodologies can be used to improve reliability in diagnosis thereby saving lot of human lives. Thus researchers started proposing various models for automating diagnosis in healthcare in various specializations including Orthopedics( estimating bone density, diagnosing rheumatoid arthritis), Radiology( interpreting mammography, computed tomography and magnetic resonance imaging scans), Ophthalmology( diagnosis of diabetic retinopathy, cataract, glaucoma), Cardiology( diagnosis of complex diseases like hypertrophic cardiomyopathy) and many more. Such diagnosis may help saving lives of human. For eg., according to a study performed by the American Cancer Society \cite{t132} it was found that more than 50\begin{math}\%\end{math} of the women who are getting annual mammograms over 10 years are having false-positive findings. These false-positive findings in mammograms can lead to unnecessary examinations like ultrasound scans, MRI, and even biopsies and this can be avoided by using ML and DL based systems.

\subsection{Impact of Computerized diagnostic systems on physicians}
Machine learning and deep learning are playing a key role in all stages of drug discovery. These techniques are used extensively for better understanding of disease mechanisms, non-disease and disease phenotypes, to identify novel targets and in development of biomarkers for drug efficiency and prognosis.  These systems are becoming more effective than medical professionals in terms of diagnosis and predictive analysis of diseases \cite{t133,t134}. This led to a major concern that these systems will replace medical professionals in near future, especially radiologists. The main reason for this concern is its ability to learn from millions of images and this learning enables the computer systems to diagnose the most complex diseases that are difficult to be diagnosed by experienced medical professionals. But there are few limitations with these systems, so we can’t completely depend on these systems for diagnosis. But these systems can be a part of a radiologist's life, by assisting them during diagnosis and making their work more accurate and efficient.
 
In the case of other medical specialties, computer systems cannot be a replacement for doctors because these systems cannot gain trust in patients and cannot have high-level interactions with the patients. Even though one might say that there is a possibility that these systems can make clinical conversations in the future. But, even if computer systems are built to an extent that they can conduct real-time MRI, CT, and other imaging examinations and perform automated surgeries, they cannot replace doctors. Doctors are still needed for the diagnosis and treatment of novel and ambiguous cases. These computer systems under-perform in case of novel diseases, side effects caused by drugs because there is no prior instance available for these systems to learn from \cite{t135}. For these reasons, it can be concluded the smart intelligent computer systems will support doctors by assisting them during the diagnosis and treatment of diseases instead of replacing them.

\section{Conclusion}
The diagnosis of diseases is a complex and time consuming process. The conventional procedures followed to diagnose diseases are not so accurate. Intelligent computer systems are being used by medical professionals for fast and accurate diagnosis of diseases. There is a lot of research that is done in this field. So the main aim of this paper is to present a comprehensive survey of different machine learning and deep learning models that are developed and used in building computer systems to diagnose diseases across various specializations of medicine. Along with the survey, we discuss the advancements in healthcare with machine learning and deep learning-based systems and also the impacts of these systems on medical professionals. 
This paper provides a  deep dive into the different frameworks, models and tools along with practical concerns and considerations in Health care domain that include Dental medicine, Haematology, Surgery, Cardiology, Pulmonology, Orthopedics, Radiology, Oncology, General medicine, Psychiatry, Endocrinology, Neurology, Dermatology, Hepatology, Nephrology, and Ophthalmology. Along with the medical specializations mentioned above, we also present the survey of machine learning and deep learning models in drug discovery.

\subsection{Highlights of the work}
\begin{itemize}
    \item Exploring prominent research works in designing the various robust Intelligent health care systems using ML and DL techniques for early diagnosis of diseases.
    \item Examining constructive research outputs demonstrating a promising supplementary diagnostic method for frontline clinical doctors and surgeons.
    \item Thorough analysis of applications of ML and DL across 16 medical specialties, namely Dental medicine, Haematology, Surgery, Cardiology, Pulmonology, Orthopaedics, Radiology, Oncology, General medicine, Psychiatry, Endocrinology, Neurology, Dermatology, Hepatology, Nephrology, Ophthalmology, and Drug discovery.
    \item Deep dive into the different ML and DL frameworks, models and tools along with practical concerns, challenges and considerations in Health care domain.
    \item Investigating the use of ML and DL mechanisms in predicting COVID-19, the deadly and infectious disease.
    \item Look into the impact of Computerized diagnostic systems on physicians with a promising discussion on future research deliberations.
	\item Recording relevant list of Public Data sets available for future works.

\end{itemize}

\section{Suggestions for future work}
\begin{itemize}
    \item Fuzzy based diagnosis systems can be built to diagnose complex diseases like cancers, heart attacks, and diabetic retinopathy.
    \item Building Neural network-based systems to diagnose diseases from image-based data like X-rays, CT, and MRI scans.
    \item Transfer learning models can be built to diagnose diseases that are difficult to detect, like novel fevers, celiac disease, brain, and ovarian cancers, diabetic eye disease, and glaucoma.
    \item Designing systems with machine learning and Internet of Things(IoT) techniques for the evaluation of organs and for diagnosis of Haematological diseases from blood samples. 
    \item Developing a pattern recognition based model for the diagnosis of Cardiovascular and Pulmonary diseases. 
    \item Building hybrid systems with image recognition approaches for the staging of various cancers, for diagnosis of neurological diseases like brain hemorrhage and hydrocephalus.
    \item Developing robots and Surgeon-assisting tools with machine learning techniques to perform robotic and minimally invasive surgeries.
    \item Building a system that estimates the stress levels and suggests lifestyle changes that are needed to reduce stress.
    \item Building systems with a different combination of algorithms using ensemble methods for diagnosing diseases.
\end{itemize}

\section{List of abbreviations}

\noindent
AI    $\>$ - $\>$  Artificial Intelligence \newline
ML    $\>$ - $\>$  Machine learning \newline
DL    $\>$ - $\>$  Deep learning \newline
MRI   $\>$ - $\>$  Magnetic Resonance Imaging\newline
RVG	  $\>$ - $\>$ Radio Videography\newline
CNN   $\>$ - $\>$ Convolutional Neural Network\newline
UCI	  $\>$ - $\>$ University of California\newline
MLP	  $\>$ - $\>$ Multi Layer Perceptron\newline
RBF	  $\>$ - $\>$ Radial Bias Function\newline
hiPCS $\>$ - $\>$ Human induced pluripotent stem cells\newline
TCMCD $\>$ - $\>$ Traditional Chinese medicine compound database\newline
ECG	  $\>$ - $\>$ Electrocardiogram\newline
ReLU  $\>$ - $\>$ Rectified Linear Unit\newline
HCM   $\>$ - $\>$ Hypertrophic cardiomyopathy\newline
SPET  $\>$ - $\>$ Single-photon emission computed tomography\newline
MACE  $\>$ - $\>$ major adverse cardiac events\newline
ARDS  $\>$ - $\>$ Acute diseases like Respiratory distress syndrome\newline
ROC   $\>$ - $\>$ Receiver operating characteristic\newline
EDR	  $\>$ - $\>$ Electrocardiogram derived respiratory\newline
CT    $\>$ - $\>$ Computed tomography\newline
JSN	  $\>$ - $\>$ Joint space narrowing\newline
SWI   $\>$ - $\>$ Susceptibility weighted imaging\newline
LASIK $\>$ - $\>$ Laser-Assisted In-Situ Keratomileusis\newline
IoT	  $\>$ - $\>$ Internet of Things\newline
SVM	  $\>$ - $\>$ Support Vector Machine\newline
KNN	  $\>$ - $\>$ K-Nearest Neighbor\newline
LMLA  $\>$ - $\>$ Levenberg Marquardt Learning Algorithm\newline
DWNN  $\>$ - $\>$ Distance-weighted -nearest-neighbor\newline
ELM	  $\>$ - $\>$ Extreme Learning Machine\newline
RVM	  $\>$ - $\>$ Relvence Vector Machine	\newline
LVQ	  $\>$ - $\>$ Linear Vector Quantization\newline
LR	  $\>$ - $\>$ Logistic Regression\newline
GA	  $\>$ - $\>$ Genetic Algorithm\newline
GAN	  $\>$ - $\>$ Generative adversarial neural networks\newline

\section{List of public datasets}
\begin{itemize}
    \item ChestXpert\newline 
    \href{https://stanfordmlgroup.github.io/competitions/chexpert/}{https://stanfordmlgroup.github.io/competitions/chexpert/}

    \item ChestXray-NIHCC\newline
    \href{https://nihcc.app.box.com/v/ChestXray-NIHCC}{https://nihcc.app.box.com/v/ChestXray-NIHCC}
    
    \item MIMIC-CXR\newline
    \href{https://physionet.org/physiobank/database/mimiccxr/}{https://physionet.org/physiobank/database/mimiccxr/}
    
    \item PadChest\newline
    \href{http://bimcv.cipf.es/bimcv-projects/padchest/}{http://bimcv.cipf.es/bimcv-projects/padchest/}
    
    \item IBM Xray Eye Gaze\newline
    \href{https://physionet.org/content/egd-cxr/1.0.0/}{https://physionet.org/content/egd-cxr/1.0.0/}
    
    \item Cancer Image Archive\newline
    \href{http://www.cancerimagingarchive.net/}{http://www.cancerimagingarchive.net/}
    
    \item National Lung Screening Trial\newline
    \href{https://wiki.cancerimagingarchive.net/display/NLST/ National+Lung+Screening+Trial}{https://wiki.cancerimagingarchive.net/display/NLST/ National+Lung+Screening+Trial}
    
    \item DeepLesion \newline
    \href{https://nihcc.app.box.com/v/DeepLesion}{https://nihcc.app.box.com/v/DeepLesion}

    \item EchoNet-Dynamic \newline
    \href{https://echonet.github.io/dynamic/}{https://echonet.github.io/dynamic/}

    \item ABCD Neurocognitive Prediction Challenge \newline
    \href{https://sibis.sri.com/abcd-np-challenge/}{https://sibis.sri.com/abcd-np-challenge/}
    
    \item AAPM Sparse-View CT Reconstruction Challenge \newline
    \href{https://www.aapm.org/GrandChallenge/DL-sparse-view-CT/}{https://www.aapm.org/GrandChallenge/DL-sparse-view-CT/}
    
    \item Cross-Sectional Multidomain Lexical Processing\newline
    \href{https://openneuro.org/datasets/ds002236}{https://openneuro.org/datasets/ds002236}

    \item Neurite-OASIS \newline
    \href{https://github.com/adalca/medical-datasets/blob/master/neurite-oasis.md}{https://github.com/adalca/medical-datasets/blob/master/neurite-oasis.md}

    \item MRNet \newline
    \href{https://stanfordmlgroup.github.io/competitions/mrnet/}{https://stanfordmlgroup.github.io/competitions/mrnet/}
    
    \item fastMRI \newline
    \href{https://fastmri.med.nyu.edu/}{https://fastmri.med.nyu.edu/}

    \item OCMR \newline
    \href{https://ocmr.info/}{https://ocmr.info/}

    \item PREVENT-AD \newline
    \href{https://openpreventad.loris.ca/}{https://openpreventad.loris.ca/}
    
    \item Medical Segmentation Decathlon \newline
    \href{http://medicaldecathlon.com/}{http://medicaldecathlon.com/}

    \item MASSIVE \newline 
    \href{http://massive-data.org/download.html}{http://massive-data.org/download.html}

    \item AOMIC: the Amsterdam Open MRI Collection \newline
    \href{https://nilab-uva.github.io/AOMIC.github.io/}{https://nilab-uva.github.io/AOMIC.github.io/}
    
    \item MRIdata \newline
    \href{http://mridata.org/}{http://mridata.org/}
    
    \item Brain MRI LGG FLAIR abnormality segmentation \newline
    \href{https://www.kaggle.com/mateuszbuda/lgg-mri-segmentation}{https://www.kaggle.com/mateuszbuda/lgg-mri-segmentation}

    \item Studyforrest \newline
    \href{http://studyforrest.org/data.html}{http://studyforrest.org/data.html}
    
    \item Lung Image Database Consortium \newline
    \href{https://wiki.cancerimagingarchive.net/display/Public/LIDC-IDRI}{https://wiki.cancerimagingarchive.net/display/Public/LIDC-IDRI}

    \item UK Biobank \newline 
    \href{https://biobank.ctsu.ox.ac.uk/crystal/download.cgi}{https://biobank.ctsu.ox.ac.uk/crystal/download.cgi}
    
    \item BrixIA: COVID19 severity score assessment databse \newline
    \href{https://brixia.github.io/}{https://brixia.github.io/}
    
    \item COVID-CT \newline
    \href{https://github.com/UCSD-AI4H/COVID-CT}{https://github.com/UCSD-AI4H/COVID-CT}

    \item Medical Imaging Data Resource Center (MIDRC) \newline
    \href{https://wiki.cancerimagingarchive.net/pages/viewpage.action?pageId=70230281}{https://wiki.cancerimagingarchive.net/pages/\newline viewpage.action?pageId=70230281}
    
    \item BIMCV-COVID19 \newline 
    \href{http://bimcv.cipf.es/bimcv-projects/bimcv-covid19/}{http://bimcv.cipf.es/bimcv-projects/bimcv-covid19/}

    \item MosMedData Covid19 \newline
    \href{https://mosmed.ai/en/}{https://mosmed.ai/en/}
    
    \item COVID-19 LUNG CT LESION SEGMENTATION CHALLENGE \newline
    \href{https://covid-segmentation.grand-challenge.org/Data/}{https://covid-segmentation.grand-challenge.org/Data/}
    
    \item MedSeg COVID-19 CT \newline
    \href{http://medicalsegmentation.com/covid19/}{http://medicalsegmentation.com/covid19/}

    \item COVID-Chest XRay \newline
    \href{https://github.com/ieee8023/covid-chestxray-dataset}{https://github.com/ieee8023/covid-chestxray-dataset}

    \item BSTI COVID19 \newline
    \href{https://bsticovid19.cimar.co.uk/worklist/?embedded=}{https://bsticovid19.cimar.co.uk/worklist/?embedded=}

    \item RICORD \newline
    \href{https://www.rsna.org/covid-19/COVID-19-RICORD/RICORD-resources}{https://www.rsna.org/covid-19/COVID-19-RICORD/RICORD-resources}

    \item FIRE (Fundus Image Registration Dataset) \newline
    \href{https://paperswithcode.com/dataset/fire}{https://paperswithcode.com/dataset/fire}

    \item DRIVE: Digital Retinal Images for Vessel Extraction \newline
    \href{https://drive.grand-challenge.org/}{https://drive.grand-challenge.org/}

    \item FLARE: Fast and Low GPU memory Abdominal Organ Segmentation \newline
    \href{https://flare.grand-challenge.org/}{https://flare.grand-challenge.org/}
    
    \item Diabetes \newline
    \href{https://archive.ics.uci.edu/ml/datasets/Diabetes}{https://archive.ics.uci.edu/ml/datasets/Diabetes}
    
    \item Thyroid Disease \newline
    \href{https://archive.ics.uci.edu/ml/datasets/Thyroid+Disease}{https://archive.ics.uci.edu/ml/datasets/Thyroid+Disease}

    \item Breast cancer \newline
    \href{https://archive.ics.uci.edu/ml/datasets/Breast+Cancer}{https://archive.ics.uci.edu/ml/datasets/Breast+Cancer}

\end{itemize}





\bibliographystyle{model2-names.bst}\biboptions{authoryear}
\bibliography{refs}

\begin{thebibliography}{134}
\expandafter\ifx\csname natexlab\endcsname\relax\def\natexlab#1{#1}\fi
\providecommand{\url}[1]{\texttt{#1}}
\providecommand{\href}[2]{#2}
\providecommand{\path}[1]{#1}
\providecommand{\DOIprefix}{doi:}
\providecommand{\ArXivprefix}{arXiv:}
\providecommand{\URLprefix}{URL: }
\providecommand{\Pubmedprefix}{pmid:}
\providecommand{\doi}[1]{\href{http://dx.doi.org/#1}{\path{#1}}}
\providecommand{\Pubmed}[1]{\href{pmid:#1}{\path{#1}}}
\providecommand{\bibinfo}[2]{#2}
\ifx\xfnm\relax \def\xfnm[#1]{\unskip,\space#1}\fi
\bibitem[{Acharya et~al.(2017)Acharya, Fujita, Oh, Hagiwara, Tan and
  Adam}]{t25}
\bibinfo{author}{Acharya, U.R.}, \bibinfo{author}{Fujita, H.},
  \bibinfo{author}{Oh, S.L.}, \bibinfo{author}{Hagiwara, Y.},
  \bibinfo{author}{Tan, J.H.}, \bibinfo{author}{Adam, M.},
  \bibinfo{year}{2017}.
\newblock \bibinfo{title}{Application of deep convolutional neural network for
  automated detection of myocardial infarction using ecg signals}.
\newblock \bibinfo{journal}{Information Sciences} \bibinfo{volume}{415},
  \bibinfo{pages}{190--198}.
\bibitem[{Ahmed and Soomrani(2016)}]{t94}
\bibinfo{author}{Ahmed, J.}, \bibinfo{author}{Soomrani, M.A.R.},
  \bibinfo{year}{2016}.
\newblock \bibinfo{title}{Tdtd: Thyroid disease type diagnostics}, in:
  \bibinfo{booktitle}{2016 International Conference on Intelligent Systems
  Engineering (ICISE)}, \bibinfo{organization}{IEEE}. pp.
  \bibinfo{pages}{44--50}.
\bibitem[{Alassaf et~al.(2018)Alassaf, Alsulaim, Alroomi, Alsharif, Aljubeir,
  Olatunji, Alahmadi, Imran, Alzahrani and Alturayeif}]{t90}
\bibinfo{author}{Alassaf, R.A.}, \bibinfo{author}{Alsulaim, K.A.},
  \bibinfo{author}{Alroomi, N.Y.}, \bibinfo{author}{Alsharif, N.S.},
  \bibinfo{author}{Aljubeir, M.F.}, \bibinfo{author}{Olatunji, S.O.},
  \bibinfo{author}{Alahmadi, A.Y.}, \bibinfo{author}{Imran, M.},
  \bibinfo{author}{Alzahrani, R.A.}, \bibinfo{author}{Alturayeif, N.S.},
  \bibinfo{year}{2018}.
\newblock \bibinfo{title}{Preemptive diagnosis of diabetes mellitus using
  machine learning}, in: \bibinfo{booktitle}{2018 21st Saudi Computer Society
  National Computer Conference (NCC)}, \bibinfo{organization}{IEEE}. pp.
  \bibinfo{pages}{1--5}.
\bibitem[{Albawi et~al.(2017)Albawi, Mohammed and Al-Zawi}]{t10}
\bibinfo{author}{Albawi, S.}, \bibinfo{author}{Mohammed, T.A.},
  \bibinfo{author}{Al-Zawi, S.}, \bibinfo{year}{2017}.
\newblock \bibinfo{title}{Understanding of a convolutional neural network}, in:
  \bibinfo{booktitle}{2017 International Conference on Engineering and
  Technology (ICET)}, \bibinfo{organization}{Ieee}. pp. \bibinfo{pages}{1--6}.
\bibitem[{Ali{\'c} et~al.(2017)Ali{\'c}, Gurbeta and Badnjevi{\'c}}]{t27}
\bibinfo{author}{Ali{\'c}, B.}, \bibinfo{author}{Gurbeta, L.},
  \bibinfo{author}{Badnjevi{\'c}, A.}, \bibinfo{year}{2017}.
\newblock \bibinfo{title}{Machine learning techniques for classification of
  diabetes and cardiovascular diseases}, in: \bibinfo{booktitle}{2017 6th
  mediterranean conference on embedded computing (MECO)},
  \bibinfo{organization}{IEEE}. pp. \bibinfo{pages}{1--4}.
\bibitem[{Angra and Ahuja(2017)}]{t1}
\bibinfo{author}{Angra, S.}, \bibinfo{author}{Ahuja, S.}, \bibinfo{year}{2017}.
\newblock \bibinfo{title}{Machine learning and its applications: A review}, in:
  \bibinfo{booktitle}{2017 International Conference on Big Data Analytics and
  Computational Intelligence (ICBDAC)}, \bibinfo{organization}{IEEE}. pp.
  \bibinfo{pages}{57--60}.
\bibitem[{Apostolopoulos and Mpesiana(2020)}]{t43}
\bibinfo{author}{Apostolopoulos, I.D.}, \bibinfo{author}{Mpesiana, T.A.},
  \bibinfo{year}{2020}.
\newblock \bibinfo{title}{Covid-19: automatic detection from x-ray images
  utilizing transfer learning with convolutional neural networks}.
\newblock \bibinfo{journal}{Physical and Engineering Sciences in Medicine}
  \bibinfo{volume}{43}, \bibinfo{pages}{635--640}.
\bibitem[{Arifin et~al.(2012)Arifin, Kibria, Firoze, Amini and Yan}]{t109}
\bibinfo{author}{Arifin, M.S.}, \bibinfo{author}{Kibria, M.G.},
  \bibinfo{author}{Firoze, A.}, \bibinfo{author}{Amini, M.A.},
  \bibinfo{author}{Yan, H.}, \bibinfo{year}{2012}.
\newblock \bibinfo{title}{Dermatological disease diagnosis using color-skin
  images}, in: \bibinfo{booktitle}{2012 international conference on machine
  learning and cybernetics}, \bibinfo{organization}{IEEE}. pp.
  \bibinfo{pages}{1675--1680}.
\bibitem[{Arun et~al.(2018)Arun, Prajwal, Krishna, Arunkumar, Padma and
  Shyam}]{t80}
\bibinfo{author}{Arun, V.}, \bibinfo{author}{Prajwal, V.},
  \bibinfo{author}{Krishna, M.}, \bibinfo{author}{Arunkumar, B.},
  \bibinfo{author}{Padma, S.}, \bibinfo{author}{Shyam, V.},
  \bibinfo{year}{2018}.
\newblock \bibinfo{title}{Neural network in a projection learning framework for
  depression classification in mynah cohort}, in: \bibinfo{booktitle}{2018 IEEE
  Symposium Series on Computational Intelligence (SSCI)},
  \bibinfo{organization}{IEEE}. pp. \bibinfo{pages}{25--32}.
\bibitem[{Ausawalaithong et~al.(2018)Ausawalaithong, Thirach, Marukatat and
  Wilaiprasitporn}]{t72}
\bibinfo{author}{Ausawalaithong, W.}, \bibinfo{author}{Thirach, A.},
  \bibinfo{author}{Marukatat, S.}, \bibinfo{author}{Wilaiprasitporn, T.},
  \bibinfo{year}{2018}.
\newblock \bibinfo{title}{Automatic lung cancer prediction from chest x-ray
  images using the deep learning approach}, in: \bibinfo{booktitle}{2018 11th
  Biomedical Engineering International Conference (BMEICON)},
  \bibinfo{organization}{IEEE}. pp. \bibinfo{pages}{1--5}.
\bibitem[{Ayeldeen et~al.(2015)Ayeldeen, Shaker, Ayeldeen and Anwar}]{t117}
\bibinfo{author}{Ayeldeen, H.}, \bibinfo{author}{Shaker, O.},
  \bibinfo{author}{Ayeldeen, G.}, \bibinfo{author}{Anwar, K.M.},
  \bibinfo{year}{2015}.
\newblock \bibinfo{title}{Prediction of liver fibrosis stages by machine
  learning model: A decision tree approach}, in: \bibinfo{booktitle}{2015 Third
  World Conference on Complex Systems (WCCS)}, \bibinfo{organization}{IEEE}.
  pp. \bibinfo{pages}{1--6}.
\bibitem[{Barakat et~al.(2010)Barakat, Bradley and Barakat}]{t91}
\bibinfo{author}{Barakat, N.}, \bibinfo{author}{Bradley, A.P.},
  \bibinfo{author}{Barakat, M.N.H.}, \bibinfo{year}{2010}.
\newblock \bibinfo{title}{Intelligible support vector machines for diagnosis of
  diabetes mellitus}.
\newblock \bibinfo{journal}{IEEE transactions on information technology in
  biomedicine} \bibinfo{volume}{14}, \bibinfo{pages}{1114--1120}.
\bibitem[{Barstugan et~al.(2020)Barstugan, Ozkaya and Ozturk}]{t50}
\bibinfo{author}{Barstugan, M.}, \bibinfo{author}{Ozkaya, U.},
  \bibinfo{author}{Ozturk, S.}, \bibinfo{year}{2020}.
\newblock \bibinfo{title}{Coronavirus (covid-19) classification using ct images
  by machine learning methods}.
\newblock \bibinfo{journal}{arXiv preprint arXiv:2003.09424} .
\bibitem[{Betancur et~al.(2018)Betancur, Otaki, Motwani, Fish, Lemley, Dey,
  Gransar, Tamarappoo, Germano, Sharir et~al.}]{t33}
\bibinfo{author}{Betancur, J.}, \bibinfo{author}{Otaki, Y.},
  \bibinfo{author}{Motwani, M.}, \bibinfo{author}{Fish, M.B.},
  \bibinfo{author}{Lemley, M.}, \bibinfo{author}{Dey, D.},
  \bibinfo{author}{Gransar, H.}, \bibinfo{author}{Tamarappoo, B.},
  \bibinfo{author}{Germano, G.}, \bibinfo{author}{Sharir, T.}, et~al.,
  \bibinfo{year}{2018}.
\newblock \bibinfo{title}{Prognostic value of combined clinical and myocardial
  perfusion imaging data using machine learning}.
\newblock \bibinfo{journal}{JACC: Cardiovascular Imaging} \bibinfo{volume}{11},
  \bibinfo{pages}{1000--1009}.
\bibitem[{Bian and Zhang(2018)}]{t58}
\bibinfo{author}{Bian, Z.}, \bibinfo{author}{Zhang, R.}, \bibinfo{year}{2018}.
\newblock \bibinfo{title}{Bone age assessment method based on deep
  convolutional neural network}, in: \bibinfo{booktitle}{2018 8th International
  Conference on Electronics Information and Emergency Communication (ICEIEC)},
  \bibinfo{organization}{IEEE}. pp. \bibinfo{pages}{194--197}.
\bibitem[{Cai et~al.(2018)Cai, Chen and Qiu}]{t118}
\bibinfo{author}{Cai, J.}, \bibinfo{author}{Chen, T.}, \bibinfo{author}{Qiu,
  X.}, \bibinfo{year}{2018}.
\newblock \bibinfo{title}{Fibrosis and inflammatory activity analysis of
  chronic hepatitis c based on extreme learning machine}, in:
  \bibinfo{booktitle}{2018 9th International Conference on Information
  Technology in Medicine and Education (ITME)}, \bibinfo{organization}{IEEE}.
  pp. \bibinfo{pages}{177--181}.
\bibitem[{Cao et~al.(2015)Cao, Wang, Moradi, Prasanna and Syeda-Mahmood}]{t61}
\bibinfo{author}{Cao, Y.}, \bibinfo{author}{Wang, H.}, \bibinfo{author}{Moradi,
  M.}, \bibinfo{author}{Prasanna, P.}, \bibinfo{author}{Syeda-Mahmood, T.F.},
  \bibinfo{year}{2015}.
\newblock \bibinfo{title}{Fracture detection in x-ray images through stacked
  random forests feature fusion}, in: \bibinfo{booktitle}{2015 IEEE 12th
  international symposium on biomedical imaging (ISBI)},
  \bibinfo{organization}{IEEE}. pp. \bibinfo{pages}{801--805}.
\bibitem[{Chambres et~al.(2018)Chambres, Hanna and Desainte-Catherine}]{t36}
\bibinfo{author}{Chambres, G.}, \bibinfo{author}{Hanna, P.},
  \bibinfo{author}{Desainte-Catherine, M.}, \bibinfo{year}{2018}.
\newblock \bibinfo{title}{Automatic detection of patient with respiratory
  diseases using lung sound analysis}, in: \bibinfo{booktitle}{2018
  International Conference on Content-Based Multimedia Indexing (CBMI)},
  \bibinfo{organization}{IEEE}. pp. \bibinfo{pages}{1--6}.
\bibitem[{Charleonnan et~al.(2016)Charleonnan, Fufaung, Niyomwong,
  Chokchueypattanakit, Suwannawach and Ninchawee}]{t119}
\bibinfo{author}{Charleonnan, A.}, \bibinfo{author}{Fufaung, T.},
  \bibinfo{author}{Niyomwong, T.}, \bibinfo{author}{Chokchueypattanakit, W.},
  \bibinfo{author}{Suwannawach, S.}, \bibinfo{author}{Ninchawee, N.},
  \bibinfo{year}{2016}.
\newblock \bibinfo{title}{Predictive analytics for chronic kidney disease using
  machine learning techniques}, in: \bibinfo{booktitle}{2016 management and
  innovation technology international conference (MITicon)},
  \bibinfo{organization}{IEEE}. pp. \bibinfo{pages}{MIT--80}.
\bibitem[{Chen et~al.(2014)Chen, Pai, Fujita, Lee, Chen, Chen and Chen}]{t120}
\bibinfo{author}{Chen, C.J.}, \bibinfo{author}{Pai, T.W.},
  \bibinfo{author}{Fujita, H.}, \bibinfo{author}{Lee, C.H.},
  \bibinfo{author}{Chen, Y.T.}, \bibinfo{author}{Chen, K.S.},
  \bibinfo{author}{Chen, Y.C.}, \bibinfo{year}{2014}.
\newblock \bibinfo{title}{Stage diagnosis for chronic kidney disease based on
  ultrasonography}, in: \bibinfo{booktitle}{2014 11th International Conference
  on Fuzzy Systems and Knowledge Discovery (FSKD)},
  \bibinfo{organization}{IEEE}. pp. \bibinfo{pages}{525--530}.
\bibitem[{Chen et~al.(2020)Chen, Wu, Zhang, Zhang, Gong, Zhao, Chen, Huang,
  Yang, Yang et~al.}]{t52}
\bibinfo{author}{Chen, J.}, \bibinfo{author}{Wu, L.}, \bibinfo{author}{Zhang,
  J.}, \bibinfo{author}{Zhang, L.}, \bibinfo{author}{Gong, D.},
  \bibinfo{author}{Zhao, Y.}, \bibinfo{author}{Chen, Q.},
  \bibinfo{author}{Huang, S.}, \bibinfo{author}{Yang, M.},
  \bibinfo{author}{Yang, X.}, et~al., \bibinfo{year}{2020}.
\newblock \bibinfo{title}{Deep learning-based model for detecting 2019 novel
  coronavirus pneumonia on high-resolution computed tomography}.
\newblock \bibinfo{journal}{Scientific reports} \bibinfo{volume}{10},
  \bibinfo{pages}{1--11}.
\bibitem[{Chowdary et~al.(2020)}]{t30}
\bibinfo{author}{Chowdary, G.J.}, et~al., \bibinfo{year}{2020}.
\newblock \bibinfo{title}{Effective prediction of cardiovascular disease using
  cluster of machine learning algorithms}.
\newblock \bibinfo{journal}{Journal of Critical Reviews} \bibinfo{volume}{7},
  \bibinfo{pages}{1865--1875}.
\bibitem[{Cole et~al.(2014)Cole, Roy, De~Luca and Nawab}]{t99}
\bibinfo{author}{Cole, B.T.}, \bibinfo{author}{Roy, S.H.},
  \bibinfo{author}{De~Luca, C.J.}, \bibinfo{author}{Nawab, S.H.},
  \bibinfo{year}{2014}.
\newblock \bibinfo{title}{Dynamical learning and tracking of tremor and
  dyskinesia from wearable sensors}.
\newblock \bibinfo{journal}{IEEE Transactions on Neural Systems and
  Rehabilitation Engineering} \bibinfo{volume}{22}, \bibinfo{pages}{982--991}.
\bibitem[{Das et~al.(2019)Das, Giri, Chourasia and Bala}]{t127}
\bibinfo{author}{Das, A.}, \bibinfo{author}{Giri, R.},
  \bibinfo{author}{Chourasia, G.}, \bibinfo{author}{Bala, A.A.},
  \bibinfo{year}{2019}.
\newblock \bibinfo{title}{Classification of retinal diseases using transfer
  learning approach}, in: \bibinfo{booktitle}{2019 International Conference on
  Communication and Electronics Systems (ICCES)}, \bibinfo{organization}{IEEE}.
  pp. \bibinfo{pages}{2080--2084}.
\bibitem[{Dinesh et~al.(2018)Dinesh, Arumugaraj, Santhosh and Mareeswari}]{t29}
\bibinfo{author}{Dinesh, K.G.}, \bibinfo{author}{Arumugaraj, K.},
  \bibinfo{author}{Santhosh, K.D.}, \bibinfo{author}{Mareeswari, V.},
  \bibinfo{year}{2018}.
\newblock \bibinfo{title}{Prediction of cardiovascular disease using machine
  learning algorithms}, in: \bibinfo{booktitle}{2018 International Conference
  on Current Trends towards Converging Technologies (ICCTCT)},
  \bibinfo{organization}{IEEE}. pp. \bibinfo{pages}{1--7}.
\bibitem[{Dormann et~al.(2013)Dormann, Elith, Bacher, Buchmann, Carl,
  Carr{\'e}, Marqu{\'e}z, Gruber, Lafourcade, Leit{\~a}o et~al.}]{t7}
\bibinfo{author}{Dormann, C.F.}, \bibinfo{author}{Elith, J.},
  \bibinfo{author}{Bacher, S.}, \bibinfo{author}{Buchmann, C.},
  \bibinfo{author}{Carl, G.}, \bibinfo{author}{Carr{\'e}, G.},
  \bibinfo{author}{Marqu{\'e}z, J.R.G.}, \bibinfo{author}{Gruber, B.},
  \bibinfo{author}{Lafourcade, B.}, \bibinfo{author}{Leit{\~a}o, P.J.}, et~al.,
  \bibinfo{year}{2013}.
\newblock \bibinfo{title}{Collinearity: a review of methods to deal with it and
  a simulation study evaluating their performance}.
\newblock \bibinfo{journal}{Ecography} \bibinfo{volume}{36},
  \bibinfo{pages}{27--46}.
\bibitem[{Dutta et~al.(2018)Dutta, Paul and Ghosh}]{t88}
\bibinfo{author}{Dutta, D.}, \bibinfo{author}{Paul, D.},
  \bibinfo{author}{Ghosh, P.}, \bibinfo{year}{2018}.
\newblock \bibinfo{title}{Analysing feature importances for diabetes prediction
  using machine learning}, in: \bibinfo{booktitle}{2018 IEEE 9th Annual
  Information Technology, Electronics and Mobile Communication Conference
  (IEMCON)}, \bibinfo{organization}{IEEE}. pp. \bibinfo{pages}{924--928}.
\bibitem[{Egan(1966)}]{t132}
\bibinfo{author}{Egan, R.L.}, \bibinfo{year}{1966}.
\newblock \bibinfo{title}{Mammography}.
\newblock \bibinfo{journal}{The American journal of nursing} ,
  \bibinfo{pages}{108--111}.
\bibitem[{Eskofier et~al.(2016)Eskofier, Lee, Daneault, Golabchi,
  Ferreira-Carvalho, Vergara-Diaz, Sapienza, Costante, Klucken, Kautz
  et~al.}]{t103}
\bibinfo{author}{Eskofier, B.M.}, \bibinfo{author}{Lee, S.I.},
  \bibinfo{author}{Daneault, J.F.}, \bibinfo{author}{Golabchi, F.N.},
  \bibinfo{author}{Ferreira-Carvalho, G.}, \bibinfo{author}{Vergara-Diaz, G.},
  \bibinfo{author}{Sapienza, S.}, \bibinfo{author}{Costante, G.},
  \bibinfo{author}{Klucken, J.}, \bibinfo{author}{Kautz, T.}, et~al.,
  \bibinfo{year}{2016}.
\newblock \bibinfo{title}{Recent machine learning advancements in sensor-based
  mobility analysis: Deep learning for parkinson's disease assessment}, in:
  \bibinfo{booktitle}{2016 38th Annual International Conference of the IEEE
  Engineering in Medicine and Biology Society (EMBC)},
  \bibinfo{organization}{IEEE}. pp. \bibinfo{pages}{655--658}.
\bibitem[{Froomkin et~al.(2019)Froomkin, Kerr and Pineau}]{t135}
\bibinfo{author}{Froomkin, A.M.}, \bibinfo{author}{Kerr, I.},
  \bibinfo{author}{Pineau, J.}, \bibinfo{year}{2019}.
\newblock \bibinfo{title}{When ais outperform doctors: confronting the
  challenges of a tort-induced over-reliance on machine learning}.
\newblock \bibinfo{journal}{Ariz. L. Rev.} \bibinfo{volume}{61},
  \bibinfo{pages}{33}.
\bibitem[{Gamage et~al.(2018)Gamage, Azad, Taebi, Sandler and Mansy}]{t32}
\bibinfo{author}{Gamage, P.T.}, \bibinfo{author}{Azad, M.K.},
  \bibinfo{author}{Taebi, A.}, \bibinfo{author}{Sandler, R.H.},
  \bibinfo{author}{Mansy, H.A.}, \bibinfo{year}{2018}.
\newblock \bibinfo{title}{Clustering seismocardiographic events using
  unsupervised machine learning}, in: \bibinfo{booktitle}{2018 IEEE Signal
  Processing in Medicine and Biology Symposium (SPMB)},
  \bibinfo{organization}{IEEE}. pp. \bibinfo{pages}{1--5}.
\bibitem[{Garcia-Ceja et~al.(2018)Garcia-Ceja, Riegler, Jakobsen, Torresen,
  Nordgreen, Oedegaard and Fasmer}]{t86}
\bibinfo{author}{Garcia-Ceja, E.}, \bibinfo{author}{Riegler, M.},
  \bibinfo{author}{Jakobsen, P.}, \bibinfo{author}{Torresen, J.},
  \bibinfo{author}{Nordgreen, T.}, \bibinfo{author}{Oedegaard, K.J.},
  \bibinfo{author}{Fasmer, O.B.}, \bibinfo{year}{2018}.
\newblock \bibinfo{title}{Motor activity based classification of depression in
  unipolar and bipolar patients}, in: \bibinfo{booktitle}{2018 IEEE 31st
  International Symposium on Computer-Based Medical Systems (CBMS)},
  \bibinfo{organization}{IEEE}. pp. \bibinfo{pages}{316--321}.
\bibitem[{Gerard et~al.(2018)Gerard, Patton, Christensen, Bayouth and
  Reinhardt}]{t40}
\bibinfo{author}{Gerard, S.E.}, \bibinfo{author}{Patton, T.J.},
  \bibinfo{author}{Christensen, G.E.}, \bibinfo{author}{Bayouth, J.E.},
  \bibinfo{author}{Reinhardt, J.M.}, \bibinfo{year}{2018}.
\newblock \bibinfo{title}{Fissurenet: a deep learning approach for pulmonary
  fissure detection in ct images}.
\newblock \bibinfo{journal}{IEEE transactions on medical imaging}
  \bibinfo{volume}{38}, \bibinfo{pages}{156--166}.
\bibitem[{Godkhindi and Gowda(2017)}]{t69}
\bibinfo{author}{Godkhindi, A.M.}, \bibinfo{author}{Gowda, R.M.},
  \bibinfo{year}{2017}.
\newblock \bibinfo{title}{Automated detection of polyps in ct colonography
  images using deep learning algorithms in colon cancer diagnosis}, in:
  \bibinfo{booktitle}{2017 International Conference on Energy, Communication,
  Data Analytics and Soft Computing (ICECDS)}, \bibinfo{organization}{IEEE}.
  pp. \bibinfo{pages}{1722--1728}.
\bibitem[{Grewal et~al.(2018)Grewal, Oloumi, Rubin and Tennant}]{t122}
\bibinfo{author}{Grewal, P.S.}, \bibinfo{author}{Oloumi, F.},
  \bibinfo{author}{Rubin, U.}, \bibinfo{author}{Tennant, M.T.},
  \bibinfo{year}{2018}.
\newblock \bibinfo{title}{Deep learning in ophthalmology: a review}.
\newblock \bibinfo{journal}{Canadian Journal of Ophthalmology}
  \bibinfo{volume}{53}, \bibinfo{pages}{309--313}.
\bibitem[{Hashem et~al.(2017)Hashem, Esmat, Elakel, Habashy, Raouf, Elhefnawi,
  Eladawy and ElHefnawi}]{t115}
\bibinfo{author}{Hashem, S.}, \bibinfo{author}{Esmat, G.},
  \bibinfo{author}{Elakel, W.}, \bibinfo{author}{Habashy, S.},
  \bibinfo{author}{Raouf, S.A.}, \bibinfo{author}{Elhefnawi, M.},
  \bibinfo{author}{Eladawy, M.I.}, \bibinfo{author}{ElHefnawi, M.},
  \bibinfo{year}{2017}.
\newblock \bibinfo{title}{Comparison of machine learning approaches for
  prediction of advanced liver fibrosis in chronic hepatitis c patients}.
\newblock \bibinfo{journal}{IEEE/ACM transactions on computational biology and
  bioinformatics} \bibinfo{volume}{15}, \bibinfo{pages}{861--868}.
\bibitem[{Hegde et~al.(2018)Hegde, Shenoy and Shekar}]{t111}
\bibinfo{author}{Hegde, P.R.}, \bibinfo{author}{Shenoy, M.M.},
  \bibinfo{author}{Shekar, B.}, \bibinfo{year}{2018}.
\newblock \bibinfo{title}{Comparison of machine learning algorithms for skin
  disease classification using color and texture features}, in:
  \bibinfo{booktitle}{2018 International Conference on Advances in Computing,
  Communications and Informatics (ICACCI)}, \bibinfo{organization}{IEEE}. pp.
  \bibinfo{pages}{1825--1828}.
\bibitem[{Hemdan et~al.(2020)Hemdan, Shouman and Karar}]{t41}
\bibinfo{author}{Hemdan, E.E.D.}, \bibinfo{author}{Shouman, M.A.},
  \bibinfo{author}{Karar, M.E.}, \bibinfo{year}{2020}.
\newblock \bibinfo{title}{Covidx-net: A framework of deep learning classifiers
  to diagnose covid-19 in x-ray images}.
\newblock \bibinfo{journal}{arXiv preprint arXiv:2003.11055} .
\bibitem[{Heraz et~al.(2007)Heraz, Razaki and Frasson}]{t87}
\bibinfo{author}{Heraz, A.}, \bibinfo{author}{Razaki, R.},
  \bibinfo{author}{Frasson, C.}, \bibinfo{year}{2007}.
\newblock \bibinfo{title}{Using machine learning to predict learner emotional
  state from brainwaves}, in: \bibinfo{booktitle}{Seventh IEEE International
  Conference on Advanced Learning Technologies (ICALT 2007)},
  \bibinfo{organization}{IEEE}. pp. \bibinfo{pages}{853--857}.
\bibitem[{Hijazi et~al.(2016)Hijazi, Page, Kantarci and Soyata}]{t24}
\bibinfo{author}{Hijazi, S.}, \bibinfo{author}{Page, A.},
  \bibinfo{author}{Kantarci, B.}, \bibinfo{author}{Soyata, T.},
  \bibinfo{year}{2016}.
\newblock \bibinfo{title}{Machine learning in cardiac health monitoring and
  decision support}.
\newblock \bibinfo{journal}{Computer} \bibinfo{volume}{49},
  \bibinfo{pages}{38--48}.
\bibitem[{Inoue et~al.(2018)Inoue, Yoshioka, Yagi, Nagami and Oku}]{t77}
\bibinfo{author}{Inoue, K.}, \bibinfo{author}{Yoshioka, M.},
  \bibinfo{author}{Yagi, N.}, \bibinfo{author}{Nagami, S.},
  \bibinfo{author}{Oku, Y.}, \bibinfo{year}{2018}.
\newblock \bibinfo{title}{Using machine learning and a combination of
  respiratory flow, laryngeal motion, and swallowing sounds to classify safe
  and unsafe swallowing}.
\newblock \bibinfo{journal}{IEEE Transactions on Biomedical Engineering}
  \bibinfo{volume}{65}, \bibinfo{pages}{2529--2541}.
\bibitem[{Joseph et~al.(2017a)Joseph, Hijal, Kildea, Hendren and Herrera}]{t73}
\bibinfo{author}{Joseph, A.}, \bibinfo{author}{Hijal, T.},
  \bibinfo{author}{Kildea, J.}, \bibinfo{author}{Hendren, L.},
  \bibinfo{author}{Herrera, D.}, \bibinfo{year}{2017}a.
\newblock \bibinfo{title}{Predicting waiting times in radiation oncology using
  machine learning}, in: \bibinfo{booktitle}{2017 16th IEEE International
  Conference on Machine Learning and Applications (ICMLA)},
  \bibinfo{organization}{IEEE}. pp. \bibinfo{pages}{1024--1029}.
\bibitem[{Joseph et~al.(2017b)Joseph, Sanghani and Ren}]{t106}
\bibinfo{author}{Joseph, N.}, \bibinfo{author}{Sanghani, P.},
  \bibinfo{author}{Ren, H.}, \bibinfo{year}{2017}b.
\newblock \bibinfo{title}{Semi-automated segmentation of glioblastomas in brain
  mri using machine learning techniques}, in: \bibinfo{booktitle}{2017 16th
  IEEE International Conference on Machine Learning and Applications (ICMLA)},
  \bibinfo{organization}{IEEE}. pp. \bibinfo{pages}{1149--1152}.
\bibitem[{Kaburlasos et~al.(1999)Kaburlasos, Petridis, Brett and Baker}]{t21}
\bibinfo{author}{Kaburlasos, V.G.}, \bibinfo{author}{Petridis, V.},
  \bibinfo{author}{Brett, P.N.}, \bibinfo{author}{Baker, D.A.},
  \bibinfo{year}{1999}.
\newblock \bibinfo{title}{Estimation of the stapes-bone thickness in the
  stapedotomy surgical procedure using a machine-learning technique}.
\newblock \bibinfo{journal}{IEEE Transactions on Information Technology in
  Biomedicine} \bibinfo{volume}{3}, \bibinfo{pages}{268--277}.
\bibitem[{Khalaf et~al.(2017)Khalaf, Hussain, Keight, Al-Jumeily, Keenan,
  Chalmers, Fergus, Salih, Abd and Idowu}]{t11}
\bibinfo{author}{Khalaf, M.}, \bibinfo{author}{Hussain, A.J.},
  \bibinfo{author}{Keight, R.}, \bibinfo{author}{Al-Jumeily, D.},
  \bibinfo{author}{Keenan, R.}, \bibinfo{author}{Chalmers, C.},
  \bibinfo{author}{Fergus, P.}, \bibinfo{author}{Salih, W.},
  \bibinfo{author}{Abd, D.H.}, \bibinfo{author}{Idowu, I.O.},
  \bibinfo{year}{2017}.
\newblock \bibinfo{title}{Recurrent neural network architectures for analysing
  biomedical data sets}, in: \bibinfo{booktitle}{2017 10th International
  Conference on Developments in eSystems Engineering (DeSE)},
  \bibinfo{organization}{IEEE}. pp. \bibinfo{pages}{232--237}.
\bibitem[{Khalaf et~al.(2016a)Khalaf, Hussain, Keight, Al-Jumeily, Keenan,
  Fergus and Idowu}]{t18}
\bibinfo{author}{Khalaf, M.}, \bibinfo{author}{Hussain, A.J.},
  \bibinfo{author}{Keight, R.}, \bibinfo{author}{Al-Jumeily, D.},
  \bibinfo{author}{Keenan, R.}, \bibinfo{author}{Fergus, P.},
  \bibinfo{author}{Idowu, I.O.}, \bibinfo{year}{2016}a.
\newblock \bibinfo{title}{The utilisation of composite machine learning models
  for the classification of medical datasets for sickle cell disease}, in:
  \bibinfo{booktitle}{2016 Sixth International Conference on Digital
  Information Processing and Communications (ICDIPC)},
  \bibinfo{organization}{IEEE}. pp. \bibinfo{pages}{37--41}.
\bibitem[{Khalaf et~al.(2016b)Khalaf, Hussain, Keight, Al-Jumeily, Keenan,
  Fergus and Idowu}]{t130}
\bibinfo{author}{Khalaf, M.}, \bibinfo{author}{Hussain, A.J.},
  \bibinfo{author}{Keight, R.}, \bibinfo{author}{Al-Jumeily, D.},
  \bibinfo{author}{Keenan, R.}, \bibinfo{author}{Fergus, P.},
  \bibinfo{author}{Idowu, I.O.}, \bibinfo{year}{2016}b.
\newblock \bibinfo{title}{The utilisation of composite machine learning models
  for the classification of medical datasets for sickle cell disease}, in:
  \bibinfo{booktitle}{2016 Sixth International Conference on Digital
  Information Processing and Communications (ICDIPC)},
  \bibinfo{organization}{IEEE}. pp. \bibinfo{pages}{37--41}.
\bibitem[{Kim et~al.(2013)Kim, Yoo and Kim}]{t56}
\bibinfo{author}{Kim, S.K.}, \bibinfo{author}{Yoo, T.K.}, \bibinfo{author}{Kim,
  D.W.}, \bibinfo{year}{2013}.
\newblock \bibinfo{title}{Osteoporosis risk prediction using machine learning
  and conventional methods}, in: \bibinfo{booktitle}{2013 35th Annual
  International Conference of the IEEE Engineering in Medicine and Biology
  Society (EMBC)}, \bibinfo{organization}{IEEE}. pp. \bibinfo{pages}{188--191}.
\bibitem[{Kobashi et~al.(2016)Kobashi, Hossain, Nii, Kambara, Morooka, Okuno
  and Yoshiya}]{t19}
\bibinfo{author}{Kobashi, S.}, \bibinfo{author}{Hossain, B.},
  \bibinfo{author}{Nii, M.}, \bibinfo{author}{Kambara, S.},
  \bibinfo{author}{Morooka, T.}, \bibinfo{author}{Okuno, M.},
  \bibinfo{author}{Yoshiya, S.}, \bibinfo{year}{2016}.
\newblock \bibinfo{title}{Prediction of post-operative implanted knee function
  using machine learning in clinical big data}, in: \bibinfo{booktitle}{2016
  International Conference on Machine Learning and Cybernetics (ICMLC)},
  \bibinfo{organization}{IEEE}. pp. \bibinfo{pages}{195--200}.
\bibitem[{Kumar et~al.(2016)Kumar, Kumar and Saboo}]{t108}
\bibinfo{author}{Kumar, V.B.}, \bibinfo{author}{Kumar, S.S.},
  \bibinfo{author}{Saboo, V.}, \bibinfo{year}{2016}.
\newblock \bibinfo{title}{Dermatological disease detection using image
  processing and machine learning}, in: \bibinfo{booktitle}{2016 Third
  International Conference on Artificial Intelligence and Pattern Recognition
  (AIPR)}, \bibinfo{organization}{IEEE}. pp. \bibinfo{pages}{1--6}.
\bibitem[{Lahijanian et~al.(2016)Lahijanian, Farahani and Zarandi}]{t112}
\bibinfo{author}{Lahijanian, B.}, \bibinfo{author}{Farahani, F.V.},
  \bibinfo{author}{Zarandi, M.F.}, \bibinfo{year}{2016}.
\newblock \bibinfo{title}{A new multiple classifier system for diagnosis of
  erythemato-squamous diseases based on rough set feature selection}, in:
  \bibinfo{booktitle}{2016 IEEE International Conference on Fuzzy Systems
  (FUZZ-IEEE)}, \bibinfo{organization}{IEEE}. pp. \bibinfo{pages}{2309--2316}.
\bibitem[{Lambrou et~al.(2009)Lambrou, Papadopoulos and Gammerman}]{t68}
\bibinfo{author}{Lambrou, A.}, \bibinfo{author}{Papadopoulos, H.},
  \bibinfo{author}{Gammerman, A.}, \bibinfo{year}{2009}.
\newblock \bibinfo{title}{Evolutionary conformal prediction for breast cancer
  diagnosis}, in: \bibinfo{booktitle}{2009 9th international conference on
  information technology and applications in biomedicine},
  \bibinfo{organization}{IEEE}. pp. \bibinfo{pages}{1--4}.
\bibitem[{Lauzon(2012)}]{t8}
\bibinfo{author}{Lauzon, F.Q.}, \bibinfo{year}{2012}.
\newblock \bibinfo{title}{An introduction to deep learning}, in:
  \bibinfo{booktitle}{2012 11th International Conference on Information
  Science, Signal Processing and their Applications (ISSPA)},
  \bibinfo{organization}{IEEE}. pp. \bibinfo{pages}{1438--1439}.
\bibitem[{Liang et~al.(2016)Liang, Powell, Ersoy, Poostchi, Silamut,
  Palaniappan, Guo, Hossain, Sameer, Maude et~al.}]{t78}
\bibinfo{author}{Liang, Z.}, \bibinfo{author}{Powell, A.},
  \bibinfo{author}{Ersoy, I.}, \bibinfo{author}{Poostchi, M.},
  \bibinfo{author}{Silamut, K.}, \bibinfo{author}{Palaniappan, K.},
  \bibinfo{author}{Guo, P.}, \bibinfo{author}{Hossain, M.A.},
  \bibinfo{author}{Sameer, A.}, \bibinfo{author}{Maude, R.J.}, et~al.,
  \bibinfo{year}{2016}.
\newblock \bibinfo{title}{Cnn-based image analysis for malaria diagnosis}, in:
  \bibinfo{booktitle}{2016 IEEE international conference on bioinformatics and
  biomedicine (BIBM)}, \bibinfo{organization}{IEEE}. pp.
  \bibinfo{pages}{493--496}.
\bibitem[{Lin et~al.(2008)Lin, Cervino, Tang, Vasconcelos and Jiang}]{t59}
\bibinfo{author}{Lin, T.}, \bibinfo{author}{Cervino, L.},
  \bibinfo{author}{Tang, X.}, \bibinfo{author}{Vasconcelos, N.},
  \bibinfo{author}{Jiang, S.B.}, \bibinfo{year}{2008}.
\newblock \bibinfo{title}{Tumor targeting for lung cancer radiotherapy using
  machine learning techniques}, in: \bibinfo{booktitle}{2008 Seventh
  International Conference on Machine Learning and Applications},
  \bibinfo{organization}{IEEE}. pp. \bibinfo{pages}{533--538}.
\bibitem[{Liu et~al.(2017)Liu, Liu, Lee, Weissman, Posner, Cha and Yoo}]{t83}
\bibinfo{author}{Liu, A.}, \bibinfo{author}{Liu, B.}, \bibinfo{author}{Lee,
  D.}, \bibinfo{author}{Weissman, M.}, \bibinfo{author}{Posner, J.},
  \bibinfo{author}{Cha, J.}, \bibinfo{author}{Yoo, S.}, \bibinfo{year}{2017}.
\newblock \bibinfo{title}{Machine learning aided prediction of family history
  of depression}, in: \bibinfo{booktitle}{2017 New York Scientific Data Summit
  (NYSDS)}, \bibinfo{organization}{IEEE}. pp. \bibinfo{pages}{1--4}.
\bibitem[{Loh(2018)}]{t134}
\bibinfo{author}{Loh, E.}, \bibinfo{year}{2018}.
\newblock \bibinfo{title}{Medicine and the rise of the robots: a qualitative
  review of recent advances of artificial intelligence in health}.
\newblock \bibinfo{journal}{BMJ Leader} , \bibinfo{pages}{leader--2018}.
\bibitem[{Louis et~al.(2017)Louis, Alosco, Rowland, Liao, Wang, Koerte,
  Shenton, Stern, Joshi and Lin}]{t107}
\bibinfo{author}{Louis, M.S.}, \bibinfo{author}{Alosco, M.},
  \bibinfo{author}{Rowland, B.}, \bibinfo{author}{Liao, H.},
  \bibinfo{author}{Wang, J.}, \bibinfo{author}{Koerte, I.},
  \bibinfo{author}{Shenton, M.}, \bibinfo{author}{Stern, R.},
  \bibinfo{author}{Joshi, A.}, \bibinfo{author}{Lin, A.P.},
  \bibinfo{year}{2017}.
\newblock \bibinfo{title}{Using machine learning techniques for identification
  of chronic traumatic encephalopathy related spectroscopic biomarkers}, in:
  \bibinfo{booktitle}{2017 IEEE Applied Imagery Pattern Recognition Workshop
  (AIPR)}, \bibinfo{organization}{IEEE}. pp. \bibinfo{pages}{1--5}.
\bibitem[{Lu et~al.(2017)Lu, Lu, Hou, Cheng and Wang}]{t98}
\bibinfo{author}{Lu, S.}, \bibinfo{author}{Lu, Z.}, \bibinfo{author}{Hou, X.},
  \bibinfo{author}{Cheng, H.}, \bibinfo{author}{Wang, S.},
  \bibinfo{year}{2017}.
\newblock \bibinfo{title}{Detection of cerebral microbleeding based on deep
  convolutional neural network}, in: \bibinfo{booktitle}{2017 14th
  International Computer Conference on Wavelet Active Media Technology and
  Information Processing (ICCWAMTIP)}, \bibinfo{organization}{IEEE}. pp.
  \bibinfo{pages}{93--96}.
\bibitem[{Luo et~al.(2018)Luo, Yang, Dai and Liu}]{t20}
\bibinfo{author}{Luo, J.}, \bibinfo{author}{Yang, C.}, \bibinfo{author}{Dai,
  S.L.}, \bibinfo{author}{Liu, Z.}, \bibinfo{year}{2018}.
\newblock \bibinfo{title}{Tremor attenuation for surgical robots using support
  vector machine with parameters optimization}, in: \bibinfo{booktitle}{2018
  Tenth International Conference on Advanced Computational Intelligence
  (ICACI)}, \bibinfo{organization}{IEEE}. pp. \bibinfo{pages}{667--672}.
\bibitem[{Mahdy et~al.(2020)Mahdy, Ezzat, Elmousalami, Ella and
  Hassanien}]{t51}
\bibinfo{author}{Mahdy, L.N.}, \bibinfo{author}{Ezzat, K.A.},
  \bibinfo{author}{Elmousalami, H.H.}, \bibinfo{author}{Ella, H.A.},
  \bibinfo{author}{Hassanien, A.E.}, \bibinfo{year}{2020}.
\newblock \bibinfo{title}{Automatic x-ray covid-19 lung image classification
  system based on multi-level thresholding and support vector machine}.
\newblock \bibinfo{journal}{MedRxiv} , \bibinfo{pages}{2020--03}.
\bibitem[{Malik et~al.(2019)Malik, Kanwal, Asghar, Sadiq, Karamat and
  Fleury}]{t124}
\bibinfo{author}{Malik, S.}, \bibinfo{author}{Kanwal, N.},
  \bibinfo{author}{Asghar, M.N.}, \bibinfo{author}{Sadiq, M.A.A.},
  \bibinfo{author}{Karamat, I.}, \bibinfo{author}{Fleury, M.},
  \bibinfo{year}{2019}.
\newblock \bibinfo{title}{Data driven approach for eye disease classification
  with machine learning}.
\newblock \bibinfo{journal}{Applied Sciences} \bibinfo{volume}{9},
  \bibinfo{pages}{2789}.
\bibitem[{Martinez-Manzanera et~al.(2015)Martinez-Manzanera, Roosma, Beudel,
  Borgemeester, van Laar and Maurits}]{t104}
\bibinfo{author}{Martinez-Manzanera, O.}, \bibinfo{author}{Roosma, E.},
  \bibinfo{author}{Beudel, M.}, \bibinfo{author}{Borgemeester, R.},
  \bibinfo{author}{van Laar, T.}, \bibinfo{author}{Maurits, N.M.},
  \bibinfo{year}{2015}.
\newblock \bibinfo{title}{A method for automatic and objective scoring of
  bradykinesia using orientation sensors and classification algorithms}.
\newblock \bibinfo{journal}{IEEE Transactions on Biomedical Engineering}
  \bibinfo{volume}{63}, \bibinfo{pages}{1016--1024}.
\bibitem[{Maysanjaya et~al.(2015)Maysanjaya, Nugroho and Setiawan}]{t93}
\bibinfo{author}{Maysanjaya, I.M.D.}, \bibinfo{author}{Nugroho, H.A.},
  \bibinfo{author}{Setiawan, N.A.}, \bibinfo{year}{2015}.
\newblock \bibinfo{title}{A comparison of classification methods on diagnosis
  of thyroid diseases}, in: \bibinfo{booktitle}{2015 International seminar on
  intelligent technology and its applications (ISITIA)},
  \bibinfo{organization}{IEEE}. pp. \bibinfo{pages}{89--92}.
\bibitem[{Milo{\v{s}}evi{\'c} et~al.(2014)Milo{\v{s}}evi{\'c}, Van~de Vel,
  Bonroy, Ceulemans, Lagae, Vanrumste and Van~Huffel}]{t102}
\bibinfo{author}{Milo{\v{s}}evi{\'c}, M.}, \bibinfo{author}{Van~de Vel, A.},
  \bibinfo{author}{Bonroy, B.}, \bibinfo{author}{Ceulemans, B.},
  \bibinfo{author}{Lagae, L.}, \bibinfo{author}{Vanrumste, B.},
  \bibinfo{author}{Van~Huffel, S.}, \bibinfo{year}{2014}.
\newblock \bibinfo{title}{Detection of epileptic convulsions from accelerometry
  signals through machine learning approach}, in: \bibinfo{booktitle}{2014 IEEE
  International Workshop on Machine Learning for Signal Processing (MLSP)},
  \bibinfo{organization}{IEEE}. pp. \bibinfo{pages}{1--6}.
\bibitem[{Morita et~al.(2017)Morita, Tashita, Nii and Kobashi}]{t57}
\bibinfo{author}{Morita, K.}, \bibinfo{author}{Tashita, A.},
  \bibinfo{author}{Nii, M.}, \bibinfo{author}{Kobashi, S.},
  \bibinfo{year}{2017}.
\newblock \bibinfo{title}{Computer-aided diagnosis system for rheumatoid
  arthritis using machine learning}, in: \bibinfo{booktitle}{2017 International
  Conference on Machine Learning and Cybernetics (ICMLC)},
  \bibinfo{organization}{IEEE}. pp. \bibinfo{pages}{357--360}.
\bibitem[{Muhamedyev et~al.(2015)Muhamedyev, Yakunin, Iskakov, Sainova,
  Abdilmanova and Kuchin}]{t3}
\bibinfo{author}{Muhamedyev, R.}, \bibinfo{author}{Yakunin, K.},
  \bibinfo{author}{Iskakov, S.}, \bibinfo{author}{Sainova, S.},
  \bibinfo{author}{Abdilmanova, A.}, \bibinfo{author}{Kuchin, Y.},
  \bibinfo{year}{2015}.
\newblock \bibinfo{title}{Comparative analysis of classification algorithms},
  in: \bibinfo{booktitle}{2015 9th International Conference on Application of
  Information and Communication Technologies (AICT)},
  \bibinfo{organization}{IEEE}. pp. \bibinfo{pages}{96--101}.
\bibitem[{Mulyani et~al.(2016)Mulyani, Rahman, Riza et~al.}]{t75}
\bibinfo{author}{Mulyani, Y.}, \bibinfo{author}{Rahman, E.F.},
  \bibinfo{author}{Riza, L.S.}, et~al., \bibinfo{year}{2016}.
\newblock \bibinfo{title}{A new approach on prediction of fever disease by
  using a combination of dempster shafer and na{\"\i}ve bayes}, in:
  \bibinfo{booktitle}{2016 2nd International Conference on Science in
  Information Technology (ICSITech)}, \bibinfo{organization}{IEEE}. pp.
  \bibinfo{pages}{367--371}.
\bibitem[{Najar et~al.(2018)Najar, Irawan and Adzkiya}]{t76}
\bibinfo{author}{Najar, A.M.}, \bibinfo{author}{Irawan, M.I.},
  \bibinfo{author}{Adzkiya, D.}, \bibinfo{year}{2018}.
\newblock \bibinfo{title}{Extreme learning machine method for dengue
  hemorrhagic fever outbreak risk level prediction}, in:
  \bibinfo{booktitle}{2018 International Conference on Smart Computing and
  Electronic Enterprise (ICSCEE)}, \bibinfo{organization}{IEEE}. pp.
  \bibinfo{pages}{1--5}.
\bibitem[{Napolitano et~al.(2016)Napolitano, Marshall, Hamilton and
  Gavin}]{t23}
\bibinfo{author}{Napolitano, G.}, \bibinfo{author}{Marshall, A.},
  \bibinfo{author}{Hamilton, P.}, \bibinfo{author}{Gavin, A.T.},
  \bibinfo{year}{2016}.
\newblock \bibinfo{title}{Machine learning classification of surgical pathology
  reports and chunk recognition for information extraction noise reduction}.
\newblock \bibinfo{journal}{Artificial intelligence in medicine}
  \bibinfo{volume}{70}, \bibinfo{pages}{77--83}.
\bibitem[{Narayanan et~al.(2017)Narayanan, Unnikrishnan, Paul and Joseph}]{t6}
\bibinfo{author}{Narayanan, U.}, \bibinfo{author}{Unnikrishnan, A.},
  \bibinfo{author}{Paul, V.}, \bibinfo{author}{Joseph, S.},
  \bibinfo{year}{2017}.
\newblock \bibinfo{title}{A survey on various supervised classification
  algorithms}, in: \bibinfo{booktitle}{2017 International Conference on Energy,
  Communication, Data Analytics and Soft Computing (ICECDS)},
  \bibinfo{organization}{IEEE}. pp. \bibinfo{pages}{2118--2124}.
\bibitem[{Narin et~al.(2021)Narin, Kaya and Pamuk}]{t44}
\bibinfo{author}{Narin, A.}, \bibinfo{author}{Kaya, C.},
  \bibinfo{author}{Pamuk, Z.}, \bibinfo{year}{2021}.
\newblock \bibinfo{title}{Automatic detection of coronavirus disease (covid-19)
  using x-ray images and deep convolutional neural networks}.
\newblock \bibinfo{journal}{Pattern Analysis and Applications} ,
  \bibinfo{pages}{1--14}.
\bibitem[{Nasr et~al.(2017)Nasr, El-Bahnasy, Hamdy and Kamal}]{t116}
\bibinfo{author}{Nasr, M.}, \bibinfo{author}{El-Bahnasy, K.},
  \bibinfo{author}{Hamdy, M.}, \bibinfo{author}{Kamal, S.M.},
  \bibinfo{year}{2017}.
\newblock \bibinfo{title}{A novel model based on non invasive methods for
  prediction of liver fibrosis}, in: \bibinfo{booktitle}{2017 13th
  International Computer Engineering Conference (ICENCO)},
  \bibinfo{organization}{IEEE}. pp. \bibinfo{pages}{276--281}.
\bibitem[{Nguyen and Patrick(2014)}]{t62}
\bibinfo{author}{Nguyen, D.H.}, \bibinfo{author}{Patrick, J.D.},
  \bibinfo{year}{2014}.
\newblock \bibinfo{title}{Supervised machine learning and active learning in
  classification of radiology reports}.
\newblock \bibinfo{journal}{Journal of the American Medical Informatics
  Association} \bibinfo{volume}{21}, \bibinfo{pages}{893--901}.
\bibitem[{Nimkar and Kubal(2018)}]{t96}
\bibinfo{author}{Nimkar, A.V.}, \bibinfo{author}{Kubal, D.R.},
  \bibinfo{year}{2018}.
\newblock \bibinfo{title}{Optimization of schizophrenia diagnosis prediction
  using machine learning techniques}, in: \bibinfo{booktitle}{2018 4th
  International Conference on Computer and Information Sciences (ICCOINS)},
  \bibinfo{organization}{IEEE}. pp. \bibinfo{pages}{1--6}.
\bibitem[{Obulesu et~al.(2018)Obulesu, Mahendra and ThrilokReddy}]{t5}
\bibinfo{author}{Obulesu, O.}, \bibinfo{author}{Mahendra, M.},
  \bibinfo{author}{ThrilokReddy, M.}, \bibinfo{year}{2018}.
\newblock \bibinfo{title}{Machine learning techniques and tools: A survey}, in:
  \bibinfo{booktitle}{2018 International Conference on Inventive Research in
  Computing Applications (ICIRCA)}, \bibinfo{organization}{IEEE}. pp.
  \bibinfo{pages}{605--611}.
\bibitem[{Olczak et~al.(2017)Olczak, Fahlberg, Maki, Razavian, Jilert, Stark,
  Sk{\"o}ldenberg and Gordon}]{t65}
\bibinfo{author}{Olczak, J.}, \bibinfo{author}{Fahlberg, N.},
  \bibinfo{author}{Maki, A.}, \bibinfo{author}{Razavian, A.S.},
  \bibinfo{author}{Jilert, A.}, \bibinfo{author}{Stark, A.},
  \bibinfo{author}{Sk{\"o}ldenberg, O.}, \bibinfo{author}{Gordon, M.},
  \bibinfo{year}{2017}.
\newblock \bibinfo{title}{Artificial intelligence for analyzing orthopedic
  trauma radiographs: deep learning algorithms—are they on par with humans
  for diagnosing fractures?}
\newblock \bibinfo{journal}{Acta orthopaedica} \bibinfo{volume}{88},
  \bibinfo{pages}{581--586}.
\bibitem[{Ongun et~al.(2001a)Ongun, Halici, Leblebicioglu, Atalay, Beksac and
  Beksac}]{t16}
\bibinfo{author}{Ongun, G.}, \bibinfo{author}{Halici, U.},
  \bibinfo{author}{Leblebicioglu, K.}, \bibinfo{author}{Atalay, V.},
  \bibinfo{author}{Beksac, M.}, \bibinfo{author}{Beksac, S.},
  \bibinfo{year}{2001}a.
\newblock \bibinfo{title}{An automated differential blood count system}, in:
  \bibinfo{booktitle}{2001 Conference Proceedings of the 23rd Annual
  International Conference of the IEEE Engineering in Medicine and Biology
  Society}, \bibinfo{organization}{IEEE}. pp. \bibinfo{pages}{2583--2586}.
\bibitem[{Ongun et~al.(2001b)Ongun, Halici, Leblebicioglu, Atalay, Beksac and
  Beksac}]{t128}
\bibinfo{author}{Ongun, G.}, \bibinfo{author}{Halici, U.},
  \bibinfo{author}{Leblebicioglu, K.}, \bibinfo{author}{Atalay, V.},
  \bibinfo{author}{Beksac, M.}, \bibinfo{author}{Beksac, S.},
  \bibinfo{year}{2001}b.
\newblock \bibinfo{title}{An automated differential blood count system}, in:
  \bibinfo{booktitle}{2001 Conference Proceedings of the 23rd Annual
  International Conference of the IEEE Engineering in Medicine and Biology
  Society}, \bibinfo{organization}{IEEE}. pp. \bibinfo{pages}{2583--2586}.
\bibitem[{Palaniappan et~al.(2014)Palaniappan, Sundaraj and Sundaraj}]{t37}
\bibinfo{author}{Palaniappan, R.}, \bibinfo{author}{Sundaraj, K.},
  \bibinfo{author}{Sundaraj, S.}, \bibinfo{year}{2014}.
\newblock \bibinfo{title}{A comparative study of the svm and k-nn machine
  learning algorithms for the diagnosis of respiratory pathologies using
  pulmonary acoustic signals}.
\newblock \bibinfo{journal}{BMC bioinformatics} \bibinfo{volume}{15},
  \bibinfo{pages}{1--8}.
\bibitem[{Pandit and Banday(2020)}]{to53}
\bibinfo{author}{Pandit, M.K.}, \bibinfo{author}{Banday, S.A.},
  \bibinfo{year}{2020}.
\newblock \bibinfo{title}{Sars n-cov2-19 detection from chest x-ray images
  using deep neural networks}.
\newblock \bibinfo{journal}{International Journal of Pervasive Computing and
  Communications} .
\bibitem[{Panicker and Gayathri(2019)}]{t85}
\bibinfo{author}{Panicker, S.S.}, \bibinfo{author}{Gayathri, P.},
  \bibinfo{year}{2019}.
\newblock \bibinfo{title}{A survey of machine learning techniques in physiology
  based mental stress detection systems}.
\newblock \bibinfo{journal}{Biocybernetics and Biomedical Engineering}
  \bibinfo{volume}{39}, \bibinfo{pages}{444--469}.
\bibitem[{Papavasileiou et~al.(2017)Papavasileiou, Zhang, Wang, Bi, Zhang and
  Han}]{t105}
\bibinfo{author}{Papavasileiou, I.}, \bibinfo{author}{Zhang, W.},
  \bibinfo{author}{Wang, X.}, \bibinfo{author}{Bi, J.}, \bibinfo{author}{Zhang,
  L.}, \bibinfo{author}{Han, S.}, \bibinfo{year}{2017}.
\newblock \bibinfo{title}{Classification of neurological gait disorders using
  multi-task feature learning}, in: \bibinfo{booktitle}{2017 IEEE/ACM
  International Conference on Connected Health: Applications, Systems and
  Engineering Technologies (CHASE)}, \bibinfo{organization}{IEEE}. pp.
  \bibinfo{pages}{195--204}.
\bibitem[{Pauly et~al.(2015)Pauly, Diotte, Fallavollita, Weidert, Euler and
  Navab}]{t22}
\bibinfo{author}{Pauly, O.}, \bibinfo{author}{Diotte, B.},
  \bibinfo{author}{Fallavollita, P.}, \bibinfo{author}{Weidert, S.},
  \bibinfo{author}{Euler, E.}, \bibinfo{author}{Navab, N.},
  \bibinfo{year}{2015}.
\newblock \bibinfo{title}{Machine learning-based augmented reality for improved
  surgical scene understanding}.
\newblock \bibinfo{journal}{Computerized Medical Imaging and Graphics}
  \bibinfo{volume}{41}, \bibinfo{pages}{55--60}.
\bibitem[{Peter et~al.(2012a)Peter, Brieu, Jansen, Smethurst, Ouwehand and
  Navab}]{t17}
\bibinfo{author}{Peter, L.}, \bibinfo{author}{Brieu, N.},
  \bibinfo{author}{Jansen, S.}, \bibinfo{author}{Smethurst, P.A.},
  \bibinfo{author}{Ouwehand, W.H.}, \bibinfo{author}{Navab, N.},
  \bibinfo{year}{2012}a.
\newblock \bibinfo{title}{Automatic segmentation and tracking of thrombus
  formation within in vitro microscopic video sequences}, in:
  \bibinfo{booktitle}{2012 9th IEEE International Symposium on Biomedical
  Imaging (ISBI)}, \bibinfo{organization}{IEEE}. pp.
  \bibinfo{pages}{1635--1638}.
\bibitem[{Peter et~al.(2012b)Peter, Brieu, Jansen, Smethurst, Ouwehand and
  Navab}]{t129}
\bibinfo{author}{Peter, L.}, \bibinfo{author}{Brieu, N.},
  \bibinfo{author}{Jansen, S.}, \bibinfo{author}{Smethurst, P.A.},
  \bibinfo{author}{Ouwehand, W.H.}, \bibinfo{author}{Navab, N.},
  \bibinfo{year}{2012}b.
\newblock \bibinfo{title}{Automatic segmentation and tracking of thrombus
  formation within in vitro microscopic video sequences}, in:
  \bibinfo{booktitle}{2012 9th IEEE International Symposium on Biomedical
  Imaging (ISBI)}, \bibinfo{organization}{IEEE}. pp.
  \bibinfo{pages}{1635--1638}.
\bibitem[{Prajapati et~al.(2017)Prajapati, Nagaraj and Mitra}]{t13}
\bibinfo{author}{Prajapati, S.A.}, \bibinfo{author}{Nagaraj, R.},
  \bibinfo{author}{Mitra, S.}, \bibinfo{year}{2017}.
\newblock \bibinfo{title}{Classification of dental diseases using cnn and
  transfer learning}, in: \bibinfo{booktitle}{2017 5th International Symposium
  on Computational and Business Intelligence (ISCBI)},
  \bibinfo{organization}{IEEE}. pp. \bibinfo{pages}{70--74}.
\bibitem[{Proch{\'a}zka et~al.(2018)Proch{\'a}zka, Charv{\'a}tov{\'a}, Vaseghi
  and Vy{\v{s}}ata}]{t34}
\bibinfo{author}{Proch{\'a}zka, A.}, \bibinfo{author}{Charv{\'a}tov{\'a}, H.},
  \bibinfo{author}{Vaseghi, S.}, \bibinfo{author}{Vy{\v{s}}ata, O.},
  \bibinfo{year}{2018}.
\newblock \bibinfo{title}{Machine learning in rehabilitation assessment for
  thermal and heart rate data processing}.
\newblock \bibinfo{journal}{IEEE Transactions on Neural Systems and
  Rehabilitation Engineering} \bibinfo{volume}{26},
  \bibinfo{pages}{1209--1214}.
\bibitem[{Puaschunder(2020)}]{t133}
\bibinfo{author}{Puaschunder, J.M.}, \bibinfo{year}{2020}.
\newblock \bibinfo{title}{The potential for artificial intelligence in
  healthcare}.
\newblock \bibinfo{journal}{Available at SSRN 3525037} .
\bibitem[{Punia et~al.(2020)Punia, Kumar, Mujahid and Rohilla}]{t54}
\bibinfo{author}{Punia, R.}, \bibinfo{author}{Kumar, L.},
  \bibinfo{author}{Mujahid, M.}, \bibinfo{author}{Rohilla, R.},
  \bibinfo{year}{2020}.
\newblock \bibinfo{title}{Computer vision and radiology for covid-19
  detection}, in: \bibinfo{booktitle}{2020 International Conference for
  Emerging Technology (INCET)}, \bibinfo{organization}{IEEE}. pp.
  \bibinfo{pages}{1--5}.
\bibitem[{Rahman et~al.(2015)Rahman, Tereshchenko, Kongkatong, Abraham, Abraham
  and Shatkay}]{t28}
\bibinfo{author}{Rahman, Q.A.}, \bibinfo{author}{Tereshchenko, L.G.},
  \bibinfo{author}{Kongkatong, M.}, \bibinfo{author}{Abraham, T.},
  \bibinfo{author}{Abraham, M.R.}, \bibinfo{author}{Shatkay, H.},
  \bibinfo{year}{2015}.
\newblock \bibinfo{title}{Utilizing ecg-based heartbeat classification for
  hypertrophic cardiomyopathy identification}.
\newblock \bibinfo{journal}{IEEE transactions on nanobioscience}
  \bibinfo{volume}{14}, \bibinfo{pages}{505--512}.
\bibitem[{Reamaroon et~al.(2018)Reamaroon, Sjoding, Lin, Iwashyna and
  Najarian}]{t35}
\bibinfo{author}{Reamaroon, N.}, \bibinfo{author}{Sjoding, M.W.},
  \bibinfo{author}{Lin, K.}, \bibinfo{author}{Iwashyna, T.J.},
  \bibinfo{author}{Najarian, K.}, \bibinfo{year}{2018}.
\newblock \bibinfo{title}{Accounting for label uncertainty in machine learning
  for detection of acute respiratory distress syndrome}.
\newblock \bibinfo{journal}{IEEE journal of biomedical and health informatics}
  \bibinfo{volume}{23}, \bibinfo{pages}{407--415}.
\bibitem[{Sadr and de~Chazal(2016)}]{t38}
\bibinfo{author}{Sadr, N.}, \bibinfo{author}{de~Chazal, P.},
  \bibinfo{year}{2016}.
\newblock \bibinfo{title}{Comparing ecg derived respiratory signals and chest
  respiratory signal for the detection of obstructive sleep apnoea}, in:
  \bibinfo{booktitle}{2016 Computing in Cardiology Conference (CinC)},
  \bibinfo{organization}{IEEE}. pp. \bibinfo{pages}{1029--1032}.
\bibitem[{Sahoo et~al.(2020)Sahoo, Pradhan and Das}]{t81}
\bibinfo{author}{Sahoo, A.K.}, \bibinfo{author}{Pradhan, C.},
  \bibinfo{author}{Das, H.}, \bibinfo{year}{2020}.
\newblock \bibinfo{title}{Performance evaluation of different machine learning
  methods and deep-learning based convolutional neural network for health
  decision making}, in: \bibinfo{booktitle}{Nature inspired computing for data
  science}. \bibinfo{publisher}{Springer}, pp. \bibinfo{pages}{201--212}.
\bibitem[{Santhakumar et~al.(2016)Santhakumar, Tandur, Rajkumar, Geetha, Haritz
  and Rajamani}]{t126}
\bibinfo{author}{Santhakumar, R.}, \bibinfo{author}{Tandur, M.},
  \bibinfo{author}{Rajkumar, E.}, \bibinfo{author}{Geetha, K.},
  \bibinfo{author}{Haritz, G.}, \bibinfo{author}{Rajamani, K.T.},
  \bibinfo{year}{2016}.
\newblock \bibinfo{title}{Machine learning algorithm for retinal image
  analysis}, in: \bibinfo{booktitle}{2016 IEEE Region 10 Conference (TENCON)},
  \bibinfo{organization}{IEEE}. pp. \bibinfo{pages}{1236--1240}.
\bibitem[{Sau and Bhakta(2019)}]{t84}
\bibinfo{author}{Sau, A.}, \bibinfo{author}{Bhakta, I.}, \bibinfo{year}{2019}.
\newblock \bibinfo{title}{Screening of anxiety and depression among seafarers
  using machine learning technology}.
\newblock \bibinfo{journal}{Informatics in Medicine Unlocked}
  \bibinfo{volume}{16}, \bibinfo{pages}{100228}.
\bibitem[{Seixas et~al.(2015)Seixas, Barbon and Mantovani}]{t110}
\bibinfo{author}{Seixas, J.L.}, \bibinfo{author}{Barbon, S.},
  \bibinfo{author}{Mantovani, R.G.}, \bibinfo{year}{2015}.
\newblock \bibinfo{title}{Pattern recognition of lower member skin ulcers in
  medical images with machine learning algorithms}, in:
  \bibinfo{booktitle}{2015 IEEE 28th International Symposium on Computer-Based
  Medical Systems}, \bibinfo{organization}{IEEE}. pp. \bibinfo{pages}{50--53}.
\bibitem[{Selvathi and Sharnitha(2011)}]{t95}
\bibinfo{author}{Selvathi, D.}, \bibinfo{author}{Sharnitha, V.},
  \bibinfo{year}{2011}.
\newblock \bibinfo{title}{Thyroid classification and segmentation in ultrasound
  images using machine learning algorithms}, in: \bibinfo{booktitle}{2011
  International Conference on Signal Processing, Communication, Computing and
  Networking Technologies}, \bibinfo{organization}{IEEE}. pp.
  \bibinfo{pages}{836--841}.
\bibitem[{Selvathi and Suganya(2019)}]{t125}
\bibinfo{author}{Selvathi, D.}, \bibinfo{author}{Suganya, K.},
  \bibinfo{year}{2019}.
\newblock \bibinfo{title}{Support vector machine based method for automatic
  detection of diabetic eye disease using thermal images}, in:
  \bibinfo{booktitle}{2019 1st International Conference on Innovations in
  Information and Communication Technology (ICIICT)},
  \bibinfo{organization}{IEEE}. pp. \bibinfo{pages}{1--6}.
\bibitem[{Sethy and Behera(2020)}]{t45}
\bibinfo{author}{Sethy, P.K.}, \bibinfo{author}{Behera, S.K.},
  \bibinfo{year}{2020}.
\newblock \bibinfo{title}{Detection of coronavirus disease (covid-19) based on
  deep features} .
\bibitem[{Sharma and Rani(2017)}]{t71}
\bibinfo{author}{Sharma, A.}, \bibinfo{author}{Rani, R.}, \bibinfo{year}{2017}.
\newblock \bibinfo{title}{Classification of cancerous profiles using machine
  learning}, in: \bibinfo{booktitle}{2017 International Conference on Machine
  Learning and Data Science (MLDS)}, \bibinfo{organization}{IEEE}. pp.
  \bibinfo{pages}{31--36}.
\bibitem[{Shetty and Shah(2018)}]{t66}
\bibinfo{author}{Shetty, A.}, \bibinfo{author}{Shah, V.}, \bibinfo{year}{2018}.
\newblock \bibinfo{title}{Survey of cervical cancer prediction using machine
  learning: A comparative approach}, in: \bibinfo{booktitle}{2018 9th
  International Conference on Computing, Communication and Networking
  Technologies (ICCCNT)}, \bibinfo{organization}{IEEE}. pp.
  \bibinfo{pages}{1--6}.
\bibitem[{Shibly et~al.(2020)Shibly, Dey, Islam and Rahman}]{t55}
\bibinfo{author}{Shibly, K.H.}, \bibinfo{author}{Dey, S.K.},
  \bibinfo{author}{Islam, M.T.U.}, \bibinfo{author}{Rahman, M.M.},
  \bibinfo{year}{2020}.
\newblock \bibinfo{title}{Covid faster r--cnn: A novel framework to diagnose
  novel coronavirus disease (covid-19) in x-ray images}.
\newblock \bibinfo{journal}{Informatics in Medicine Unlocked}
  \bibinfo{volume}{20}, \bibinfo{pages}{100405}.
\bibitem[{Shrestha and Mahmood(2019)}]{t9}
\bibinfo{author}{Shrestha, A.}, \bibinfo{author}{Mahmood, A.},
  \bibinfo{year}{2019}.
\newblock \bibinfo{title}{Review of deep learning algorithms and
  architectures}.
\newblock \bibinfo{journal}{IEEE Access} \bibinfo{volume}{7},
  \bibinfo{pages}{53040--53065}.
\bibitem[{Singh et~al.(2016)Singh, Thakur and Sharma}]{t2}
\bibinfo{author}{Singh, A.}, \bibinfo{author}{Thakur, N.},
  \bibinfo{author}{Sharma, A.}, \bibinfo{year}{2016}.
\newblock \bibinfo{title}{A review of supervised machine learning algorithms},
  in: \bibinfo{booktitle}{2016 3rd International Conference on Computing for
  Sustainable Global Development (INDIACom)}, \bibinfo{organization}{Ieee}. pp.
  \bibinfo{pages}{1310--1315}.
\bibitem[{Song et~al.(2021)Song, Zheng, Li, Zhang, Zhang, Huang, Chen, Wang,
  Zhao, Zha et~al.}]{t46}
\bibinfo{author}{Song, Y.}, \bibinfo{author}{Zheng, S.}, \bibinfo{author}{Li,
  L.}, \bibinfo{author}{Zhang, X.}, \bibinfo{author}{Zhang, X.},
  \bibinfo{author}{Huang, Z.}, \bibinfo{author}{Chen, J.},
  \bibinfo{author}{Wang, R.}, \bibinfo{author}{Zhao, H.}, \bibinfo{author}{Zha,
  Y.}, et~al., \bibinfo{year}{2021}.
\newblock \bibinfo{title}{Deep learning enables accurate diagnosis of novel
  coronavirus (covid-19) with ct images}.
\newblock \bibinfo{journal}{IEEE/ACM Transactions on Computational Biology and
  Bioinformatics} .
\bibitem[{Stamate et~al.(2018)Stamate, Alghamdi, Ogg, Hoile and Murtagh}]{t100}
\bibinfo{author}{Stamate, D.}, \bibinfo{author}{Alghamdi, W.},
  \bibinfo{author}{Ogg, J.}, \bibinfo{author}{Hoile, R.},
  \bibinfo{author}{Murtagh, F.}, \bibinfo{year}{2018}.
\newblock \bibinfo{title}{A machine learning framework for predicting dementia
  and mild cognitive impairment}, in: \bibinfo{booktitle}{2018 17th IEEE
  International Conference on Machine Learning and Applications (ICMLA)},
  \bibinfo{organization}{IEEE}. pp. \bibinfo{pages}{671--678}.
\bibitem[{Stoean et~al.(2006)Stoean, Stoean, Preuss, El-Darzi and
  Dumitrescu}]{t89}
\bibinfo{author}{Stoean, R.}, \bibinfo{author}{Stoean, C.},
  \bibinfo{author}{Preuss, M.}, \bibinfo{author}{El-Darzi, E.},
  \bibinfo{author}{Dumitrescu, D.}, \bibinfo{year}{2006}.
\newblock \bibinfo{title}{Evolutionary support vector machines for diabetes
  mellitus diagnosis}, in: \bibinfo{booktitle}{2006 3rd International IEEE
  Conference Intelligent Systems}, \bibinfo{organization}{IEEE}. pp.
  \bibinfo{pages}{182--187}.
\bibitem[{Subhani et~al.(2017)Subhani, Mumtaz, Saad, Kamel and Malik}]{t82}
\bibinfo{author}{Subhani, A.R.}, \bibinfo{author}{Mumtaz, W.},
  \bibinfo{author}{Saad, M.N.B.M.}, \bibinfo{author}{Kamel, N.},
  \bibinfo{author}{Malik, A.S.}, \bibinfo{year}{2017}.
\newblock \bibinfo{title}{Machine learning framework for the detection of
  mental stress at multiple levels}.
\newblock \bibinfo{journal}{IEEE Access} \bibinfo{volume}{5},
  \bibinfo{pages}{13545--13556}.
\bibitem[{Tabassian et~al.(2018)Tabassian, Sunderji, Erdei, Sanchez-Martinez,
  Degiovanni, Marino, Fraser and D'hooge}]{t26}
\bibinfo{author}{Tabassian, M.}, \bibinfo{author}{Sunderji, I.},
  \bibinfo{author}{Erdei, T.}, \bibinfo{author}{Sanchez-Martinez, S.},
  \bibinfo{author}{Degiovanni, A.}, \bibinfo{author}{Marino, P.},
  \bibinfo{author}{Fraser, A.G.}, \bibinfo{author}{D'hooge, J.},
  \bibinfo{year}{2018}.
\newblock \bibinfo{title}{Diagnosis of heart failure with preserved ejection
  fraction: machine learning of spatiotemporal variations in left ventricular
  deformation}.
\newblock \bibinfo{journal}{Journal of the American society of
  echocardiography} \bibinfo{volume}{31}, \bibinfo{pages}{1272--1284}.
\bibitem[{Tamin and Iswari(2017)}]{t92}
\bibinfo{author}{Tamin, F.}, \bibinfo{author}{Iswari, N.M.S.},
  \bibinfo{year}{2017}.
\newblock \bibinfo{title}{Implementation of c4. 5 algorithm to determine
  hospital readmission rate of diabetes patient}, in: \bibinfo{booktitle}{2017
  4th International Conference on New Media Studies (CONMEDIA)},
  \bibinfo{organization}{IEEE}. pp. \bibinfo{pages}{15--18}.
\bibitem[{Tataru et~al.(2017)Tataru, Yi, Shenoyas and Ma}]{t60}
\bibinfo{author}{Tataru, C.}, \bibinfo{author}{Yi, D.},
  \bibinfo{author}{Shenoyas, A.}, \bibinfo{author}{Ma, A.},
  \bibinfo{year}{2017}.
\newblock \bibinfo{title}{Deep learning for abnormality detection in chest
  x-ray images}, in: \bibinfo{booktitle}{IEEE Conference on Deep Learning}.
\bibitem[{Telnoni et~al.(2019)Telnoni, Budiawan and Qana’a}]{t4}
\bibinfo{author}{Telnoni, P.A.}, \bibinfo{author}{Budiawan, R.},
  \bibinfo{author}{Qana’a, M.}, \bibinfo{year}{2019}.
\newblock \bibinfo{title}{Comparison of machine learning classification method
  on text-based case in twitter}, in: \bibinfo{booktitle}{2019 International
  Conference on ICT for Smart Society (ICISS)}, \bibinfo{organization}{IEEE}.
  pp. \bibinfo{pages}{1--5}.
\bibitem[{Ting et~al.(2019)Ting, Pasquale, Peng, Campbell, Lee, Raman, Tan,
  Schmetterer, Keane and Wong}]{t123}
\bibinfo{author}{Ting, D.S.W.}, \bibinfo{author}{Pasquale, L.R.},
  \bibinfo{author}{Peng, L.}, \bibinfo{author}{Campbell, J.P.},
  \bibinfo{author}{Lee, A.Y.}, \bibinfo{author}{Raman, R.},
  \bibinfo{author}{Tan, G.S.W.}, \bibinfo{author}{Schmetterer, L.},
  \bibinfo{author}{Keane, P.A.}, \bibinfo{author}{Wong, T.Y.},
  \bibinfo{year}{2019}.
\newblock \bibinfo{title}{Artificial intelligence and deep learning in
  ophthalmology}.
\newblock \bibinfo{journal}{British Journal of Ophthalmology}
  \bibinfo{volume}{103}, \bibinfo{pages}{167--175}.
\bibitem[{Tsehay et~al.(2017)Tsehay, Lay, Wang, Kwak, Turkbey, Choyke, Pinto,
  Wood and Summers}]{t70}
\bibinfo{author}{Tsehay, Y.}, \bibinfo{author}{Lay, N.}, \bibinfo{author}{Wang,
  X.}, \bibinfo{author}{Kwak, J.T.}, \bibinfo{author}{Turkbey, B.},
  \bibinfo{author}{Choyke, P.}, \bibinfo{author}{Pinto, P.},
  \bibinfo{author}{Wood, B.}, \bibinfo{author}{Summers, R.M.},
  \bibinfo{year}{2017}.
\newblock \bibinfo{title}{Biopsy-guided learning with deep convolutional neural
  networks for prostate cancer detection on multiparametric mri}, in:
  \bibinfo{booktitle}{2017 IEEE 14th International Symposium on Biomedical
  Imaging (ISBI 2017)}, \bibinfo{organization}{IEEE}. pp.
  \bibinfo{pages}{642--645}.
\bibitem[{Tuan et~al.(2017)Tuan, Duc, Van~Hai et~al.}]{t14}
\bibinfo{author}{Tuan, T.M.}, \bibinfo{author}{Duc, N.T.},
  \bibinfo{author}{Van~Hai, P.}, et~al., \bibinfo{year}{2017}.
\newblock \bibinfo{title}{Dental diagnosis from x-ray images using fuzzy
  rule-based systems}.
\newblock \bibinfo{journal}{International Journal of Fuzzy System Applications
  (IJFSA)} \bibinfo{volume}{6}, \bibinfo{pages}{1--16}.
\bibitem[{Turki(2018)}]{t67}
\bibinfo{author}{Turki, T.}, \bibinfo{year}{2018}.
\newblock \bibinfo{title}{An empirical study of machine learning algorithms for
  cancer identification}, in: \bibinfo{booktitle}{2018 IEEE 15th International
  Conference on Networking, Sensing and Control (ICNSC)},
  \bibinfo{organization}{IEEE}. pp. \bibinfo{pages}{1--5}.
\bibitem[{Vamathevan et~al.(2019)Vamathevan, Clark, Czodrowski, Dunham, Ferran,
  Lee, Li, Madabhushi, Shah, Spitzer et~al.}]{t131}
\bibinfo{author}{Vamathevan, J.}, \bibinfo{author}{Clark, D.},
  \bibinfo{author}{Czodrowski, P.}, \bibinfo{author}{Dunham, I.},
  \bibinfo{author}{Ferran, E.}, \bibinfo{author}{Lee, G.}, \bibinfo{author}{Li,
  B.}, \bibinfo{author}{Madabhushi, A.}, \bibinfo{author}{Shah, P.},
  \bibinfo{author}{Spitzer, M.}, et~al., \bibinfo{year}{2019}.
\newblock \bibinfo{title}{Applications of machine learning in drug discovery
  and development}.
\newblock \bibinfo{journal}{Nature Reviews Drug Discovery}
  \bibinfo{volume}{18}, \bibinfo{pages}{463--477}.
\bibitem[{Vera et~al.(2010)Vera, Garcia, Suarez, Hernando, Redondo, Corchado,
  Sanchez, Gil and Sedano}]{t15}
\bibinfo{author}{Vera, V.}, \bibinfo{author}{Garcia, A.E.},
  \bibinfo{author}{Suarez, M.J.}, \bibinfo{author}{Hernando, B.},
  \bibinfo{author}{Redondo, R.}, \bibinfo{author}{Corchado, E.},
  \bibinfo{author}{Sanchez, M.A.}, \bibinfo{author}{Gil, A.},
  \bibinfo{author}{Sedano, J.}, \bibinfo{year}{2010}.
\newblock \bibinfo{title}{Optimizing a dental milling process by means of soft
  computing techniques}, in: \bibinfo{booktitle}{2010 10th International
  Conference on Intelligent Systems Design and Applications},
  \bibinfo{organization}{IEEE}. pp. \bibinfo{pages}{1430--1435}.
\bibitem[{Vyas et~al.(2014)Vyas, Meyerle and Burlina}]{t114}
\bibinfo{author}{Vyas, S.}, \bibinfo{author}{Meyerle, J.},
  \bibinfo{author}{Burlina, P.}, \bibinfo{year}{2014}.
\newblock \bibinfo{title}{Cross validating hyperspectral with ultrasound-based
  skin thickness estimation}, in: \bibinfo{booktitle}{2014 6th Workshop on
  Hyperspectral Image and Signal Processing: Evolution in Remote Sensing
  (WHISPERS)}, \bibinfo{organization}{IEEE}. pp. \bibinfo{pages}{1--4}.
\bibitem[{Wahid et~al.(2015)Wahid, Begg, Hass, Halgamuge and Ackland}]{t101}
\bibinfo{author}{Wahid, F.}, \bibinfo{author}{Begg, R.K.},
  \bibinfo{author}{Hass, C.J.}, \bibinfo{author}{Halgamuge, S.},
  \bibinfo{author}{Ackland, D.C.}, \bibinfo{year}{2015}.
\newblock \bibinfo{title}{Classification of parkinson's disease gait using
  spatial-temporal gait features}.
\newblock \bibinfo{journal}{IEEE journal of biomedical and health informatics}
  \bibinfo{volume}{19}, \bibinfo{pages}{1794--1802}.
\bibitem[{Wang et~al.(2020)Wang, Lin and Wong}]{t42}
\bibinfo{author}{Wang, L.}, \bibinfo{author}{Lin, Z.Q.}, \bibinfo{author}{Wong,
  A.}, \bibinfo{year}{2020}.
\newblock \bibinfo{title}{Covid-net: A tailored deep convolutional neural
  network design for detection of covid-19 cases from chest x-ray images}.
\newblock \bibinfo{journal}{Scientific Reports} \bibinfo{volume}{10},
  \bibinfo{pages}{1--12}.
\bibitem[{Wei et~al.(2005)Wei, Yang, Nishikawa and Jiang}]{t63}
\bibinfo{author}{Wei, L.}, \bibinfo{author}{Yang, Y.},
  \bibinfo{author}{Nishikawa, R.M.}, \bibinfo{author}{Jiang, Y.},
  \bibinfo{year}{2005}.
\newblock \bibinfo{title}{A study on several machine-learning methods for
  classification of malignant and benign clustered microcalcifications}.
\newblock \bibinfo{journal}{IEEE transactions on medical imaging}
  \bibinfo{volume}{24}, \bibinfo{pages}{371--380}.
\bibitem[{Wickramasinghe et~al.(2017)Wickramasinghe, Perera and
  Kahandawaarachchi}]{t121}
\bibinfo{author}{Wickramasinghe, M.}, \bibinfo{author}{Perera, D.},
  \bibinfo{author}{Kahandawaarachchi, K.}, \bibinfo{year}{2017}.
\newblock \bibinfo{title}{Dietary prediction for patients with chronic kidney
  disease (ckd) by considering blood potassium level using machine learning
  algorithms}, in: \bibinfo{booktitle}{2017 IEEE Life Sciences Conference
  (LSC)}, \bibinfo{organization}{IEEE}. pp. \bibinfo{pages}{300--303}.
\bibitem[{Wongkoblap et~al.(2018)Wongkoblap, Vadillo and Curcin}]{t79}
\bibinfo{author}{Wongkoblap, A.}, \bibinfo{author}{Vadillo, M.A.},
  \bibinfo{author}{Curcin, V.}, \bibinfo{year}{2018}.
\newblock \bibinfo{title}{Classifying depressed users with multiple instance
  learning from social network data}, in: \bibinfo{booktitle}{2018 IEEE
  International Conference on Healthcare Informatics (ICHI)},
  \bibinfo{organization}{IEEE}. pp. \bibinfo{pages}{436--436}.
\bibitem[{Wu et~al.(1993)Wu, Giger, Doi, Vyborny, Schmidt and Metz}]{t64}
\bibinfo{author}{Wu, Y.}, \bibinfo{author}{Giger, M.L.}, \bibinfo{author}{Doi,
  K.}, \bibinfo{author}{Vyborny, C.J.}, \bibinfo{author}{Schmidt, R.A.},
  \bibinfo{author}{Metz, C.E.}, \bibinfo{year}{1993}.
\newblock \bibinfo{title}{Artificial neural networks in mammography:
  application to decision making in the diagnosis of breast cancer.}
\newblock \bibinfo{journal}{Radiology} \bibinfo{volume}{187},
  \bibinfo{pages}{81--87}.
\bibitem[{Xie et~al.(2017)Xie, Wang, Zhang, Meng, Kong, Mao, Xu and
  Zhang}]{t31}
\bibinfo{author}{Xie, J.}, \bibinfo{author}{Wang, H.}, \bibinfo{author}{Zhang,
  J.}, \bibinfo{author}{Meng, C.}, \bibinfo{author}{Kong, Y.},
  \bibinfo{author}{Mao, S.}, \bibinfo{author}{Xu, L.}, \bibinfo{author}{Zhang,
  W.}, \bibinfo{year}{2017}.
\newblock \bibinfo{title}{A novel hybrid subset-learning method for predicting
  risk factors of atherosclerosis}, in: \bibinfo{booktitle}{2017 IEEE
  International Conference on Bioinformatics and Biomedicine (BIBM)},
  \bibinfo{organization}{IEEE}. pp. \bibinfo{pages}{2124--2131}.
\bibitem[{Xu and Meng()}]{t47}
\bibinfo{author}{Xu, B.}, \bibinfo{author}{Meng, X.}, .
\newblock \bibinfo{title}{A deep learning algorithm using ct images to screen
  for corona virus disease (covid-19)}.
\newblock \bibinfo{journal}{Preprint} .
\bibitem[{Xu et~al.(2020)Xu, Jiang, Ma, Du, Li, Lv, Yu, Ni, Chen, Su
  et~al.}]{t49}
\bibinfo{author}{Xu, X.}, \bibinfo{author}{Jiang, X.}, \bibinfo{author}{Ma,
  C.}, \bibinfo{author}{Du, P.}, \bibinfo{author}{Li, X.}, \bibinfo{author}{Lv,
  S.}, \bibinfo{author}{Yu, L.}, \bibinfo{author}{Ni, Q.},
  \bibinfo{author}{Chen, Y.}, \bibinfo{author}{Su, J.}, et~al.,
  \bibinfo{year}{2020}.
\newblock \bibinfo{title}{A deep learning system to screen novel coronavirus
  disease 2019 pneumonia}.
\newblock \bibinfo{journal}{Engineering} \bibinfo{volume}{6},
  \bibinfo{pages}{1122--1129}.
\bibitem[{Yang et~al.(2017)Yang, Fasching and Tresp}]{t12}
\bibinfo{author}{Yang, Y.}, \bibinfo{author}{Fasching, P.A.},
  \bibinfo{author}{Tresp, V.}, \bibinfo{year}{2017}.
\newblock \bibinfo{title}{Predictive modeling of therapy decisions in
  metastatic breast cancer with recurrent neural network encoder and
  multinomial hierarchical regression decoder}, in: \bibinfo{booktitle}{2017
  IEEE International Conference on Healthcare Informatics (ICHI)},
  \bibinfo{organization}{IEEE}. pp. \bibinfo{pages}{46--55}.
\bibitem[{Zakeri and Tavakolian(2015)}]{t39}
\bibinfo{author}{Zakeri, V.}, \bibinfo{author}{Tavakolian, K.},
  \bibinfo{year}{2015}.
\newblock \bibinfo{title}{Identification of respiratory phases using
  seismocardiogram: A machine learning approach}, in: \bibinfo{booktitle}{2015
  Computing in Cardiology Conference (CinC)}, \bibinfo{organization}{IEEE}. pp.
  \bibinfo{pages}{305--308}.
\bibitem[{Zhang and Wang(2007)}]{t97}
\bibinfo{author}{Zhang, W.L.}, \bibinfo{author}{Wang, X.Z.},
  \bibinfo{year}{2007}.
\newblock \bibinfo{title}{Feature extraction and classification for human brain
  ct images}, in: \bibinfo{booktitle}{2007 International Conference on Machine
  Learning and Cybernetics}, \bibinfo{organization}{IEEE}. pp.
  \bibinfo{pages}{1155--1159}.
\bibitem[{Zhang et~al.(2017)Zhang, Wang, Liu and Tao}]{t113}
\bibinfo{author}{Zhang, X.}, \bibinfo{author}{Wang, S.}, \bibinfo{author}{Liu,
  J.}, \bibinfo{author}{Tao, C.}, \bibinfo{year}{2017}.
\newblock \bibinfo{title}{Computer-aided diagnosis of four common cutaneous
  diseases using deep learning algorithm}, in: \bibinfo{booktitle}{2017 IEEE
  International Conference on Bioinformatics and Biomedicine (BIBM)},
  \bibinfo{organization}{IEEE}. pp. \bibinfo{pages}{1304--1306}.
\bibitem[{Zheng et~al.(2020)Zheng, Deng, Fu, Zhou, Feng, Ma, Liu and
  Wang}]{t48}
\bibinfo{author}{Zheng, C.}, \bibinfo{author}{Deng, X.}, \bibinfo{author}{Fu,
  Q.}, \bibinfo{author}{Zhou, Q.}, \bibinfo{author}{Feng, J.},
  \bibinfo{author}{Ma, H.}, \bibinfo{author}{Liu, W.}, \bibinfo{author}{Wang,
  X.}, \bibinfo{year}{2020}.
\newblock \bibinfo{title}{Deep learning-based detection for covid-19 from chest
  ct using weak label}.
\newblock \bibinfo{journal}{MedRxiv} .

\end{thebibliography}

\end{document}